\documentclass[12pt]{arxiv}

\title[Communication trade-offs for synchronized distributed SGD]{Communication trade-offs for synchronized distributed SGD with large step size}

\arxivauthor{%
 \Name{Kumar Kshitij Patel} \Email{kishinmh@cse.iitk.ac.in}\\
\addr Department of Computer Science, IIT Kanpur, India
 \AND
 \Name{Aymeric Dieuleveut} \Email{aymeric.dieuleveut@epfl.ch}\\
\addr Department of Computer Science, EPFL, Switzerland
}

\usepackage{adjustbox}

\usepackage[utf8]{inputenc} 
\usepackage[T1]{fontenc}    
\usepackage{url}            
\usepackage{booktabs}       
\usepackage{amsfonts}       
\usepackage{nicefrac}       
\usepackage{microtype}      
\usepackage{natbib}
\usepackage{algorithmicx}
\usepackage{times}

\usepackage[utf8]{inputenc}
\usepackage{url}            
\usepackage{booktabs}       
\usepackage{microtype}      

\usepackage{algorithmicx}
\usepackage[boxed]{algorithm}
\usepackage[noend]{algpseudocode}

\RequirePackage{amsmath}

\usepackage{stmaryrd}

\usepackage{amssymb}
\usepackage{makeidx}
\usepackage[english]{babel}
\usepackage{graphicx}
\usepackage{amsfonts,amssymb}
\usepackage{oldgerm}
\usepackage{mathrsfs}
\usepackage[active]{srcltx}
\usepackage{verbatim}
\usepackage{enumitem, kantlipsum} 
\usepackage{aliascnt}
\usepackage{bbm}
\usepackage{array}
\usepackage[textwidth=4cm,textsize=footnotesize]{todonotes}
\usepackage{xargs}
\usepackage{cellspace}
\usepackage{xr}

\usepackage[Symbolsmallscale]{upgreek}
\usepackage{xcolor}
\definecolor{mydarkblue}{rgb}{0,0.08,0.45}
\usepackage{multicol} 
\usepackage{subcaption} 
\usepackage{wrapfig}

\usepackage{dsfont}
\usepackage{aliascnt}
\usepackage{cleveref}
\makeatletter

\crefname{theorem}{theorem}{Theorems}
\Crefname{Theorem}{Theorem}{Theorems}

\newtheorem*{lemma_nonumber*}{Lemma}

\newaliascnt{lemma}{theorem}
\aliascntresetthe{lemma}
\crefname{lemma}{lemma}{lemmas}
\Crefname{Lemma}{Lemma}{Lemmas}

\newaliascnt{corollary}{theorem}
\aliascntresetthe{corollary}
\crefname{corollary}{corollary}{corollaries}
\Crefname{Corollary}{Corollary}{Corollaries}

\newaliascnt{proposition}{theorem}
\aliascntresetthe{proposition}
\crefname{proposition}{proposition}{propositions}
\Crefname{Proposition}{Proposition}{Propositions}

\newaliascnt{definition}{theorem}
\aliascntresetthe{definition}
\crefname{definition}{definition}{definitions}
\Crefname{Definition}{Definition}{Definitions}

\newaliascnt{remark}{theorem}
\aliascntresetthe{remark}
\crefname{remark}{remark}{remarks}
\Crefname{Remark}{Remark}{Remarks}

\crefname{example}{example}{examples}
\Crefname{Example}{Example}{Examples}

\crefname{figure}{figure}{figures}
\Crefname{Figure}{Figure}{Figures}

\newtheorem{assumption}{\textbf{A}\hspace{-3pt}}
\crefformat{assumption}{{\textbf{A}}#2#1#3}

\Crefname{assumptionG}{\textbf{G}\hspace{-3pt}}{\textbf{G}\hspace{-3pt}}
\crefname{assumptionG}{\textbf{G}}{\textbf{G}}

\newtheorem{assumptionQ}{\textbf{Q}\hspace{-3pt}}
\Crefname{assumptionQ}{\textbf{Q}\hspace{-3pt}}{\textbf{Q}\hspace{-3pt}}
\crefname{assumptionQ}{\textbf{Q}}{\textbf{Q}}

\newcounter{cmtcounter}
\let\OLDthebibliography\thebibliography
\renewcommand\thebibliography[1]{
	\OLDthebibliography{#1}
	\setlength{\parskip}{2.0pt}
	\setlength{\itemsep}{1pt plus 1ex}
}

\usepackage[boxed]{algorithm}
\usepackage[noend]{algpseudocode}
\algblock{ParFor}{EndParFor}
\algnewcommand\algorithmicparfor{\textbf{parfor}}
\algnewcommand\algorithmicpardo{\textbf{do}}
\algnewcommand\algorithmicendparfor{\textbf{end\ parfor}}
\algrenewtext{ParFor}[1]{\algorithmicparfor\ #1\ \algorithmicpardo}
\algrenewtext{EndParFor}{\algorithmicendparfor}\allowdisplaybreaks

\newcommandx{\functionspace}[2][1=+]{\mathbb{F}_{#1}(#2)}

\newcommandx{\VarDeux}[3][3=]{\operatorname{Var}^{#3}_{#1}\left\{#2 \right\}}

\newcommand{\LeftEqNo}{\let\veqno\@@leqno}

\newcommand{\N}{\ensuremath{\mathbb{N}}}

\newcommand{\R}{\ensuremath{\mathbb{R}}}

\newcommand{\PE}{\mathbb{E}}

\newcommandx{\Vnorm}[2][1=V]{\| #2 \|_{#1}}
\newcommandx{\VnormEq}[2][1=V]{\left\| #2 \right\|_{#1}}
\newcommandx{\mynorm}[2][1=]{\ifthenelse{\equal{#1}{}}{\left\Vert #2 \right\Vert}{\left\Vert #2 \right\Vert^{#1}}}
\newcommandx{\normLigne}[2][1=]{\ifthenelse{\equal{#1}{}}{\Vert #2 \Vert}{\Vert #2\Vert^{#1}}}

\newcommand{\ps}[2]{\left\langle#1,#2 \right\rangle}

\def\rset{\mathbb{R}}
\def\nset{\mathbb{N}}

\def\nsets{\mathbb{N}^*}

\newcommandx\probaMarkovTilde[2][2=]
{\ifthenelse{\equal{#2}{}}{{\widetilde{\mathbb{P}}_{#1}}}{\widetilde{\mathbb{P}}_{#1}\left[ #2\right]}}

\newcommand{\expe}[1]{\PE \left[ #1 \right]}

\newcommand{\expeLigne}[1]{\PE [ #1 ]}
\newcommand{\expeLine}[1]{\PE [ #1 ]}

\newcounter{hypoconbis}
\newcounter{saveconbis}
\newcommand\debutH{\begin{list}
{\textbf{H\arabic{hypoconbis}}}{\usecounter{hypoconbis}}\setcounter{hypoconbis}{\value{saveconbis}}}
\newcommand\finH{\end{list}\setcounter{saveconbis}{\value{hypoconbis}}}

\def\ie{\textit{i.e.}}
\def\eqsp{\;}

\newcommandx{\weight}[2][2=n]{\omega_{#1,#2}^N}

\def\as{\ensuremath{\text{a.s.}}}

\newcommandx\sequence[3][2=,3=]
{\ifthenelse{\equal{#3}{}}{\ensuremath{\{ #1_{#2}\}}}{\ensuremath{\{ #1_{#2}, \eqsp #2 \in #3 \}}}}
\newcommandx{\sequencen}[2][2=n\in\N]{\ensuremath{\{ #1, \eqsp #2 \}}}
\newcommandx\sequenceDouble[4][3=,4=]
{\ifthenelse{\equal{#3}{}}{\ensuremath{\{ (#1_{#3},#2_{#3}) \}}}{\ensuremath{\{  (#1_{#3},#2_{#3}), \eqsp #3 \in #4 \}}}}
\newcommandx{\sequencenDouble}[3][3=n\in\N]{\ensuremath{\{ (#1_{n},#2_{n}), \eqsp #3 \}}}

\def\iid{i.i.d.}

\def\eg{\textit{e.g.}}

\newcommand{\BEAS}{\begin{eqnarray*}}
\newcommand{\EEAS}{\end{eqnarray*}}
\newcommand{\BEA}{\begin{eqnarray}}
\newcommand{\EEA}{\end{eqnarray}}
\newcommand{\BEQ}{\begin{equation}}
\newcommand{\EEQ}{\end{equation}}
\newcommand{\BIT}{\begin{itemize}}
\newcommand{\EIT}{\end{itemize}}
\newcommand{\BNUM}{\begin{enumerate}}
\newcommand{\ENUM}{\end{enumerate}}
\newcommand{\BA}{\begin{array}}
\newcommand{\EA}{\end{array}}

\def \E{{\mathbb E}}
\def \R{{\mathbb R}}

\def \E{{\mathbb E}}

\def \N{{\mathbb N}}

\newcommand{\h}[2]{\mathcal{H}_{#2}^{#1}}

\newcommandx{\tharg}[2][2=\gamma]{\theta_{#1}^{(#2)}}

\newcommand{\tavd}[1]{\bar \theta_{2\gamma}}

\renewcommand{\epsilon}{\varepsilon}

\newcommand{\opnorm}[1]{{\left\vert\kern-0.25ex\left\vert\kern-0.25ex\left\vert #1 
    \right\vert\kern-0.25ex\right\vert\kern-0.25ex\right\vert}}

\newcommandx{\CPE}[3][1=]{{\mathbb E}_{#1}\left[ \left. #2 \middle \vert #3 \right. \right]} 
\newcommandx{\CPVar}[3][1=]{\mathrm{Var}^{#3}_{#1}\left\{ #2 \right\}}
\newcommand{\CPP}[3][]
{\ifthenelse{\equal{#1}{}}{{\mathbb P}\left(\left. #2 \, \right| #3 \right)}{{\mathbb P}_{#1}\left(\left. #2 \, \right | #3 \right)}}

\def \rd {\rset^{d}}

\renewcommand{\epsilon}{\varepsilon}

\def\loss{\ell}

\def\ite{\boldsymbol{w} }
\def\ww{\ite} 
\def\ws{\ite^{\star}}
\newcommand{\iter}[3]{\ite_{#1, #2}^{#3}} 

\def\ites{\boldsymbol{v}} 

\def\itePRav{\overline{\overline{\ite}}^C}

\def \rd {\rset^{d}}

\def\Nt{N^{t}} 
\def\Ntp{N^{t'}} 

\newcommand{\un}[1]{[#1]}

\def\gg{\boldsymbol{g}}

\newcommand\inner[2]{\big\langle #1, #2 \big\rangle}
\newcommand{\e}[1]{\mathbb{E}\left[#1\right]}
\newcommand{\norm}[1]{\left\lVert#1\right\rVert}

\def\siginf{\sigma ^2_\infty}

\def\cm{\checkmark}

\def\CA{L}
\def\Lone{\sigma^{2}}

\def\forptk{p\in \un{P},t\in \un{C}, k\in \un{\Nt}}
\def\fortk{t\in \un{C}, k\in \un{\Nt}}
\begin{document}

\maketitle
	
\begin{abstract}	
	Synchronous mini-batch SGD is state-of-the-art for large-scale distributed machine learning. However, in practice, its convergence is bottlenecked by slow communication rounds between worker nodes. A natural solution to reduce communication is to use the \emph{``local-SGD''}  model in which the workers train their model independently and synchronize every once in a while. This algorithm improves the computation-communication trade-off but its convergence is not understood very well. We propose a non-asymptotic error analysis, which enables comparison to \emph{one-shot averaging} i.e., a single communication round among independent workers, and \emph{mini-batch averaging} i.e., communicating at every step. We also provide adaptive lower bounds on the communication frequency for large step-sizes ($ t^{-\alpha} $, $ \alpha\in (1/2 , 1 ) $) and show that \emph{Local-SGD} reduces communication by a factor of $O\Big(\frac{\sqrt{T}}{P^{3/2}}\Big)$, with $T$ the total number of gradients and $P$ machines.     
\end{abstract}

\addtocontents{toc}{\protect\setcounter{tocdepth}{0}}

\section{Introduction}
We consider the minimization of an objective function which is accessible through unbiased estimates of its gradients. This problem has received attention from various communities over the last fifty years in optimization, stochastic approximation, and machine learning~\citep{Pol_Jud_1992,Rup_1988,Fab_1968,Nes_Via_2008,Nem_Jud_Lan_2009,Sha_Sha_Sre_2009,Zha_2004}. The most widely used algorithms are stochastic gradient descent (SGD), a.k.a. Robbins-Monro algorithm~\citep{Rob_Mon_1951}, and some of its modifications based on averaging of the iterates~\citep{Pol_Jud_1992,Rup_1988,Sha_Zha_2013}. For a convex differentiable function $ F :\rd \to \mathbb{R} $, SGD  iteratively updates an estimator $ (\ites_t)_{t\geq 0} $ for any $ t\geq 1 $
\begin{equation}\label{eq:defSGD}
	\ites_{t} = \ites_{t-1} -\eta_t \gg_t(\ites_{t-1}), 
\end{equation}
where $ (\eta_t)_{t\geq 0}  $ is a deterministic sequence of positive scalars, referred to as the \emph{learning rate} and $ \gg_t(\ites_{t-1}) $ is an oracle on the gradient of the function $ F $ at $ \ites_{t-1} $. We focus on objective functions that are both smooth and strongly convex \citep{Bac_Mou_2011}. While these assumptions might be restrictive in practice, they enable to provide a tight analysis of the error of SGD. In such a setting, two types of proofs have been used traditionally. On one hand, \emph{Lyapunov}-type proofs rely on controlling  the expected squared distance to the optimal point \citep{Zha_Zha_2015}. Such analysis suggests using \emph{small} decaying steps, inversely proportional to the number of iterations ($t^{-1}$). On the other hand, studying the recursion as a stochastic process~\citep{Pol_Jud_1992} enables to better capture the reduction of the noise through averaging.  It results in optimal convergence rates for larger steps, typically scaling as $ t^{-\alpha} $, $ \alpha\in (1/2 , 1 ) $~\citep{Bac_Mou_2011}.

\begin{figure}        
	\includegraphics[angle=90, width=\linewidth]{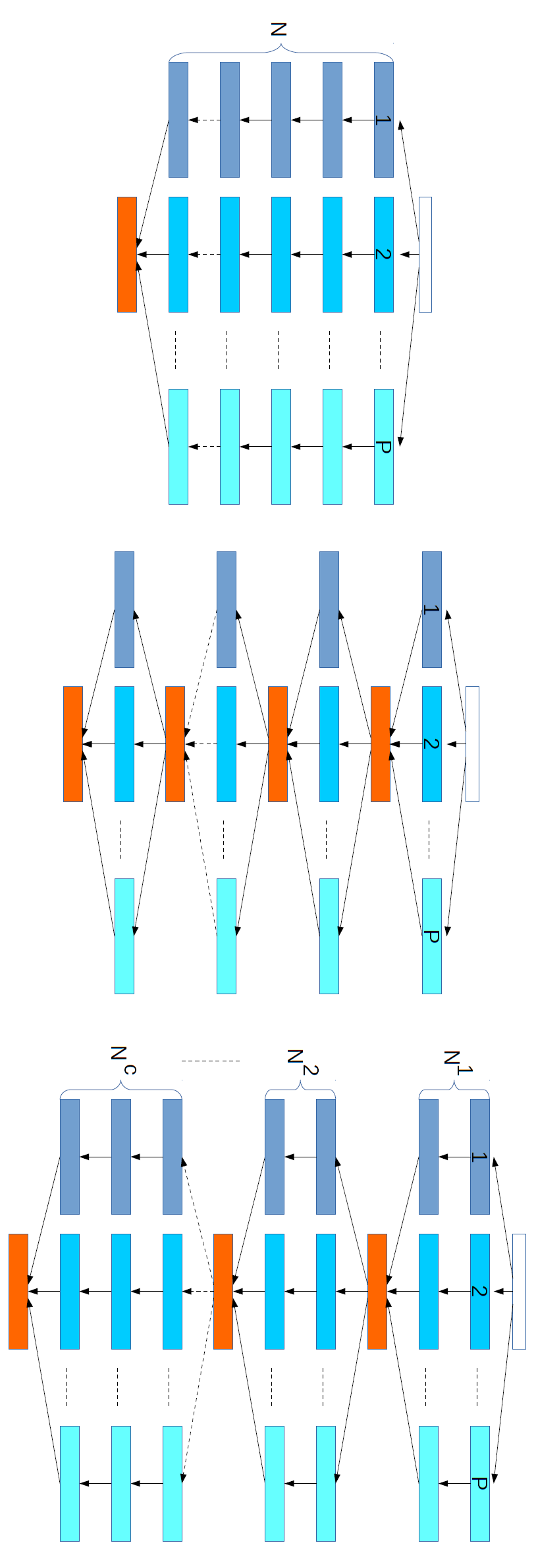}
	\caption{Schematic representation of one-shot averaging (left), mini-batch averaging (middle) and local-SGD (right). Vertical threads correspond to  machines and orange boxes to communication rounds.}\label{fig:2}
	\vspace{-1.5em}
\end{figure}

Over the past decade, the amount of available data has steadily increased: to adapt SGD to such situations, it has become necessary to \emph{distribute} the workload between several machines, also referred to as \emph{workers} \citep{Del_Ben_2007, Zin_2010, Rec_Re_2011}. For SGD, two extreme approaches have received attention: 1) workers  run SGD independently and at the end aggregate their results, called \emph{one-shot averaging} (\textbf{OSA}) \citep{Zin_2010,God_Saa_2017} or \emph{parameter mixing}, and 2) \emph{mini-batch averaging} (\textbf{MBA})~\citep{Dek_2012,Tak_2013,Li_2014,Goy_2017,Jai_Kak_Kid_2016}, where workers communicate after every iteration: all gradients are thus computed at the \emph{same} support point (iterate)
and the algorithm is equivalent to using mini-batches of size $ P $, with $ P $ the number of workers. While OSA requires only a single communication step, it typically does not perform very well in practice~\citep{Zha_DeS_Re_2016}. At the other extreme, MBA performs better in practice, but the number of communications equals the number of steps, which is a major burden, as communication is highly time consuming~\citep{Zhazip_2016}. To optimize this computation-communication-convergence trade-off, we consider the \emph{Local-SGD} framework: $ P $ workers run SGD iterations in parallel and communicate  periodically. This framework encompasses \emph{one-shot averaging} and \emph{mini-batch averaging} as special cases (see \Cref{fig:2}).

We make the following contributions: \\
1) We provide the  first non-asymptotic analysis for local-SGD with large step sizes (typically scaling as $ t^{-\alpha} $, for $ \alpha\in (1/2; 1) $), in both on-line and finite horizon settings. Our assumptions encompass the ubiquitous \emph{least-squares regression} and \emph{logistic regression}. \\
2) Our comparison of the two extreme cases, OSA and MBA, underlines the communication trade-offs. While both of these algorithms are  asymptotically equivalent for a fixed number of machines, mini-batch theoretically outperforms one-shot averaging when we consider the precise bias-variance split. In the regime where both the number of machines and  gradients grow simultaneously we  show that mini-batch SGD outperforms one-shot averaging. \\
3) Under three different sets of assumptions, we quantify the \emph{frequency of communication} necessary for Local SGD to be optimal (i.e., as good as mini-batch). Precisely, we show that the communication frequency can be reduced by as much as $O\Big(\frac{\sqrt{T}}{P^{3/2}}\Big)$, with $T$ gradients and $P$ workers. Moreover, our bounds suggest an adaptive communication frequency for logistic regression, which depending on the expected distance to the optimal point (a phenomenon observed by \cite{Zha_DeS_Re_2016}).\\
4) We support our analysis by experiments illustrating the behavior of the algorithms. 

The paper is organized as follows:  in \Cref{sec:setting}, we introduce the general setting, notations and algorithms, then in  \Cref{sec:related}, we describe the related literature. Next, in \Cref{subsec:asspt}, we describe assumptions made on the objective function. In \Cref{sec:results}, we provide our main results, their interpretation, consequence and comparison with other results. Results in the on-line setting and experiments are presented in the~\Cref{subsec:onlinesetting} and \Cref{sec:experiments}.

\section{Algorithms and setting}
\label{sec:setting}

We first introduce a couple of notations.  We consider the finite
dimensional Euclidean space $ \rset^{d} $ embedded with its canonical
inner product $ \langle\cdot, \cdot\rangle $. For any integer $ \ell \in \nset^{*} $, we denote by $ \un{\ell} $ the set  $\{ 1, \ldots, \ell\} $.
We consider a strongly-convex differentiable function $ F:\rd \to \mathbb{R}$.  We denote by $ \ws $ the point such that $ \ws= \arg\!\min_{\ite}\ F(\ite) $. With only one machine, \emph{Serial-SGD} performs a sequence of updates according to Equation~\eqref{eq:defSGD}. 
In the next section, we describe  Local-SGD, the subject of this study.

\subsection{Local-SGD algorithm}

We  consider $ P $ machines, each of them running SGD.
Periodically, workers  aggregate (i.e., average) their models and restart from the resulting model. We denote by $ C $ the number of communication steps. We define a \emph{phase} as the time between two communication rounds. At \emph{phase} $ t\in \un{C} $, for any worker $ p\in \un{P} $, we perform~$ \Nt $ \emph{local steps} of SGD. 
Iterations are thus naturally indexed by $ (t, k) \in \un{C}\times \un{\Nt} $. We consider the lexicographic order $  \preccurlyeq $ on such pairs, which matches the order in which  iterations are processed. Note that we assume the number of local steps is the same over all machines~$ p $. While this assumption can be relaxed in practice, is facilitates our proof technique and notation.  At any $ k\in \un{\Nt} $, we denote by $ \iter{p}{k}{t} $ the model proposed by worker $ p $, at phase~$ t  $, after $ k $ local iterations. All machines initially start from the same point $ \ite_0 $, that is for any $p\in \un{P}  $, $ \iter{p}{0}{1} = \ite_0$.
The update rule is thus the following, for any $ p\in \un{P} , t\in \un{C},  k\in \un{\Nt} $: \begin{equation}\label{eq:SGDmachineP}
	\iter{p}{k}{t}= \iter{p}{k-1}{t} - \eta_k^{t} g_{p,k}^{t} ( \iter{p}{k-1}{t}  ).
\end{equation}
Aggregation steps consist in averaging the final local iterates of a phase: for any $ t\in \un{C} $,	$	\hat\ite^t = \frac{1}{P} \sum_{p=1}^{P} \iter{p}{\Nt}{t}.$
At phase $ t+1 $, every worker $ p\in \un{P} $  restarts from the averaged model:  $ \iter{p}{0}{t+1}:=\hat \ite^t$.
Eventually, we are interested in controlling the excess risk of the Polyak-Ruppert averaged iterate: \begin{align*}
	&\itePRav= \frac{1}{\sum_{t=1}^{C} \Nt} \sum_{t=1}^{C} \Nt \overline{\ite}^{t} = \frac{1}{ P \sum_{t=1}^{C} \Nt} \sum_{t=1}^{C} \sum_{p=1}^{P}  \sum_{k=1}^{\Nt}  \iter{p}{k}{t}, 
\end{align*}
{with }   $ \overline{\ite}^{t} =\frac{1}{P \Nt} \sum_{k=1}^{\Nt}  \sum_{p=1}^{P} \iter{p}{k}{t} $. We use the notation $ \overline{\ite} $ to underline the fact that iterates are averaged over one phase and  $ \overline{\overline{\ite} }$  when averaging is made over all iterations. All averaged iterates can be computed on-line.

The algorithm, called \emph{local-SGD}, is thus parameterized by the number of machines $ P $, communication steps $ C $,  local iterations $ (\Nt)_{ t\in \un{C}}  $, the starting point $ \ite_0 $, the learning rate $ (\eta^t_k )_{ (t, k) \in \un{C}\times \un{\Nt}}$, and the first order oracle on the gradient. Pseudo-code of the algorithm is given in Table~\ref{alg:locSGDmain}.

\textbf{Link with classical algorithms.}
Special cases of Local-SGD correspond to \emph{one-shot averaging} or \emph{mini-batch averaging}, as summarized in Table~\ref{tab:link}. More precisely, for a total number of gradients $ T $, with $ P $ workers,  $ C=T/P $ communication rounds,  and $ (\Nt)_{ t\in \un{C}} = (1,.\dots, 1) $, we realize an instance of P-mini-batch averaging (P-MBA). On the other hand,  with $ P $ workers,  $ C=1 $ communication,  and $ (N^1)=T/P $, we realize an instance of one shot-averaging.  Our goal is to get general convergence bounds for Local-SGD that recover classical bounds for both these settings when we choose the correct parameters. While comparing to Serial-SGD (which is also a particular case of the algorithm), would also be interesting,  we focus here  on the comparison between Local-SGD, \emph{one-shot averaging} and \emph{mini-batch averaging}. Indeed, the step size is generally increased for mini-batch with respect to Serial SGD, and the running efficiency of algorithms is harder to compare: we only focus on different algorithms that use the \emph{same number of machines}. 

\begin{table}
	\centering
\begin{tabular}{cc}
	\begin{minipage}{0.48\linewidth}
\resizebox{\linewidth}{!}{%
			\begin{tabular}{lcccc}
			\hline
			Algo.	&Work. &  Com. & Phases  & $ T $\\ \hline 
			Local  &  $ P $ & $ C $ & $ (N^1\dots  N^C) $ &  $ P\sum_{t=1}^{C} \Nt $\\
			Serial& 1 & - & $ (N)  $ & $ N $ \\
			P-MBA &  $ P $ &$  C $ &  $ (1, \dots , 1) $  & $ PC $  \\
			OSA & $ P  $ &$ 1 $ &  $ (N^1) $  & $  N^1 P$  \\
			\hline
		\end{tabular}}
		\vspace{0.5em}
		\caption{One shot averaging and mini-batch SGD can be seen as particular instances of our algorithm, depending on the number of Workers, Communication Rounds,  Phase lengths and total number of gradients.\label{tab:link}}	
	\end{minipage}%
&
	\begin{minipage}{0.48\linewidth}
		\begin{algorithm}[H]
			\KwIn{$F:\R^d\rightarrow\R$}
			$\hat{\ww}^0=\ww^0 \gets \textbf{Initialize}$\\
			\For{$t=1,2,\dots, C $}{
				\ForPar{$ i=1,2,\dots, P $}{
					$\ww_{i,0}^t\gets \hat{\ww}^{t-1}$\\
					\For{$ k=0,1,\dots, N^t$}{
						$\gg_{i,k}^t(\ww_{i,k-1}^t)\gets \textbf{SFO}(F,\ww_{i,k-1}^t)$
						$\ww_{i,k}^t\gets\ww_{i,k-1}^t-\eta_{k}^t\gg_{i,k}^t(\ww_{i,k-1}^t)$
					}
					$\overline{\ww}_i^t \gets \frac{1}{N_t}\sum_{k=1}^{N_t}\ww_{i,k}^t$    
				}
				$\overline{\ww}^t \gets \frac{1}{P}\sum_{i=1}^{P}\overline{\ww}_i^t$; 
				$\hat{\ww}^t \gets \frac{1}{p}\sum_{i=1}^{P}\ww_{i,N_t}^t$
			}
			\KwOut{$\overline{\overline{\ww}}^T=\frac{1}{C}\sum_{t=1}^{C}\overline{\ww}^{t}\in\R^d$}
		\end{algorithm}
	\vspace{-1em}
			\caption{Pseudo code for Local-SGD}
		\label{alg:locSGDmain}
	\end{minipage}
\end{tabular}	
\end{table}

\subsection{Related Work}
\label{sec:related}

\textbf{On Stochastic Gradient Descent.}
Bounds on the excess risk of  SGD for convex functions have been widely studied: most proofs rely on controlling the decay of the mean squared distance $ \E[\mynorm[2]{\ites_t-\ite^\star}] $, which results in an upper bound on the mean excess of risk $ \E[F(\bar \ites_t) - F(\ite^\star)] $~\citep{Lac_Sch_Bac_2012,Rak_Sha_Sri_2011}. This upper bound is composed of a  ``bias'' term that depends on the initial condition, and a ``variance'' term that involves either an upper bound on the \emph{norm} of the noisy gradient (in the non-smooth case), or an upper bound on the \emph{variance} of the noisy gradient in the smooth case~\citep{Zha_Zha_2015}. In the strongly convex case such an approach advocates for the use of \emph{small} step sizes, scaling as $ (\mu t) ^{-1}$. However, in practice, this is not a very satisfying result, as the constant $ \mu $ is typically unknown, and convergence is very sensitive to ill-conditioning. On the other hand, in the smooth and strongly-convex case, the classical analysis by~\citet{Pol_Jud_1992}, relies on an explicit decomposition of the stochastic process $ (\bar \ites_t-\ite^\star)_{t\geq 1}$: the effect of averaging on the noise term is better taken into account, and this analysis thus suggests to use larger steps, and results in the optimal rate for $ \eta_t \varpropto t^{-\alpha} $, with $ \alpha \in (0;1) $. This type of analysis has been successfully used recently~\citep{Bac_Mou_2011,Die_Dur_2018,God_Saa_2017, Gad_Pan_2107}. 

For quadratic functions, larger steps can be used, as pointed by \cite{Bac_Mou_2013}. Indeed, even with \emph{non-decaying} step size, the averaged process converges to the optimal point. Several studies focus on understanding properties of SGD for quadratic functions: a detailed non-asymptotic analysis is provided by  \cite{Def_Bac_2015}, acceleration under the additive noise oracle (see Assumption~\Cref{hyp:addit_noise} below) is studied by \cite{Die_Fla_Bac_2016} (without this assumption by \cite{Jai_Kak_Kid_2017}), and~\cite{Jai_Kak_Kid_2016} analyze the effects of mini-batch and tail averaging.

\textbf{One shot averaging.} In this approach, the $ P $-independent workers compute several steps of stochastic gradient descent, and a unique communication step is used to average the different models \citep{mcdonald2009efficient,mcdonald2010distributed,Zin_2010}.  \cite{Zin_2010} show a reduction of the variance when multiple workers are used, but neither consider the Polyak-Ruppert averaged iterate as the final output, nor provide non-asymptotic rates. \cite{zhang2012communication} provide the first non-asymptotic results for OSA but their dependence on constants (like strong convexity constant $\mu$, moment bounds, etc.) is worse; as well as their single machine convergence bound \citep{rakhlin2012making} is not truly non-asymptotic (like for e.g., \cite{Bac_Mou_2011}). More importantly, their results hold only for small learning rates like $\frac{c}{\mu t}$. \cite{rosenblatt2016optimality} have also discussed the asymptotic equivalence of OSA with vanilla-SGD by providing an analysis up to the second order terms. Further, \cite{Jai_Kak_Kid_2016} have provided non-asymptotic results for least-square regression using similar Polyak-Juditsky analysis of the stochastic process, while our results apply to more general problems. Their approach encompasses one shot averaging and the effect of tail averaging, that we do not consider here. Recently,~\cite{God_Saa_2017} proposed an approach similar to ours (but only for one shot averaging). However, their result relies on an asymptotic bound, namely $ \E[\mynorm[2]{\ite_t-\ws}] \le C_1 \eta_t$ (as in \cite{rakhlin2012making}), while our analysis is purely non-asymptotic and we also improve the upper bound on the noise term which results from the analysis. 

\textbf{Mini-batch averaging.} Mini-batch averaging has been studied by  \cite{Dek_2012,Tak_2013}. These papers show an improvement in the variance of the process, and make comparisons to SGD. It has been found that increasing the mini-batch size often leads to increasing generalization errors, which limits their distributivity \citep{li2014efficient}. \cite{Jai_Kak_Kid_2016} have provided upper bounds on learning-rate and mini-batch size  for optimal performance. Recently, large mini-batches have been leveraged successfully in deep learning as in~\citep{keskar2016large,you2017large,goyal2017imagenet} by properly tuning learning rates, etc.  

\textbf{Local-SGD.}  \cite{Zha_DeS_Re_2016} empirically show that local SGD performs well. They also provide a theoretical guarantee on the variance of the process, however, they assume the variance of the estimated gradients to be uniformly upper bounded (Assumption \Cref{hyp:addit_noise} below). Such an assumption is  restrictive in practice, for example it is not satisfied for least squares regression. In a simultaneous work, \cite{stich2018local} has  provided an analysis for local-SGD. The limitation with their analysis is that they also assume bounded gradients and use a small step size scaling as $\frac{c}{\mu t}$. More importantly, their analysis doesn't extend to the extreme case of one-shot averaging like ours. \cite{Tao2018local} have experimentally shown that Local-SGD is better than the synchronous mini-batch techniques, in terms of overcoming the large communication bottleneck. Recently, \cite{yu_2018} have given convergence rates for the non-convex synchronous and a stale synchronous settings.   

We have summarized the major limitations of some of these analyses in \Cref{tab:limit}, given in \Cref{app:relatedwork}. Our motivation is to get away with some of these restrictive assumptions, and provide tight upper bounds for the above three averaging schemes.
In the following section, we present the set of assumptions under which our analysis is conducted.

\subsection{Assumptions}\label{subsec:asspt}

We first make the following  classical assumptions on the objective function $F:\rset^d \to \rset$. In the following, we use different subsets of these assumptions:
\begin{assumption}[\textbf{Strong convexity}]
	\label{hyp:strong_convex}
	The function $F$ is strongly-convex with convexity constant $\mu>0$.
	
\end{assumption}

\begin{assumption}[\textbf{Smoothness and regularity}]
	\label{hyp:regularity}
	The function $F$ is three times continuously differentiable with second and third uniformly bounded derivatives: $\sup_{\ite\in \rset^d} \opnorm{F^{(2)}(\ite)} < L$, and $\sup_{\ite\in \rset^d} \opnorm{F^{(3)}(\ite)} < M$. Especially  $F$ is $L$-smooth.
\end{assumption}
\begin{assumptionQ}[\textbf{Quadratic function}]\label{hyp:quad}
	There exists a positive definite matrix $\Sigma \in \R^{d\times d}$, such that  the function $F$ is the quadratic function $ \ite \mapsto \normLigne{\Sigma^{1/2} (\ite -\ws)}^2/2$, 
\end{assumptionQ}
If \Cref{hyp:quad} is satisfied, then Assumptions~\Cref{hyp:strong_convex},~\Cref{hyp:regularity} are satisfied, and $ L$ and $ \mu $ are respectively the largest and smallest eigenvalues of $ \Sigma $.  At any iteration $ (t, k) \in \un{C}\times \un{\Nt}$, any machine can query an unbiased estimator of the gradient $ g_{p,k}^{t}(\ite) $ at a point $ \ite $. Formally, we make the following assumption:
\begin{assumption}[\textbf{Oracle on the gradient}]\label{ass:def_filt}
	{There exists a filtration $(\h{t}{k})_ {(t, k) \in \un{C}\times \un{\Nt}}$  on some probability space $(\Omega, \mathcal{F}, \mathbb{P})$ such that  for any  $(t, k) \in \un{C}\times \un{\Nt}$ and $\ite \in \rset^{d}$, $g_{p,k+1}^{t}(\ite)$ is a  $\h{t}{k+1}$-measurable random variable and $\expe{g_{p,k+1}^{t}(\ite)|\h{t}{k}}=F'(\ite)$.
		In addition, we assume the functions $(g^t_{p,k}) _{(t, k) \in \un{C}\times \un{\Nt}}  $ to be independent and identically distributed (\iid) random fields.}
\end{assumption}
A filtration is an increasing (i.e., for all $(t,k)\preccurlyeq(t',k')$, $\h{t}{k} \subset \h{t'}{k'}$), sequence of $ \sigma$-algebras. \Cref{ass:def_filt}  expresses that  we have access to an \iid~sequence $(g^t_{p,k})_{(t, k) \in \un{C}\times \un{\Nt}}$ of unbiased estimators of $F'$.
Remark that with such notations, for any $ t\in \un{C}, k\in \un{\Nt}, p\in \un{P} $,  $ \iter{p}{k}{t} $ is $ \h{t}{k}  $-measurable. 
In \Cref{prop:conv_quad_Simple}
, we make the additional, stronger assumption that the variance of gradient estimates is uniformly upper bounded, a standard assumption in the SGD literature, see e.g.~\cite{Zha_DeS_Re_2016}:
\begin{assumption}[\textbf{Uniformly bounded variance}]\label{hyp:addit_noise}
	We assume the variance of the error, $ \expeLigne{\normLigne[2]{g^t_{p,k} (\iter{p}{k}{t})- F'(\iter{p}{k}{t}) }} $ to be  uniformly upper bounded by $ \siginf $, a constant which does not depend on the iteration.
\end{assumption}
Assumption \Cref{hyp:addit_noise} is for example true if the sequence of random vectors~$ ( g^t_{p,k+1} (\iter{p}{k}{t})- F'(\iter{p}{k}{t}) )_{t\in \un{C}, k\in \un{\Nt}, p\in \un{P}}$ is  \iid. This setting is referred to as the semi-stochastic setting in~\cite{Die_Fla_Bac_2016}.

We also consider the following conditions on the regularity of the gradients, for $ p\geq 2 $:
\begin{assumption}[\textbf{Cocoercivity of the random gradients}]\label{ass:lip_noisy_gradient_AS} 
	For any $ t\in \un{C}$, $k\in \un{\Nt}$, $p\in \un{P}$, $g^t_{p,k}$ is almost surely $L$-co-coercive (with the same constant as in \Cref{hyp:regularity}): that is, for any $ \ite_1, \ite_2\in \R^{d} $, $ L \ps{g^t_{p,k}(\ite_1)-g^t_{p,k}(\ite_2)}{\ite_1-\ite_2} \geq \normLigne[2]{g^t_{p,k}(\ite_1)-g^t_{p,k}(\ite_2)} $.
\end{assumption}
Almost sure $L$-co-coercivity~\citep{Zhu_Mar_1995} is for example
satisfied if for any $(p,k)\in \un{P}\times \un{\Nt} $, there exist a random function
$ f^t_{p,k}$ such that $ g^t_{p,k} = (f^t_{p,k})' $ and which is \as~convex and
$ L$-smooth.  Finally, we assume the fourth order moment of the random gradients at $ \ws $ to be well defined:
\begin{assumption}[\textbf{Finite variance at the optimal point}]\label{hyp:variancebound}
	There exists $ \sigma \geq 0 $, such that  for any  ${t\in \un{C}, k\in \un{\Nt}, p\in \un{P}}$, 
	$\PE[\normLigne{g^t_{p,k} (\ws)}^4] \le \sigma^{4}$.
\end{assumption}
It must be noted that \Cref{hyp:variancebound} is a much weaker assumption than \Cref{hyp:addit_noise}, for e.g., least-square regression satisfies former but not latter. Most of these assumptions are classical in machine learning. SGD for least squares regression  satisfies \Cref{hyp:quad}, \Cref{ass:def_filt}, \Cref{ass:lip_noisy_gradient_AS} and \Cref{hyp:variancebound}. On the other hand, SGD for logistic regression satisfies \Cref{hyp:strong_convex}, {\Cref{hyp:regularity}}, \Cref{ass:def_filt} and {\Cref{hyp:addit_noise}}. Our main result~\Cref{th:LocalSGDconv}  (lower bounding the frequency of communications)  applies to both these sets of assumptions. 
In~\Cref{ex:iid_observation_learning} we  further detail how these assumptions apply in machine learning.

\textbf{Learning rate.}
We consider two different types of learning rates:\\
1) in the \emph{finite horizon} case, the step size $(\eta^t_k )_{ (t, k) \in \un{C}\times \un{\Nt}}$ is a constant $ \eta $, that can depend on the number of iterations eventually performed by the algorithm; 
2) in the \emph{on-line} case, the sequence of step size is a subsequence of a universal sequence $( \tilde{\eta}_\ell)_{\ell \geq 0} $.  Moreover, in our analysis, when using decaying learning rate, the step size only depends on the number of iterations processed in the past: $ \eta^t_k  = \tilde{\eta}_{\lbrace\sum_{t'=1}^{t-1} \Ntp +k \rbrace} $. Especially, the step size at iteration $ (t,k) $ does not depend on the machine. 

Though both of these approaches are often considered to be nearly equivalent \citep{Bac_2014, Die_Bac_2015}, fundamental differences exist in their convergence properties. The \emph{on-line} case is harder to analyze, but ultimately provides a better convergence rate.  However as the behavior is easier to interpret in the finite horizon case, we postpone results for on-line setting to~\Cref{subsec:onlinesetting}.

Moreover, we always assume 
that for any $\fortk  $, the learning rate satisfies $ 2 \eta_k^t L \le 1 $. 	In the following section, we present our main results. 

\section{Main Results}

\label{sec:results}
\textbf{Sketch of the proof.} We follow the approach by Polyak and Juditsky, which relies on the following decomposition: for any $ p\in \un{P}, t\in \un{C}, k\in \un{\Nt}$,  Equation~\eqref{eq:SGDmachineP} is trivially equivalent to:
\begin{align}
		\eta_k^{t}  F''(\ws) (\iter{p}{k-1}{t}-\ws) & =   \iter{p}{k-1}{t}- \iter{p}{k}{t}  - \eta_k^{t} \left [g_{p,k}^{t} (\iter{p}{k-1}{t})- F'  (\iter{p}{k-1}{t})\right ] \nonumber\\
	& -  \eta_k^{t}\left [ F'  (\iter{p}{k-1}{t}) - F''(\ws) (\iter{p}{k-1}{t}-\ws) \right ]. \nonumber
\end{align}
We have used  a first order Taylor expansion around the optimal value $ \ws $ of the gradient. Thus, using the definition of $ \itePRav $:
\begin{align}
	F''(\ws) \left (\itePRav-\ws\right ) &=    \frac{1}{P\sum_{t=1}^{C}\Nt}   \sum_{t=1}^{C} \sum_{p=1}^{P}  \sum_{k=1}^{\Nt}        \bigg     (    \frac{\iter{p}{k-1}{t}- \iter{p}{k}{t}}{\eta_k^{t}}  -  \left [g_{p,k}^{t} (\iter{p}{k-1}{t})- F'  (\iter{p}{k-1}{t})\right ] \nonumber\\
	& - \left [ F'  (\iter{p}{k-1}{t}) - F''(\ws) (\iter{p}{k-1}{t}-\ws) \right ] \bigg).  \label{eq:dec_Pol_Jud}
\end{align}
In other words, the error can be decomposed into three terms: the first one mainly depends on the \emph{initial condition}, the second one is a \emph{noise term}: it is the mean of centered random variables  (as $ \expeLigne{g_{p,k}^{t} (\iter{p}{k-1}{t})- F'  (\iter{p}{k-1}{t})}=0 $), and the third is a \emph{residual term} that accounts for the fact that the function is not quadratic (if $ F $ is quadratic, then $  F'  (\iter{p}{k-1}{t}) - F''(\ws) (\iter{p}{k-1}{t}-\ws)=0 $).

\textbf{Controlling different terms in Equation~\eqref{eq:dec_Pol_Jud}.} The variance of the noise $ g_{p,k}^{t}(\iter{p}{k-1}{t})- F'  (\iter{p}{k-1}{t}) $ and the residual term both directly depend on the distance $  \normLigne{\iter{p}{k-1}{t}- \ws}^{2}$. The proof is thus composed of two aspects: (1) we first provide a tight control for this quantity, with or without communication: in the following propositions, this corresponds to an upper bound on $ \expeLigne{\normLigne{\iter{p}{k}{t}- \ws}^{2}} $ \footnote{more precisely, on  $\expeLigne{\normLigne{\hat \ite^{t} -\ws}^2}$ and $\expeLigne{\normLigne{ \iter{p}{k}{1} -\ws}^2}$ for MBA and OSA respectively.}, (2)  we provide the subsequent upper bound on $ \expeLigne{\normLigne{F''(\ws)(\itePRav- \ws)}^{2}}  $.


We first compare the results for \textit{Mini-batch averaging} and \textit{One-shot averaging} for \emph{finite horizon (FH)}  setting, and then provide these results for local-SGD. 

\subsection{Results for MBA and OSA, FH setting}	
\label{sec:OSA_MBA}

First we assume the step size $ \eta_k^{t} $ to be a constant $ \eta $ at every iteration for any $ (t,k)\in \un{C}\times \un{\Nt} $. Our first contribution is to provide \emph{non-asymptotic} convergence rates for \emph{mini-batch SGD} and \emph{one shot averaging}, that allow a simple comparison. For the benefit of presentation, we define following quantities:
\begin{align*}
	Q_{bias} &= 1 + \frac{M^2 \eta }{\mu}\norm{\ww^0-\ws}^2 +\frac{L^2\eta}{\mu P},
\  \	Q_{1,var}(X) =  \frac{L^2\eta}{\mu}+ \frac{P}{X \eta\mu} , \ \  Q_{2, var}(X)= \frac{M^2 X P \eta^2 \sigma^2}{\mu^2} .
\end{align*}

We have the following result for mini-batch averaging:

\begin{proposition}[Mini-batch Averaging]\label{prop:conver_Mini_Batch}
	Under Assumptions~\Cref{hyp:strong_convex},~\Cref{hyp:regularity},~\Cref{ass:def_filt},~\Cref{ass:lip_noisy_gradient_AS},~\Cref{hyp:variancebound}, we have the following bound for mini-batch SGD: for any $ t\in \un{C} $, 
	\begin{align}
&		\expe{\mynorm[2]{\hat \ite^{t} -\ws}} \le (1-\eta \mu)^{t}  \mynorm[2]{\ite_0-\ws}+ \frac{2\sigma^{2} \eta}{P} \frac{1 - (1-\eta \mu)^{t}}{\mu},\label{eq:prop11} \\
		&		\expe{\mynorm[2]{F''(\ws)(\itePRav -\ws)}} \precsim  \frac{\norm{\ww^0-\ww^\star}^2}{\eta^2C^2 }Q_{bias}  +\frac{\sigma^2}{T}\Big(1+\frac{Q_{1,var}(C)}{P}+\frac{Q_{2,var}(C)}{P^2}\Big)\label{eq:prop12}.
	\end{align}
\end{proposition}

The notation $ \precsim $ denotes inequality up to an absolute constant. Recall that for mini-batch, the total number of gradients processed is  $ T= PC $.

On the other hand, we also have the following result for one-shot averaging:
\begin{proposition}[One-shot Averaging]\label{prop:conver_One_Shot}
	Under Assumptions~\Cref{hyp:strong_convex},~\Cref{hyp:regularity},~\Cref{ass:def_filt},~\Cref{ass:lip_noisy_gradient_AS},~\Cref{hyp:variancebound}, we have the following bound for one shot averaging: $ p\in \un{P}, t=1, k \in \un{N} $,
	\begin{align}
&		\expe{\mynorm[2]{ \iter{p}{k}{1} -\ws}} \le (1-\eta \mu)^{k}  \mynorm[2]{\ite_0-\ws}+ 2 {\sigma^{2} \eta} \frac{1 - (1-\eta \mu)^{k}}{\mu}, \label{eq:prop21} \\
		&			\expe{\mynorm[2]{F''(\ws)(\itePRav -\ws)}}  \precsim  	\frac{\norm{\ww^0-\ww^\star}^2}{\eta^2N^2}Q_{bias} + \frac{\sigma^2}{T}\big(1+Q_{1,var}(N)+Q_{2,var}(N)\big) \label{eq:prop22}.
	\end{align}
\end{proposition}

Note that for one-shot averaging, the total number of gradients used is $ T= PN $.

\textbf{Interpretation, fixed $ P $.}	Using mini-batch naturally reduces the variance of the process $ (\iter{p}{k}{t}) _{\forptk}$.  \Cref{eq:prop11,eq:prop21} show that the speed at which the initial condition is forgotten remains the same, but that the variance of the local process is reduced by a factor $ P $.

\Cref{eq:prop12,eq:prop22} show that the convergence depends on an \emph{initial condition} term and a \emph{variance term}. For a fixed number of machines $ P $, and a step size scaling  as $ \eta = X^{-\alpha} $, $ 0.5< \alpha<1 $,  $ X \in \{N,C\} $, the speed at which the \emph{initial condition} is forgotten is asymptotically dictated by  $ Q_{bias}/(\eta X)^{2} $ where $ X \in \{N,C\} $, for  \emph{both algorithms} (if we use the same number of gradients for both algorithms, naturally, $ N=C $.)
As for the variance term, it scales as $  \sigma^2 T^{-1}  $ as $ T\to \infty $ , as the remaining terms $ Q_{var}(X) $ asymptotically vanish for $  \eta = X^{-\alpha}  $. 
It reduces with the total number $ T $ of gradients used in the process. Interestingly, this term is \emph{the same} for the two extreme cases (MBA and OSA): it does not depend on the number of communication rounds. This phenomenon is often described as \emph{``the noise is the noise and SGD doesn't care''} (for asynchronous SGD, \citep{duchi2015asynchronous}). Though we recover this asymptotic equivalence here, our belief is that this asymptotic point of view is typically misleading as the asymptotic regime is not always reached, and the residual terms do then  matter. 

Indeed, the lower order terms do have a dependence on the number of communication rounds: when the number of communications increases, the overall effect of the noise is reduced. More precisely, since $Q_{var}(N) = Q_{var}(C)$ the remaining terms are respectively $ P $ or $ P^2 $ times smaller for  mini-batch. This provides a theoretical explanation of why mini-batch SGD outperforms one shot averaging in practice. It also highlights the weakness of an asymptotic analysis: the dominant term might be equivalent, without reflecting the actual behavior of the algorithm. Disregarding communication aspects, mini-batch SGD is  in that sense \emph{optimal}.

Note that for quadratic functions, $ Q_{2, var} =0 $ as $ M=0 $. The conditions on the step size can thus be relaxed, and the asymptotic rates described above would be valid for any step size satisfying $ \eta\le \mu $ \citep{Jai_Kak_Kid_2016}.

Extension to the on-line setting, eventually leading to a better convergence rate, is given in \Cref{prop:conver_Mini_Batch_DSS} in \Cref{subsec:onlinesetting}.

\textbf{Interpretation,  $ P, T \to \infty$.}	When both the total number of gradients used $ T $ and the number of machines $ P $ are allowed to grow simultaneously, the asymptotic regime is not necessarily the same for MBA and OSA,  as remaining terms are not always negligible. For example, if fixing $ \eta = X^{-2/3} $, $ X\in \lbrace N,C\rbrace $ (we chose $ \alpha=2/3 $ to balance $ Q_{1,var} $ and $ Q_{2,var} $),   the variance term would be controlled by $ \sigma^2 T^{-1} (1+ \frac{P}{\mu C^{1/3}} )$. Thus, unless $ P\le \mu {C^{1/3}} $, MBA could outperform OSA by a factor as large as $ P $.

\textbf{Novelty and proofs.}
Both \Cref{prop:conver_Mini_Batch,prop:conver_One_Shot} are proved in the Appendix~\ref{app:mainproofs}.  Importantly, \Cref{eq:prop11,eq:prop21} respectively imply \Cref{eq:prop12,eq:prop22} under the stated conditions: this is the reason why we only focus on proving equations similar to \Cref{eq:prop11,eq:prop21} for Local-SGD.

\Cref{prop:conver_Mini_Batch} is  similar to the  analysis of \emph{Serial-SGD} for large step size, but with a reduction in the variance proportional to the number of machines. Such a result is derived from the analysis by~\cite{Die_Dur_2018}, combining the approach of \cite{Bac_Mou_2013} with the correct upper bound for smooth strongly convex SGD \citep{Nee_War_Sre_2014}, and controlling similarly higher order moments. While this result is expected, we have not found it under such a simple form in the literature.
\Cref{prop:conver_One_Shot} follows a similar approach, we combine the proof for mini-batch with a control of the iterates of each of the machines. This is closely related to \cite{God_Saa_2017}, but we preserve a non-asymptotic approach.


\textbf{Remark:  link with convergence in function values.} We mainly focus on proving convergence results on the Mahalanobis distance $ \normLigne[2]{F''(\ws)(\itePRav -\ws)} $, which is the natural quantity in such a setting \citep{Bac_Mou_2011, Bac_Mou_2013, God_Saa_2017}.  These results could be translated into function value convergence $ F(\itePRav )- F(\ws) $, using the inequality $  F(\itePRav )- F(\ws)  \le L \mu ^{-2} \normLigne[2]{F''(\ws)(\itePRav -\ws)}$ but the dependence on $ \mu $ would be pessimistic and sub-optimal. However, a similar approach has been used  by \cite{Bac_2014}, under a slightly different set of assumptions (including self-concordance, e.g., for logistic regression), recovering optimal rates. Extension to such a set of assumptions, which relies on tracking  other quantities, is an important direction. 

While  the ``classical proof'', which provides rates  for function values directly (with smoothness, or with uniformly bounded gradients) has a better dependence on $ \mu $, one cannot easily obtain a noise reduction when averaging between machines. Similarly, there is no proof showing that one-shot averaging is asymptotically optimal that relies only on function values. In other words, these proofs do not adequately capture the noise reduction due to averaging. Moreover, such proof techniques relying on function values typically involve a small step size $ 1/(\mu t) $ (because the noise reduction is captured inefficiently). Such step size performs  poorly in practice (initial condition is forgotten slowly), and $ \mu $ is unknown.

{{In conclusion, though they do not directly result  in optimal dependence on $ \mu $  for function values, we believe our approach allows to correctly capture the effect of the noise, and is thus suitable for capturing the effect of local SGD.}}

\textbf{Conclusion:}  for a fixed or limited number of machines, asymptotically, the convergence rate is similar for OSA and MBA. However, non-asymptotically, or when the number of machines also increases, the dominant terms can be as much as $ P^2 $ times smaller for MBA. In the following we provide conditions for Local-SGD to perform as well as MBA (while requiring much fewer communication rounds).

\subsection{Convergence of Local-SGD, FH setting}

For local-SGD we first consider the case of a quadratic function, under the assumption that the noise has a uniformly upper bounded variance. While this set of assumptions is not realistic, it allows an intuitive presentation of  the results. Similar results for settings encompassing LSR and LR follow.  We provide a bound on the moment of an iterate after the communication step $ \hat\ite^{t} $ (i.e., the restart point of the next phase), and on the second order moment of any iterate.    

For $ t \in \un{C}$, we denote $ \boldsymbol{N}_1^t:= \sum_{t'=1}^{t} \Ntp $.

\begin{proposition}[Local-SGD: Quadratic Functions with Bounded Noise]\label{prop:conv_quad_Simple}
	Under Assumptions~\Cref{hyp:quad},~\Cref{ass:def_filt},~\Cref{hyp:addit_noise}, we have the following bound for Local-SGD:  for any $ p\in \un{P}, t\in \un{C}, k \in \un{\Nt} $, 
	\begin{align*}
		&	\expe{\mynorm[2]{\hat \ite^{t-1} -\ws}} \le (1-\eta \mu)^{\boldsymbol{N}_1^{t-1}}  \mynorm[2]{\ite_0-\ws} + +  \frac{\siginf \eta}{P} \frac{1 - (1-\eta \mu)^{\boldsymbol{N}_1^{t-1}}}{\mu}\\
		&	\expe{\mynorm[2]{ \iter{p}{k}{t} -\ws}} \le  (1-\eta \mu)^{\boldsymbol{N}_1^{t-1}+k}  \mynorm[2]{\ite_0-\ws}	+  \siginf \eta \left ( \underbrace{ \frac{1 - (1-\eta \mu)^{\boldsymbol{N}_1^{t-1}}}{P \mu}}_{\text{long term reduced variance}}  + \underbrace{\frac{1 - (1-\eta \mu)^{k}}{\mu} }_{\text{local iteration variance}}\right ).
	\end{align*}
\end{proposition}

To prove such a result, we use the classical technique, and introduce a \emph{ghost} sequence $ \breve \ite_k^{t}:=\frac{1}{P}\sum_{p=1}^{P}  \iter{p}{k}{t} $, and recursively control $ \mynorm[2]{\breve \ite_k^{t}-\ws} $. We conclude by remarking that $ \breve \ite_{\Nt}^{t} =\hat \ite ^{t} $. This proof is given in \Cref{sec:proofquadsimple}. 

\textbf{Interpretation. } The variance bound for the iterates after communication, $\hat{w}^{t}$ exactly  behaves as in mini-batch case: the initialization term decays linearly with the number of local steps, and the variance is reduced proportionally to the number of workers  $P$. On the other hand, the bound on the iterates $ \iter{p}{k}{t} $ shows that the variance of this process is composed of a ``long term'' reduced variance, that accumulates through phases, and is increasingly converging to $ \frac{\siginf \eta}{P \mu} $ and of an extra  variance $\eta \siginf \frac{1 - (1-\eta \mu)^{k}}{\mu} $, that increases within the phase, and is upper bounded by $ \siginf \eta^{2} k $.

In the case of constant step size, the iterates of serial SGD converge to a limit distribution $ \pi_\eta $ that depends on the step size \citep{Die_Dur_2018}. Here, the iterates after communication (or the mini-batch iterates) converge to a distribution with reduced variance $ \pi_{\eta/P} $, thus local iterates periodically restart from a distribution with reduced variance, then slowly ``diverge'' to the distribution with large variance. If the number of local iterations is small enough, the iterates keep a reduced variance. More precisely, we have the following result.
\begin{corollary}\label{cor:maintext}
	If for all $t\in \un{C}  $, $ \Nt\le \frac{1}{\mu \eta P} $, then the second order moment of $ \iter{p}{k}{t} $ admits the same upper bound as the mini-batch iterate $ \hat \ite_{MB}^{\boldsymbol{N}_1^{t-1}+k} $ (Equation~\eqref{eq:prop11}) up to a constant factor of 2. As a consequence, Equation~\eqref{eq:prop12} is still valid, and Local-SGD performs optimally. 
\end{corollary}

\textbf{Interpretation.} This result shows that if the algorithm communicates often enough, the convergence of the Polyak Ruppert iterate $ \itePRav $ is as good as in the mini-batch case, thus it is ``optimal''. Moreover, the minimal number of communication rounds  is easy to define: the maximal number of local steps $ \Nt $ decays as the number of workers and the step size increases. This bound implies that more communication steps are necessary when more machines are used. Note that $(\eta P)^{-1}$ is a large number, as a typical value for $ \eta $ is inversely proportional  to (a power of) the number of local steps for e.g., $ (\sum_{t'=1}^{t} \Ntp)^{-\alpha}$, $ \alpha\in (1/2;1) $.

\begin{example}\label{cor:maintext2}
	With constant number of local steps $\Nt=N$, and learning rate $\eta = \frac{c}{\sqrt{NC}}$ in order to obtain an optimal $O(\frac{\sigma^2}{T})$ parallel convergence rate, local-SGD communicates $O(\frac{\sqrt{NC}}{P\mu})$ times less as compared to mini-batch averaging.   
\end{example}

We believe that this is the first result (with \cite{stich2018local}) that shows a communication reduction proportional to  a power of the number of local steps of a local solver (i.e., $O({\sqrt{NC}})$), compared to mini-batch averaging. 

In the following, we alternatively relax the  bounded variance assumption \Cref{hyp:addit_noise} and the quadratic assumption \Cref{hyp:quad}, and show similar results for local SGD. This allows us to successively cover the cases of least squares regression (LSR) and logistic regression (LR).

\label{sec:CSS_case_quad}
\begin{theorem}\label{th:LocalSGDconv}
	Under either of the following sets of assumptions, the convergence of the Polyak Ruppert iterate $ \itePRav $ is as good as in the mini-batch case, up to a constant:  
	\begin{enumerate}
		\item Assume~\Cref{hyp:quad},~\Cref{ass:def_filt},~\Cref{ass:lip_noisy_gradient_AS},~\Cref{hyp:variancebound}, and for any $t\in \un{C}  $, $ \Nt\le \frac{1}{\mu \eta P} $  and $   \mu \eta^{2}  \boldsymbol{N}_1^t =O(1) $.
		\item Assume~\Cref{hyp:strong_convex},~\Cref{hyp:regularity},~\Cref{ass:def_filt},~\Cref{hyp:addit_noise}, and  for any $t\in \un{C}  $, $ \Nt\le \inf \Big(\frac{1}{\eta P M \expe{\mynorm{\hat{\ite}^{t}-\ws}} },\frac{1}{\mu \eta P}\Big)$. 
	\end{enumerate}
\end{theorem}
These results are derived from  \Cref{prop:conv_quad_DSS} and \Cref{prop:conv_general_function_DSS} which generalize \Cref{prop:conv_quad_Simple}. Those results are proved in Appendix~\ref{app:auxiliary} and \ref{app:aux4} and constitute the main technical challenge of the paper. 

\textbf{Interpretation.} We note that in both of these situations, the optimal rates can be achieved if the communications happen often enough, and beyond such a number of communication rounds, there is no substantial improvement in the convergence. This result corresponds to the effect observed in practice~\citep{Zha_DeS_Re_2016}. The first set of assumption is valid for LSR, the second for LR. In the first case, the maximal number of local steps before communication is upper bounded by the same ratio as in \Cref{cor:maintext}, but the ``constant'' that appears is $\exp(\mu \eta^{2} \boldsymbol{N}_1^t)$, so we need this quantity to be small (which is typically always satisfied in practice) in order to be optimal w.r.t. mini-batch averaging. A similar result as \Cref{cor:maintext2} can be provided reducing the communication by a factor of $O(\frac{\sqrt{NC}}{P\mu})$. 

In the second case, the maximal number of local steps is smaller than before, by a factor $ \mu^{-1} $, but the allowed maximal number of local steps can increase along with the epochs, as $\expe{\mynorm{\hat{\ite}^{t}-\ws}}$ is typically decaying. This adaptive communication frequency has been observed to work well in practice \citep{Zha_DeS_Re_2016}. Assuming optimization on a compact space with radius $R$ for instance, one can obtain a $O(\frac{\sqrt{NC}}{{P^{2}}})$ times improvement in communication, similar to \Cref{cor:maintext2}. 

It is important to remark that these results are only based on upper bounds. While they provide some intuition, comparisons should be handled with caution. Proving corresponding lower bounds is an interesting and important open direction. Moreover, such results might be difficult to use directly in practice, as $ \mu $ is unknown. However, as it is not the limiting factor in \Cref{th:LocalSGDconv}.2, an estimation of $ \expe{\mynorm{\hat{\ite}^{t}-\ws}} $ could allow us to use adaptive phases lengths to minimize communications. 

\section{Main results: On-line Setting}\label{subsec:onlinesetting}
In the on-line setting we consider the particular case of a decaying sequence $ \eta^t_k =(\sum_{t'=1}^{t-1} \Ntp + k) ^{-\alpha} $, for some $ \alpha \in (\frac{1}{2}, 1) $. The analysis is slightly more involved as \Cref{eq:dec_Pol_Jud} results in more terms than in the finite horizon setting (sums do not directly telescope). While the decaying step-size case enables to improve some terms with respect to the finite horizon case (\eg~the speed at which one forgets the initial condition),  the trade-offs concerning communication remain unchanged. We define the following constants to make the presentation clear, for $ \alpha\in (1/2;1) $:

\begin{align*}
R_{bias}(X) &= 1+ X^{2\alpha}\exp\left(-\mu c_\eta X^{1-\alpha}\right) +\frac{1}{(\mu c_\eta)^{\frac{1}{1-\alpha}}} + \frac{M^2c_\eta^2\norm{\ww^0-\ws}^2}{(\mu c_\eta)^{\frac{2}{1-\alpha}}} + \frac{2L^2c_\eta^2}{P(\mu c_\eta)^{\frac{1}{1-\alpha}}},\\
R_{1,var}(X) &= \frac{ X^{2\alpha-1}P }{2\alpha-1}\exp\left(-\frac{\mu X^{1-\alpha}}{2(1-\alpha)}\right) + \frac{ P}{X^{1-\alpha}}\frac{1}{c_\eta \mu}  +  \frac{P}{X \mu^{\frac{2\alpha}{1-\alpha}} c_\eta^{\frac{2}{1-\alpha}}} +  \frac{L^2P c_\eta^{2} }{X^{\alpha} \mu^2}\\ 
\quad R_{2,var}(X) &=\frac{M^2 \sigma^2 P c_\eta^{2} }{\mu^2 X^{2\alpha -1} }.
\end{align*}
Now we present a result similar to \Cref{prop:conver_Mini_Batch} for mini-batch averaging and one shot averaging:
\begin{proposition}[On-line Mini-batch Averaging and One-shot averaging]\label{prop:conver_Mini_Batch_DSS}
	Under the Assumptions ~\Cref{hyp:strong_convex}, ~\Cref{hyp:regularity}, ~\Cref{ass:def_filt}, ~\Cref{ass:lip_noisy_gradient_AS}, ~\Cref{hyp:variancebound} using $ \eta^t_k =(\sum_{t'=1}^{t-1} \Ntp + k) ^{-\alpha} $ we have for respectively mini-batch averaging and one-shot averaging:
	\begin{align*}
	\e{\norm{\nabla^2F(\ww^\star)(\ww-\ww^\star)}^2} \precsim \frac{\norm{\ww^0-\ww^\star}^2}{X^2c_\eta^{2}} R_{bias}(X) + \frac{2\sigma^2}{T} \left (1+ \frac{R_{1,var}(X)}{\kappa}+ \frac{R_{2,var}(X)}{\kappa^2}\right),
	\end{align*}
	with respectively  $ \kappa=1 $ and $ X=N $ for one-shot averaging, and $ \kappa =  P $ and $ X=C $ for mini-batch averaging.
\end{proposition}

\paragraph{Interpretation and comparison. } This proposition is directly derived from \Cref{lem:MBAonline} in   \Cref{sec:proofsonline}. This proposition is similar to \Cref{prop:conver_Mini_Batch,prop:conver_One_Shot}, but the overall convergence rate is better as using decaying step size eventually performs better. For example, the bias term mainly  decays as $ 1/X^2 $ instead of $ 1/(\eta X)^{2} $. This underlines why in practice decaying step size is often preferable. Asymptotically, the variance term is now dominant, and as before, MBA and OSA have similar performance as $ \sigma^2 T^{-1} $.

\paragraph{Optimal step size and asymptotic regimes for $ P,T $}
For a fixed number of machine $ P $, the bias is asymptotically vanishing, and if we ignore the linearly decaying terms and  the dependence on $ \mu $, the resulting dominating  term  in $ R_{1,2 ,var} $ is controlled by $X^{-\min\lbrace(1-\alpha), 
	\alpha, 2\alpha -1\rbrace} $, which would result in an optimal choice of $ \alpha =2/3 $. 

In the non asymptotic regime, where the total number of iterations and $ P $ grow simultaneously, the variance of OSA scales as $ T^{-1} $ as long as $ P X^{-\min\lbrace(1-\alpha), 
	\alpha, 2\alpha -1\rbrace} =O(1)$. In other words, for $ \alpha=2/3 $,  we need $ P\le X^{1/3} $: the number of machines as to be smaller than the number of iterations to the power  $ 1/3 $, in other words, for 1000 iterations, one could only use 10 machines to reach the asymptotic regime where OSA performs similarly to MBA.

\section{Conclusion}

Stochastic approximation and distributed optimization are both very densely studied research areas. However, in practice most distributed applications stick to bulk synchronous mini-batch SGD. While the algorithm has desirable convergence properties, it suffers from a huge communication bottleneck. In this paper we have analyzed a natural generalization of mini-batch averaging, Local SGD. Our analysis is non-asymptotic, which helps us to better understand the exact communication trade-offs. We give feasible lower bounds on communication frequency which significantly reduce the need for communication, while providing similar non-asymptotic convergence as mini-batch averaging. Our results apply to common loss functions, and use large step sizes. Further, our analysis unifies and extends all the scattered results for one-shot averaging, mini-batch averaging and local SGD, providing an intuitive understanding of their behavior.   

Some important future directions are obtaining lower bounds, studying observable quantities to predict an adaptive communication frequency and relaxing some of the technical assumptions required by the analysis. The on-line case, experiments, proofs, additional materials and a review of distributed optimization follow in the appendix.

\section{Acknowledgments}
We thank Martin Jaggi, Sebastian Stichs, and Sai Praneeth Reddy for helpful discussions. 

\bibliography{Biblio/biblio_all}

\begin{thebibliography}{106}
\providecommand{\natexlab}[1]{#1}
\providecommand{\url}[1]{\texttt{#1}}
\expandafter\ifx\csname urlstyle\endcsname\relax
  \providecommand{\doi}[1]{doi: #1}\else
  \providecommand{\doi}{doi: \begingroup \urlstyle{rm}\Url}\fi

\bibitem[{Agarwal} and {Duchi}(2011)]{agarwal2011delayed}
A.~{Agarwal} and J.~C. {Duchi}.
\newblock {Distributed Delayed Stochastic Optimization}.
\newblock \emph{ArXiv e-prints}, April 2011.

\bibitem[{Alistarh} et~al.(2018){Alistarh}, {De Sa}, and
  {Konstantinov}]{alistarh2018convergence}
D.~{Alistarh}, C.~{De Sa}, and N.~{Konstantinov}.
\newblock {The Convergence of Stochastic Gradient Descent in Asynchronous
  Shared Memory}.
\newblock \emph{ArXiv e-prints}, March 2018.

\bibitem[Alistarh et~al.(2016)Alistarh, Li, Tomioka, and
  Vojnovic]{alistarh2016randomized}
Dan Alistarh, Jerry Li, Ryota Tomioka, and Milan Vojnovic.
\newblock {QSGD:} randomized quantization for communication-optimal stochastic
  gradient descent.
\newblock \emph{CoRR}, abs/1610.02132, 2016.
\newblock URL \url{http://arxiv.org/abs/1610.02132}.

\bibitem[Arjevani and Shamir(2015)]{arjevani2015communication}
Yossi Arjevani and Ohad Shamir.
\newblock Communication complexity of distributed convex learning and
  optimization.
\newblock \emph{CoRR}, abs/1506.01900, 2015.
\newblock URL \url{http://arxiv.org/abs/1506.01900}.

\bibitem[Bach(2014)]{Bac_2014}
F.~Bach.
\newblock {Adaptivity of averaged stochastic gradient descent to local strong
  convexity for logistic regression}.
\newblock \emph{J. Mach. Learn. Res.}, 15\penalty0 (1):\penalty0 595--627,
  January 2014.

\bibitem[{Bach} and {Moulines}(2013)]{Bac_Mou_2013}
F.~{Bach} and E.~{Moulines}.
\newblock {Non-strongly-convex smooth stochastic approximation with convergence
  rate O(1/n)}.
\newblock \emph{Advances in Neural Information Processing Systems (NIPS)},
  2013.

\bibitem[Bach and Moulines(2011)]{Bac_Mou_2011}
Francis Bach and Eric Moulines.
\newblock {Non-asymptotic Analysis of Stochastic Approximation Algorithms for
  Machine Learning}.
\newblock In \emph{{Proceedings of the 24th International Conference on Neural
  Information Processing Systems}}, {NIPS'11}, pages 451--459, USA, 2011.
  Curran Associates Inc.
\newblock ISBN 978-1-61839-599-3.
\newblock URL \url{http://dl.acm.org/citation.cfm?id=2986459.2986510}.

\bibitem[Boyd et~al.(2011)Boyd, Parikh, Chu, Peleato, and
  Eckstein]{2011boydADMM}
Stephen Boyd, Neal Parikh, Eric Chu, Borja Peleato, and Jonathan Eckstein.
\newblock Distributed optimization and statistical learning via the alternating
  direction method of multipliers.
\newblock \emph{Found. Trends Mach. Learn.}, 3\penalty0 (1):\penalty0 1--122,
  January 2011.
\newblock ISSN 1935-8237.
\newblock \doi{10.1561/2200000016}.
\newblock URL \url{http://dx.doi.org/10.1561/2200000016}.

\bibitem[Braverman et~al.(2015)Braverman, Garg, Ma, Nguyen, and
  Woodruff]{braverman2015communication}
Mark Braverman, Ankit Garg, Tengyu Ma, Huy~L. Nguyen, and David~P. Woodruff.
\newblock Communication lower bounds for statistical estimation problems via a
  distributed data processing inequality.
\newblock \emph{CoRR}, abs/1506.07216, 2015.
\newblock URL \url{http://arxiv.org/abs/1506.07216}.

\bibitem[{{\c S}im{\c s}ekli} et~al.(2018){{\c S}im{\c s}ekli}, {Y{\i}ld{\i}z},
  {Nguyen}, {Richard}, and {Taylan Cemgil}]{simsekli2018MCMC}
U.~{{\c S}im{\c s}ekli}, {\c C}.~{Y{\i}ld{\i}z}, T.~H. {Nguyen}, G.~{Richard},
  and A.~{Taylan Cemgil}.
\newblock {Asynchronous Stochastic Quasi-Newton MCMC for Non-Convex
  Optimization}.
\newblock \emph{ArXiv e-prints}, June 2018.

\bibitem[{Chen} et~al.(2016{\natexlab{a}}){Chen}, {Pan}, {Monga}, {Bengio}, and
  {Jozefowicz}]{chen2016revisiting}
J.~{Chen}, X.~{Pan}, R.~{Monga}, S.~{Bengio}, and R.~{Jozefowicz}.
\newblock {Revisiting Distributed Synchronous SGD}.
\newblock \emph{ArXiv e-prints}, April 2016{\natexlab{a}}.

\bibitem[{Chen} et~al.(2016{\natexlab{b}}){Chen}, {Luo}, and
  {Zhang}]{chen2016communication}
Z.~{Chen}, L.~{Luo}, and Z.~{Zhang}.
\newblock {Communication Lower Bounds for Distributed Convex Optimization:
  Partition Data on Features}.
\newblock \emph{ArXiv e-prints}, December 2016{\natexlab{b}}.

\bibitem[Chin et~al.(2015)Chin, Zhuang, Juan, and Lin]{Chin2015fast}
Wei-Sheng Chin, Yong Zhuang, Yu-Chin Juan, and Chih-Jen Lin.
\newblock A fast parallel stochastic gradient method for matrix factorization
  in shared memory systems.
\newblock \emph{ACM Trans. Intell. Syst. Technol.}, 6\penalty0 (1):\penalty0
  2:1--2:24, March 2015.
\newblock ISSN 2157-6904.
\newblock \doi{10.1145/2668133}.
\newblock URL \url{http://doi.acm.org/10.1145/2668133}.

\bibitem[{De} and {Goldstein}(2015)]{de2015efficient}
S.~{De} and T.~{Goldstein}.
\newblock {Efficient Distributed SGD with Variance Reduction}.
\newblock \emph{ArXiv e-prints}, December 2015.

\bibitem[D{\'e}fossez and Bach(2015)]{Def_Bac_2015}
A.~D{\'e}fossez and F.~Bach.
\newblock {Averaged least-mean-squares: bias-variance trade-offs and optimal
  sampling distributions}.
\newblock In \emph{{Proceedings of the International Conference on Artificial
  Intelligence and Statistics, {(AISTATS)}}}, 2015.

\bibitem[Dekel et~al.(2012{\natexlab{a}})Dekel, Gilad-Bachrach, Shamir, and
  Xiao]{Dek_2012}
Ofer Dekel, Ran Gilad-Bachrach, Ohad Shamir, and Lin Xiao.
\newblock Optimal distributed online prediction using mini-batches.
\newblock \emph{Journal of Machine Learning Research}, 13\penalty0
  (Jan):\penalty0 165--202, 2012{\natexlab{a}}.

\bibitem[Dekel et~al.(2012{\natexlab{b}})Dekel, Gilad-Bachrach, Shamir, and
  Xiao]{dekel2012optimal}
Ofer Dekel, Ran Gilad-Bachrach, Ohad Shamir, and Lin Xiao.
\newblock Optimal distributed online prediction using mini-batches.
\newblock \emph{Journal of Machine Learning Research}, 13\penalty0
  (Jan):\penalty0 165--202, 2012{\natexlab{b}}.

\bibitem[Delalleau and Bengio(2007)]{Del_Ben_2007}
Olivier Delalleau and Yoshua Bengio.
\newblock Parallel stochastic gradient descent.
\newblock 2007.

\bibitem[{Dieuleveut} et~al.(2016){Dieuleveut}, {Flammarion}, and
  {Bach}]{Die_Fla_Bac_2016}
A.~{Dieuleveut}, N.~{Flammarion}, and F.~{Bach}.
\newblock {Harder, Better, Faster, Stronger Convergence Rates for Least-Squares
  Regression}.
\newblock \emph{ArXiv e-prints}, February 2016.

\bibitem[Dieuleveut and Bach(2016)]{Die_Bac_2015}
Aymeric Dieuleveut and Francis Bach.
\newblock Nonparametric stochastic approximation with large step-sizes.
\newblock \emph{Ann. Statist.}, 44\penalty0 (4):\penalty0 1363--1399, 08 2016.
\newblock \doi{10.1214/15-AOS1391}.
\newblock URL \url{http://dx.doi.org/10.1214/15-AOS1391}.

\bibitem[Dieuleveut et~al.(2017)Dieuleveut, Durmus, and Bach]{Die_Dur_2018}
Aymeric Dieuleveut, Alain Durmus, and Francis Bach.
\newblock Bridging the gap between constant step size stochastic gradient
  descent and markov chains.
\newblock \emph{arXiv preprint arXiv:1707.06386}, 2017.

\bibitem[{Duchi} et~al.(2014){Duchi}, {Jordan}, {Wainwright}, and
  {Zhang}]{duchi2014optimality}
J.~C. {Duchi}, M.~I. {Jordan}, M.~J. {Wainwright}, and Y.~{Zhang}.
\newblock {Optimality guarantees for distributed statistical estimation}.
\newblock \emph{ArXiv e-prints}, May 2014.

\bibitem[{Duchi} et~al.(2015){Duchi}, {Chaturapruek}, and
  {R{\'e}}]{duchi2015asynchronous}
J.~C. {Duchi}, S.~{Chaturapruek}, and C.~{R{\'e}}.
\newblock {Asynchronous stochastic convex optimization}.
\newblock \emph{ArXiv e-prints}, August 2015.

\bibitem[Fabian(1968)]{Fab_1968}
Vaclav Fabian.
\newblock On asymptotic normality in stochastic approximation.
\newblock \emph{The Annals of Mathematical Statistics}, pages 1327--1332, 1968.

\bibitem[{Fang} and {Klabjan}(2018)]{fang2018large}
B.~{Fang} and D.~{Klabjan}.
\newblock {A Stochastic Large-scale Machine Learning Algorithm for Distributed
  Features and Observations}.
\newblock \emph{ArXiv e-prints}, March 2018.

\bibitem[{Feyzmahdavian} et~al.(2015){Feyzmahdavian}, {Aytekin}, and
  {Johansson}]{feyzmahdavian2015mini}
H.~R. {Feyzmahdavian}, A.~{Aytekin}, and M.~{Johansson}.
\newblock {An Asynchronous Mini-Batch Algorithm for Regularized Stochastic
  Optimization}.
\newblock \emph{ArXiv e-prints}, May 2015.

\bibitem[{Gadat} and {Panloup}(2017)]{Gad_Pan_2107}
S.~{Gadat} and F.~{Panloup}.
\newblock {Optimal non-asymptotic bound of the Ruppert-Polyak averaging without
  strong convexity}.
\newblock \emph{ArXiv e-prints}, September 2017.

\bibitem[{Godichon} and {Saadane}(2017)]{God_Saa_2017}
A.~Baggioni {Godichon} and S.~{Saadane}.
\newblock {On the rates of convergence of Parallelized Averaged Stochastic
  Gradient Algorithms}.
\newblock \emph{ArXiv e-prints}, October 2017.

\bibitem[Godichon and Saadane(2017)]{godichon2017rates}
Baggioni~Antoine Godichon and Sofiane Saadane.
\newblock On the rates of convergence of parallelized averaged stochastic
  gradient algorithms.
\newblock \emph{arXiv preprint arXiv:1710.07926}, 2017.

\bibitem[{Goyal} et~al.(2017){Goyal}, {Doll{\'a}r}, {Girshick}, {Noordhuis},
  {Wesolowski}, {Kyrola}, {Tulloch}, {Jia}, and {He}]{goyal2017imagenet}
P.~{Goyal}, P.~{Doll{\'a}r}, R.~{Girshick}, P.~{Noordhuis}, L.~{Wesolowski},
  A.~{Kyrola}, A.~{Tulloch}, Y.~{Jia}, and K.~{He}.
\newblock {Accurate, Large Minibatch SGD: Training ImageNet in 1 Hour}.
\newblock \emph{ArXiv e-prints}, June 2017.

\bibitem[Goyal et~al.(2017)Goyal, Doll{\'a}r, Girshick, Noordhuis, Wesolowski,
  Kyrola, Tulloch, Jia, and He]{Goy_2017}
Priya Goyal, Piotr Doll{\'a}r, Ross Girshick, Pieter Noordhuis, Lukasz
  Wesolowski, Aapo Kyrola, Andrew Tulloch, Yangqing Jia, and Kaiming He.
\newblock Accurate, large minibatch sgd: training imagenet in 1 hour.
\newblock \emph{arXiv preprint arXiv:1706.02677}, 2017.

\bibitem[Gupta et~al.(2015)Gupta, Agrawal, Gopalakrishnan, and
  Narayanan]{gupta2015deep}
Suyog Gupta, Ankur Agrawal, Kailash Gopalakrishnan, and Pritish Narayanan.
\newblock Deep learning with limited numerical precision.
\newblock \emph{CoRR}, abs/1502.02551, 2015.
\newblock URL \url{http://arxiv.org/abs/1502.02551}.

\bibitem[{Jain} et~al.(2016){Jain}, {Kakade}, {Kidambi}, {Netrapalli}, and
  {Sidford}]{Jai_Kak_Kid_2016}
P.~{Jain}, S.~M. {Kakade}, R.~{Kidambi}, P.~{Netrapalli}, and A.~{Sidford}.
\newblock {Parallelizing Stochastic Approximation Through Mini-Batching and
  Tail-Averaging}.
\newblock \emph{ArXiv e-prints}, October 2016.

\bibitem[Jain et~al.(2017)Jain, Kakade, Kidambi, Netrapalli, and
  Sidford]{Jai_Kak_Kid_2017}
P.~Jain, S.~M. Kakade, R.~Kidambi, P.~Netrapalli, and A.~Sidford.
\newblock {Accelerating Stochastic Gradient Descent}.
\newblock \emph{arXiv preprint arXiv:1704.08227}, 2017.

\bibitem[{Keuper} and {Pfreundt}(2015)]{keuper2015parallel}
J.~{Keuper} and F.-J. {Pfreundt}.
\newblock {Asynchronous Parallel Stochastic Gradient Descent - A Numeric Core
  for Scalable Distributed Machine Learning Algorithms}.
\newblock \emph{ArXiv e-prints}, May 2015.

\bibitem[{Khirirat} et~al.(2018){Khirirat}, {Feyzmahdavian}, and
  {Johansson}]{khirirat2018distributed}
S.~{Khirirat}, H.~R. {Feyzmahdavian}, and M.~{Johansson}.
\newblock {Distributed learning with compressed gradients}.
\newblock \emph{ArXiv e-prints}, June 2018.

\bibitem[Konecn{\'{y}} et~al.(2015)Konecn{\'{y}}, McMahan, and
  Ramage]{jakub2015federated}
Jakub Konecn{\'{y}}, Brendan McMahan, and Daniel Ramage.
\newblock Federated optimization: Distributed optimization beyond the
  datacenter.
\newblock \emph{CoRR}, abs/1511.03575, 2015.
\newblock URL \url{http://arxiv.org/abs/1511.03575}.

\bibitem[Konecn{\'{y}} et~al.(2016)Konecn{\'{y}}, McMahan, Yu, Richt{\'{a}}rik,
  Suresh, and Bacon]{jakub2016federated}
Jakub Konecn{\'{y}}, H.~Brendan McMahan, Felix~X. Yu, Peter Richt{\'{a}}rik,
  Ananda~Theertha Suresh, and Dave Bacon.
\newblock Federated learning: Strategies for improving communication
  efficiency.
\newblock \emph{CoRR}, abs/1610.05492, 2016.
\newblock URL \url{http://arxiv.org/abs/1610.05492}.

\bibitem[{Lacoste-Julien} et~al.(2012){Lacoste-Julien}, {Schmidt}, and
  {Bach}]{Lac_Sch_Bac_2012}
S.~{Lacoste-Julien}, M.~{Schmidt}, and F.~{Bach}.
\newblock {A simpler approach to obtaining an {O}(1/t) rate for the stochastic
  projected subgradient method}.
\newblock ArXiv e-prints 1212.2002, 2012.

\bibitem[{Langford} et~al.(2009){Langford}, {Smola}, and
  {Zinkevich}]{langford2009slow}
J.~{Langford}, A.~{Smola}, and M.~{Zinkevich}.
\newblock {Slow Learners are Fast}.
\newblock \emph{ArXiv e-prints}, November 2009.

\bibitem[{Leblond} et~al.(2018){Leblond}, {Pedregosa}, and
  {Lacoste-Julien}]{leblond2018improve}
R.~{Leblond}, F.~{Pedregosa}, and S.~{Lacoste-Julien}.
\newblock {Improved asynchronous parallel optimization analysis for stochastic
  incremental methods}.
\newblock \emph{ArXiv e-prints}, January 2018.

\bibitem[{Lee} et~al.(2015){Lee}, {Lin}, {Ma}, and {Yang}]{lee2015distributed}
J.~D. {Lee}, Q.~{Lin}, T.~{Ma}, and T.~{Yang}.
\newblock {Distributed Stochastic Variance Reduced Gradient Methods and A Lower
  Bound for Communication Complexity}.
\newblock \emph{ArXiv e-prints}, July 2015.

\bibitem[Lee et~al.(2014)Lee, Kim, Zheng, Ho, Gibson, and Xing]{lee2014model}
Seunghak Lee, Jin~Kyu Kim, Xun Zheng, Qirong Ho, Garth~A Gibson, and Eric~P
  Xing.
\newblock On model parallelization and scheduling strategies for distributed
  machine learning.
\newblock In Z.~Ghahramani, M.~Welling, C.~Cortes, N.~D. Lawrence, and K.~Q.
  Weinberger, editors, \emph{Advances in Neural Information Processing Systems
  27}, pages 2834--2842. Curran Associates, Inc., 2014.

\bibitem[Li et~al.(2014{\natexlab{a}})Li, Wu, Xu, Shi, and Shi]{li2014fast}
Fanglin Li, Bin Wu, Liutong Xu, Chuan Shi, and Jing Shi.
\newblock A fast distributed stochastic gradient descent algorithm for matrix
  factorization, 24 Aug 2014{\natexlab{a}}.
\newblock URL \url{http://proceedings.mlr.press/v36/li14.html}.

\bibitem[Li et~al.(2014{\natexlab{b}})Li, Andersen, Smola, and
  Yu]{Li2014communication}
Mu~Li, David~G. Andersen, Alexander Smola, and Kai Yu.
\newblock Communication efficient distributed machine learning with the
  parameter server.
\newblock In \emph{Proceedings of the 27th International Conference on Neural
  Information Processing Systems - Volume 1}, NIPS'14, pages 19--27, Cambridge,
  MA, USA, 2014{\natexlab{b}}. MIT Press.
\newblock URL \url{http://dl.acm.org/citation.cfm?id=2968826.2968829}.

\bibitem[Li et~al.(2014{\natexlab{c}})Li, Zhang, Chen, and Smola]{Li_2014}
Mu~Li, Tong Zhang, Yuqiang Chen, and Alexander~J Smola.
\newblock Efficient mini-batch training for stochastic optimization.
\newblock In \emph{Proceedings of the 20th ACM SIGKDD international conference
  on Knowledge discovery and data mining}, pages 661--670. ACM,
  2014{\natexlab{c}}.

\bibitem[Li et~al.(2014{\natexlab{d}})Li, Zhang, Chen, and
  Smola]{li2014efficient}
Mu~Li, Tong Zhang, Yuqiang Chen, and Alexander~J Smola.
\newblock Efficient mini-batch training for stochastic optimization.
\newblock In \emph{Proceedings of the 20th ACM SIGKDD international conference
  on Knowledge discovery and data mining}, pages 661--670. ACM,
  2014{\natexlab{d}}.

\bibitem[{Lian} et~al.(2015){Lian}, {Huang}, {Li}, and
  {Liu}]{lian2015asynchronous}
X.~{Lian}, Y.~{Huang}, Y.~{Li}, and J.~{Liu}.
\newblock {Asynchronous Parallel Stochastic Gradient for Nonconvex
  Optimization}.
\newblock \emph{ArXiv e-prints}, June 2015.

\bibitem[{Lian} et~al.(2017{\natexlab{a}}){Lian}, {Zhang}, {Zhang}, {Hsieh},
  {Zhang}, and {Liu}]{lian2017decentralized}
X.~{Lian}, C.~{Zhang}, H.~{Zhang}, C.-J. {Hsieh}, W.~{Zhang}, and J.~{Liu}.
\newblock {Can Decentralized Algorithms Outperform Centralized Algorithms? A
  Case Study for Decentralized Parallel Stochastic Gradient Descent}.
\newblock \emph{ArXiv e-prints}, May 2017{\natexlab{a}}.

\bibitem[{Lian} et~al.(2017{\natexlab{b}}){Lian}, {Zhang}, {Zhang}, and
  {Liu}]{lian2017asynchronous}
X.~{Lian}, W.~{Zhang}, C.~{Zhang}, and J.~{Liu}.
\newblock {Asynchronous Decentralized Parallel Stochastic Gradient Descent}.
\newblock \emph{ArXiv e-prints}, October 2017{\natexlab{b}}.

\bibitem[{Lin} et~al.(2018){Lin}, {Stich}, and {Jaggi}]{Tao2018local}
T.~{Lin}, S.~U. {Stich}, and M.~{Jaggi}.
\newblock {Don't Use Large Mini-Batches, Use Local SGD}.
\newblock \emph{ArXiv e-prints}, August 2018.

\bibitem[{Ma} and {Tak{\'a}{\v c}}(2015)]{ma2015partitioning}
C.~{Ma} and M.~{Tak{\'a}{\v c}}.
\newblock {Partitioning Data on Features or Samples in Communication-Efficient
  Distributed Optimization?}
\newblock \emph{ArXiv e-prints}, October 2015.

\bibitem[{Ma} et~al.(2015){Ma}, {Smith}, {Jaggi}, {Jordan}, {Richt{\'a}rik},
  and {Tak{\'a}{\v c}}]{ma2015adding}
C.~{Ma}, V.~{Smith}, M.~{Jaggi}, M.~I. {Jordan}, P.~{Richt{\'a}rik}, and
  M.~{Tak{\'a}{\v c}}.
\newblock {Adding vs. Averaging in Distributed Primal-Dual Optimization}.
\newblock \emph{ArXiv e-prints}, February 2015.

\bibitem[Ma et~al.(2017)Ma, Kone{\v{c}}n{\`y}, Jaggi, Smith, Jordan,
  Richt{\'a}rik, and Tak{\'a}{\v{c}}]{ma2017distributed}
Chenxin Ma, Jakub Kone{\v{c}}n{\`y}, Martin Jaggi, Virginia Smith, Michael~I
  Jordan, Peter Richt{\'a}rik, and Martin Tak{\'a}{\v{c}}.
\newblock Distributed optimization with arbitrary local solvers.
\newblock \emph{Optimization Methods and Software}, 32\penalty0 (4):\penalty0
  813--848, 2017.

\bibitem[{Mania} et~al.(2015){Mania}, {Pan}, {Papailiopoulos}, {Recht},
  {Ramchandran}, and {Jordan}]{mania2015perturbed}
H.~{Mania}, X.~{Pan}, D.~{Papailiopoulos}, B.~{Recht}, K.~{Ramchandran}, and
  M.~I. {Jordan}.
\newblock {Perturbed Iterate Analysis for Asynchronous Stochastic
  Optimization}.
\newblock \emph{ArXiv e-prints}, July 2015.

\bibitem[Mcdonald et~al.(2009)Mcdonald, Mohri, Silberman, Walker, and
  Mann]{mcdonald2009efficient}
Ryan Mcdonald, Mehryar Mohri, Nathan Silberman, Dan Walker, and Gideon~S Mann.
\newblock Efficient large-scale distributed training of conditional maximum
  entropy models.
\newblock In \emph{Advances in Neural Information Processing Systems}, pages
  1231--1239, 2009.

\bibitem[McDonald et~al.(2010)McDonald, Hall, and
  Mann]{mcdonald2010distributed}
Ryan McDonald, Keith Hall, and Gideon Mann.
\newblock Distributed training strategies for the structured perceptron.
\newblock In \emph{Human Language Technologies: The 2010 Annual Conference of
  the North American Chapter of the Association for Computational Linguistics},
  pages 456--464. Association for Computational Linguistics, 2010.

\bibitem[McMahan et~al.(2016)McMahan, Moore, Ramage, and
  y~Arcas]{mcmahan2016federated}
H.~Brendan McMahan, Eider Moore, Daniel Ramage, and Blaise~Ag{\"{u}}era
  y~Arcas.
\newblock Federated learning of deep networks using model averaging.
\newblock \emph{CoRR}, abs/1602.05629, 2016.
\newblock URL \url{http://arxiv.org/abs/1602.05629}.

\bibitem[Meng et~al.(2012)Meng, Wiesel, and Hero]{meng2012distributed}
Z.~Meng, A.~Wiesel, and A.~O. Hero.
\newblock Distributed principal component analysis on networks via directed
  graphical models, March 2012.
\newblock ISSN 2379-190X.

\bibitem[Moulines and Bach(2011)]{moulines2011non}
Eric Moulines and Francis~R Bach.
\newblock Non-asymptotic analysis of stochastic approximation algorithms for
  machine learning.
\newblock In \emph{Advances in Neural Information Processing Systems}, pages
  451--459, 2011.

\bibitem[Na et~al.(2017)Na, Ko, Kung, and Mukhopadhyay]{Na2017OnchipTO}
Taesik Na, Jong~Hwan Ko, Jaeha Kung, and Saibal Mukhopadhyay.
\newblock On-chip training of recurrent neural networks with limited numerical
  precision.
\newblock \emph{2017 International Joint Conference on Neural Networks
  (IJCNN)}, pages 3716--3723, 2017.

\bibitem[Najafabadi et~al.(2017)Najafabadi, Khoshgoftaar, Villanustre, and
  Holt]{najafabadi2017large}
Maryam~M Najafabadi, Taghi~M Khoshgoftaar, Flavio Villanustre, and John Holt.
\newblock Large-scale distributed l-bfgs.
\newblock \emph{Journal of Big Data}, 4\penalty0 (1):\penalty0 22, 2017.

\bibitem[Needell et~al.(2014)Needell, Ward, and Srebro]{Nee_War_Sre_2014}
Deanna Needell, Rachel Ward, and Nati Srebro.
\newblock {Stochastic Gradient Descent, Weighted Sampling, and the Randomized
  Kaczmarz algorithm}.
\newblock In Z.~Ghahramani, M.~Welling, C.~Cortes, N.~D. Lawrence, and K.~Q.
  Weinberger, editors, \emph{{Advances in Neural Information Processing Systems
  27}}, pages 1017--1025. Curran Associates, Inc., 2014.

\bibitem[Nemirovski et~al.(2009)Nemirovski, Juditsky, Lan, and
  Shapiro]{Nem_Jud_Lan_2009}
A.~Nemirovski, A.~Juditsky, G.~Lan, and A.~Shapiro.
\newblock {Robust Stochastic Approximation Approach to Stochastic Programming}.
\newblock \emph{SIAM J. on Optimization}, 19\penalty0 (4):\penalty0 1574--1609,
  January 2009.
\newblock ISSN 1052-6234.
\newblock \doi{10.1137/070704277}.
\newblock URL \url{http://dx.doi.org/10.1137/070704277}.

\bibitem[Nesterov(2004)]{Nes_2004}
Y.~Nesterov.
\newblock \emph{{Introductory Lectures on Convex Optimization: A Basic
  Course}}.
\newblock {Applied Optimization}. Springer, 2004.
\newblock ISBN 9781402075537.
\newblock URL \url{http://books.google.fr/books?id=VyYLem-l3CgC}.

\bibitem[Nesterov and Vial(2008)]{Nes_Via_2008}
Yu. Nesterov and J.~Ph. Vial.
\newblock {Confidence Level Solutions for Stochastic Programming}.
\newblock \emph{Automatica}, 44\penalty0 (6):\penalty0 1559--1568, June 2008.
\newblock ISSN 0005-1098.
\newblock \doi{10.1016/j.automatica.2008.01.017}.
\newblock URL \url{http://dx.doi.org/10.1016/j.automatica.2008.01.017}.

\bibitem[{Niu} et~al.(2011){Niu}, {Recht}, {Re}, and {Wright}]{feng2011hogwild}
F.~{Niu}, B.~{Recht}, C.~{Re}, and S.~J. {Wright}.
\newblock {HOGWILD!: A Lock-Free Approach to Parallelizing Stochastic Gradient
  Descent}.
\newblock \emph{ArXiv e-prints}, June 2011.

\bibitem[Oh et~al.(2015)Oh, Han, Yu, and Jiang]{Oh2015fast}
Jinoh Oh, Wook-Shin Han, Hwanjo Yu, and Xiaoqian Jiang.
\newblock Fast and robust parallel sgd matrix factorization.
\newblock In \emph{Proceedings of the 21th ACM SIGKDD International Conference
  on Knowledge Discovery and Data Mining}, KDD '15, pages 865--874, New York,
  NY, USA, 2015. ACM.
\newblock ISBN 978-1-4503-3664-2.
\newblock \doi{10.1145/2783258.2783322}.
\newblock URL \url{http://doi.acm.org/10.1145/2783258.2783322}.

\bibitem[{Paine} et~al.(2013){Paine}, {Jin}, {Yang}, {Lin}, and
  {Huang}]{paine2013GPU}
T.~{Paine}, H.~{Jin}, J.~{Yang}, Z.~{Lin}, and T.~{Huang}.
\newblock {GPU Asynchronous Stochastic Gradient Descent to Speed Up Neural
  Network Training}.
\newblock \emph{ArXiv e-prints}, December 2013.

\bibitem[{Pedregosa} et~al.(2017){Pedregosa}, {Leblond}, and
  {Lacoste-Julien}]{pedregosa2017Nonsmooth}
F.~{Pedregosa}, R.~{Leblond}, and S.~{Lacoste-Julien}.
\newblock {Breaking the Nonsmooth Barrier: A Scalable Parallel Method for
  Composite Optimization}.
\newblock \emph{ArXiv e-prints}, July 2017.

\bibitem[Polyak and Juditsky(1992)]{Pol_Jud_1992}
B.~T. Polyak and A.~B. Juditsky.
\newblock {Acceleration of stochastic approximation by averaging}.
\newblock \emph{SIAM J. Control Optim.}, 30\penalty0 (4):\penalty0 838--855,
  1992.

\bibitem[{Rakhlin} et~al.(2011){Rakhlin}, {Shamir}, and
  {Sridharan}]{Rak_Sha_Sri_2011}
A.~{Rakhlin}, O.~{Shamir}, and K.~{Sridharan}.
\newblock {Making Gradient Descent Optimal for Strongly Convex Stochastic
  Optimization}.
\newblock \emph{ArXiv e-prints}, September 2011.

\bibitem[Rakhlin et~al.(2012)Rakhlin, Shamir, Sridharan,
  et~al.]{rakhlin2012making}
Alexander Rakhlin, Ohad Shamir, Karthik Sridharan, et~al.
\newblock Making gradient descent optimal for strongly convex stochastic
  optimization.
\newblock In \emph{ICML}. Citeseer, 2012.

\bibitem[Recht et~al.(2011)Recht, Re, Wright, and Niu]{Rec_Re_2011}
Benjamin Recht, Christopher Re, Stephen Wright, and Feng Niu.
\newblock Hogwild: A lock-free approach to parallelizing stochastic gradient
  descent.
\newblock In \emph{Advances in neural information processing systems}, pages
  693--701, 2011.

\bibitem[{Reddi} et~al.(2015){Reddi}, {Hefny}, {Sra}, {P{\'o}czos}, and
  {Smola}]{reddi2015variance}
S.~J. {Reddi}, A.~{Hefny}, S.~{Sra}, B.~{P{\'o}czos}, and A.~{Smola}.
\newblock {On Variance Reduction in Stochastic Gradient Descent and its
  Asynchronous Variants}.
\newblock \emph{ArXiv e-prints}, June 2015.

\bibitem[{Reddi} et~al.(2016){Reddi}, {Kone{\v c}n{\'y}}, {Richt{\'a}rik},
  {P{\'o}cz{\'o}s}, and {Smola}]{reddi2016aide}
S.~J. {Reddi}, J.~{Kone{\v c}n{\'y}}, P.~{Richt{\'a}rik}, B.~{P{\'o}cz{\'o}s},
  and A.~{Smola}.
\newblock {AIDE: Fast and Communication Efficient Distributed Optimization}.
\newblock \emph{ArXiv e-prints}, August 2016.

\bibitem[{Robbins} and {Monro}(1951)]{Rob_Mon_1951}
H.~{Robbins} and S.~{Monro}.
\newblock {A stochastic approxiation method}.
\newblock \emph{The Annals of mathematical Statistics}, 22\penalty0
  (3):\penalty0 400--407, 1951.

\bibitem[Rosenblatt and Nadler(2016)]{rosenblatt2016optimality}
Jonathan~D. Rosenblatt and Boaz Nadler.
\newblock On the optimality of averaging in distributed statistical learning.
\newblock \emph{Information and Inference: A Journal of the IMA}, 5\penalty0
  (4):\penalty0 379--404, 2016.
\newblock \doi{10.1093/imaiai/iaw013}.
\newblock URL \url{http://dx.doi.org/10.1093/imaiai/iaw013}.

\bibitem[Ruppert(1988)]{Rup_1988}
D.~Ruppert.
\newblock {Efficient estimations from a slowly convergent Robbins-Monro
  process}.
\newblock Technical report, Cornell University Operations Research and
  Industrial Engineering, 1988.

\bibitem[Sa et~al.(2015)Sa, Zhang, Olukotun, and R{\'{e}}]{sa2015taming}
Christopher~De Sa, Ce~Zhang, Kunle Olukotun, and Christopher R{\'{e}}.
\newblock Taming the wild: {A} unified analysis of hogwild!-style algorithms.
\newblock \emph{CoRR}, abs/1506.06438, 2015.
\newblock URL \url{http://arxiv.org/abs/1506.06438}.

\bibitem[{Scaman} et~al.(2017){Scaman}, {Bach}, {Bubeck}, {Tat Lee}, and
  {Massouli{\'e}}]{scaman2017optimal}
K.~{Scaman}, F.~{Bach}, S.~{Bubeck}, Y.~{Tat Lee}, and L.~{Massouli{\'e}}.
\newblock {Optimal algorithms for smooth and strongly convex distributed
  optimization in networks}.
\newblock \emph{ArXiv e-prints}, February 2017.

\bibitem[Shalev-Shwartz et~al.(2009)Shalev-Shwartz, Shamir, Srebro, and
  Sridharan]{Sha_Sha_Sre_2009}
S.~Shalev-Shwartz, O.~Shamir, N.~Srebro, and K.~Sridharan.
\newblock {Stochastic convex optimization.}
\newblock In \emph{{Proceedings of the International Conference on Learning
  Theory (COLT)}}, 2009.

\bibitem[{Shamir} and {Zhang}(2013)]{Sha_Zha_2013}
O.~{Shamir} and T.~{Zhang}.
\newblock {Stochastic Gradient Descent for Non-smooth Optimization: Convergence
  Results and Optimal Averaging Schemes}.
\newblock \emph{Proceedings of the 30$^th$ International Conference on Machine
  Learning}, 2013.

\bibitem[Shamir et~al.(2014)Shamir, Srebro, and Zhang]{shamir2014communication}
Ohad Shamir, Nati Srebro, and Tong Zhang.
\newblock Communication-efficient distributed optimization using an approximate
  newton-type method.
\newblock In \emph{International conference on machine learning}, pages
  1000--1008, 2014.

\bibitem[{Shirish Keskar} et~al.(2016){Shirish Keskar}, {Mudigere}, {Nocedal},
  {Smelyanskiy}, and {Tang}]{keskar2016large}
N.~{Shirish Keskar}, D.~{Mudigere}, J.~{Nocedal}, M.~{Smelyanskiy}, and
  P.~T.~P. {Tang}.
\newblock {On Large-Batch Training for Deep Learning: Generalization Gap and
  Sharp Minima}.
\newblock \emph{ArXiv e-prints}, September 2016.

\bibitem[{Smith} et~al.(2016){Smith}, {Forte}, {Ma}, {Takac}, {Jordan}, and
  {Jaggi}]{smith2016cocoa}
V.~{Smith}, S.~{Forte}, C.~{Ma}, M.~{Takac}, M.~I. {Jordan}, and M.~{Jaggi}.
\newblock {CoCoA: A General Framework for Communication-Efficient Distributed
  Optimization}.
\newblock \emph{ArXiv e-prints}, November 2016.

\bibitem[{Stich}(2018)]{stich2018local}
S.~U. {Stich}.
\newblock {Local SGD Converges Fast and Communicates Little}.
\newblock \emph{ArXiv e-prints}, May 2018.

\bibitem[Tak{\'a}{\v{c}} et~al.(2013)Tak{\'a}{\v{c}}, Bijral, Richt{\'a}rik,
  and Srebro]{Tak_2013}
Martin Tak{\'a}{\v{c}}, Avleen Bijral, Peter Richt{\'a}rik, and Nathan Srebro.
\newblock Mini-batch primal and dual methods for svms.
\newblock In \emph{Proceedings of the 30th International Conference on
  International Conference on Machine Learning-Volume 28}, pages III--1022.
  JMLR. org, 2013.

\bibitem[Wangni et~al.(2017)Wangni, Wang, Liu, and Zhang]{wangni2017gradient}
Jianqiao Wangni, Jialei Wang, Ji~Liu, and Tong Zhang.
\newblock Gradient sparsification for communication-efficient distributed
  optimization.
\newblock \emph{CoRR}, abs/1710.09854, 2017.
\newblock URL \url{http://arxiv.org/abs/1710.09854}.

\bibitem[Wen et~al.(2017)Wen, Xu, Yan, Wu, Wang, Chen, and Li]{wen2017terngrad}
Wei Wen, Cong Xu, Feng Yan, Chunpeng Wu, Yandan Wang, Yiran Chen, and Hai Li.
\newblock Terngrad: Ternary gradients to reduce communication in distributed
  deep learning.
\newblock \emph{CoRR}, abs/1705.07878, 2017.
\newblock URL \url{http://arxiv.org/abs/1705.07878}.

\bibitem[{You} et~al.(2017){You}, {Gitman}, and {Ginsburg}]{you2017large}
Y.~{You}, I.~{Gitman}, and B.~{Ginsburg}.
\newblock {Large Batch Training of Convolutional Networks}.
\newblock \emph{ArXiv e-prints}, August 2017.

\bibitem[{Yu} et~al.(2018){Yu}, {Yang}, and {Zhu}]{yu_2018}
H.~{Yu}, S.~{Yang}, and S.~{Zhu}.
\newblock {Parallel Restarted SGD for Non-Convex Optimization with Faster
  Convergence and Less Communication}.
\newblock \emph{ArXiv e-prints}, July 2018.

\bibitem[Zhang et~al.(2016)Zhang, Li, Kara, Alistarh, Liu, and
  Zhang]{Zhazip_2016}
Hantian Zhang, Jerry Li, Kaan Kara, Dan Alistarh, Ji~Liu, and Ce~Zhang.
\newblock The zipml framework for training models with end-to-end low
  precision: The cans, the cannots, and a little bit of deep learning.
\newblock \emph{arXiv preprint arXiv:1611.05402}, 2016.

\bibitem[Zhang et~al.(2017)Zhang, Li, Kara, Alistarh, Liu, and
  Zhang]{zhang2017zipml}
Hantian Zhang, Jerry Li, Kaan Kara, Dan Alistarh, Ji~Liu, and Ce~Zhang.
\newblock {Z}ip{ML}: Training linear models with end-to-end low precision, and
  a little bit of deep learning, 06--11 Aug 2017.
\newblock URL \url{http://proceedings.mlr.press/v70/zhang17e.html}.

\bibitem[{Zhang} et~al.(2016){Zhang}, {De Sa}, {Mitliagkas}, and
  {R{\'e}}]{Zha_DeS_Re_2016}
J.~{Zhang}, C.~{De Sa}, I.~{Mitliagkas}, and C.~{R{\'e}}.
\newblock {Parallel SGD: When does averaging help?}
\newblock \emph{ArXiv e-prints}, June 2016.

\bibitem[{Zhang} et~al.(2014){Zhang}, {Choromanska}, and
  {LeCun}]{zhang2014deep}
S.~{Zhang}, A.~{Choromanska}, and Y.~{LeCun}.
\newblock {Deep learning with Elastic Averaging SGD}.
\newblock \emph{ArXiv e-prints}, December 2014.

\bibitem[{Zhang}(2004)]{Zha_2004}
T.~{Zhang}.
\newblock {Solving large scale linear prediction problems using stochastic
  gradient descent algorithms}.
\newblock \emph{Proceedings of the conference on machine learning (ICML)},
  2004.

\bibitem[{Zhang} and {Xiao}(2015)]{zhang2015communication}
Y.~{Zhang} and L.~{Xiao}.
\newblock {Communication-Efficient Distributed Optimization of Self-Concordant
  Empirical Loss}.
\newblock \emph{ArXiv e-prints}, January 2015.

\bibitem[Zhang et~al.(2012)Zhang, Wainwright, and
  Duchi]{zhang2012communication}
Yuchen Zhang, Martin~J Wainwright, and John~C Duchi.
\newblock Communication-efficient algorithms for statistical optimization.
\newblock In \emph{Advances in Neural Information Processing Systems}, pages
  1502--1510, 2012.

\bibitem[Zhang et~al.(2013)Zhang, Duchi, Jordan, and
  Wainwright]{zhang2013information}
Yuchen Zhang, John Duchi, Michael~I Jordan, and Martin~J Wainwright.
\newblock Information-theoretic lower bounds for distributed statistical
  estimation with communication constraints.
\newblock In C.~J.~C. Burges, L.~Bottou, M.~Welling, Z.~Ghahramani, and K.~Q.
  Weinberger, editors, \emph{Advances in Neural Information Processing Systems
  26}, pages 2328--2336. Curran Associates, Inc., 2013.

\bibitem[Zhao and Zhang(2015)]{Zha_Zha_2015}
Peilin Zhao and Tong Zhang.
\newblock Stochastic optimization with importance sampling for regularized loss
  minimization.
\newblock In \emph{International Conference on Machine Learning (ICML)}, pages
  1--9, 2015.

\bibitem[{Zhao} and {Li}(2015)]{zhao2015fast}
S.-Y. {Zhao} and W.-J. {Li}.
\newblock {Fast Asynchronous Parallel Stochastic Gradient Decent}.
\newblock \emph{ArXiv e-prints}, August 2015.

\bibitem[Zhao and Li(2016)]{zhao2016fast}
Shen-Yi Zhao and Wu-Jun Li.
\newblock Fast asynchronous parallel stochastic gradient descent: A lock-free
  approach with convergence guarantee.
\newblock In \emph{Proceedings of the Thirtieth AAAI Conference on Artificial
  Intelligence}, AAAI'16, pages 2379--2385. AAAI Press, 2016.
\newblock URL \url{http://dl.acm.org/citation.cfm?id=3016100.3016231}.

\bibitem[Zhu and Marcotte(1996)]{Zhu_Mar_1995}
Dao~Li Zhu and Patrice Marcotte.
\newblock Co-coercivity and its role in the convergence of iterative schemes
  for solving variational inequalities.
\newblock \emph{SIAM Journal on Optimization}, 6\penalty0 (3):\penalty0
  714--726, 1996.

\bibitem[Zhuang et~al.(2013)Zhuang, Chin, Juan, and Lin]{zhuang2013fast}
Yong Zhuang, Wei-Sheng Chin, Yu-Chin Juan, and Chih-Jen Lin.
\newblock A fast parallel sgd for matrix factorization in shared memory
  systems.
\newblock In \emph{Proceedings of the 7th ACM Conference on Recommender
  Systems}, RecSys '13, pages 249--256, New York, NY, USA, 2013. ACM.
\newblock ISBN 978-1-4503-2409-0.
\newblock \doi{10.1145/2507157.2507164}.
\newblock URL \url{http://doi.acm.org/10.1145/2507157.2507164}.

\bibitem[Zinkevich et~al.(2010)Zinkevich, Weimer, Li, and Smola]{Zin_2010}
Martin Zinkevich, Markus Weimer, Lihong Li, and Alex~J Smola.
\newblock Parallelized stochastic gradient descent.
\newblock In \emph{Advances in neural information processing systems}, pages
  2595--2603, 2010.

\end{thebibliography}

\newpage
\appendix
\begin{center}
	{\Large{\textbf{Communication trade-offs for synchronized distributed SGD with large step size\\ \ \\
	{\large SUPPLEMENTARY MATERIAL}}}}
\end{center}

\setcounter{equation}{0}
\setcounter{figure}{0}
\setcounter{table}{0}
\setcounter{page}{1}
\renewcommand{\theequation}{S\arabic{equation}}
\renewcommand{\thefigure}{S\arabic{figure}}
\renewcommand{\thetheorem}{S\arabic{theorem}}
\renewcommand{\thelemma}{S\arabic{lemma}}
\renewcommand{\theproposition}{S\arabic{proposition}}
\renewcommand{\thecorollary}{S\arabic{proposition}}
\renewcommand{\thetable}{S\arabic{table}}

\ \\
In this Appendix, we give the proofs of our main results, and auxiliary elements. In \Cref{sec:experiments}, we describe the experimental evaluations that illustrate the behavior of the different processes.  In ~\Cref{app:material} we provide some additional material (Tables, interpretations, etc.) which may help the reader navigate through our results. In~\Cref{app:auxiliary}, we prove contraction Lemmas for $ \expeLigne{\normLigne[2]{\iter{p}{k}{t}-\ws}} $. In ~\Cref{app:aux4}, we prove similar guarantees for moment of order 4. In \Cref{app:mainproofs}, we give the proof of the main results on $ \normLigne[2]{F''(\ws)( \itePRav-ws)} $ for mini-batch, one-shot averaging,  and Local-SGD in the Finite Horizon setting. In \Cref{sec:proofsonline} we give similar results in the online setting (for decaying step size). Finally, we provide a brief survey of distributed optimization techniques in \Cref{app:relatedwork}. 

%

\addtocontents{toc}{\protect\setcounter{tocdepth}{1}}
\setcounter{tocdepth}{1}
\tableofcontents

\section{Experimental results}\label{sec:experiments}

\begin{table}[H]
	\centering
	\caption{Data-sets for experimentation.  \label{tab:exp_data}}
	\resizebox{\linewidth}{!}{%
		\begin{tabular}{|c|c|c|c|c|}
			\hline
			\textbf{Name of the Data-set} & \textbf{Task} & \textbf{Algorithm} & \textbf{Number of Samples} & \textbf{Number of Features}\\
			\hline
			Epsilon & Classification & Logistic & 400000 & 2000 \\
			Year Prediction MSD & Regression  & Least-Squares & 463715 & 90\\
			CPU Stall & Regression & Least-Squares & 8192 & 12\\
			\hline
		\end{tabular}%
	}
\end{table}

\begin{figure}[!htb]
	\begin{minipage}{0.48\textwidth}
		\centering
		\includegraphics[width=\linewidth]{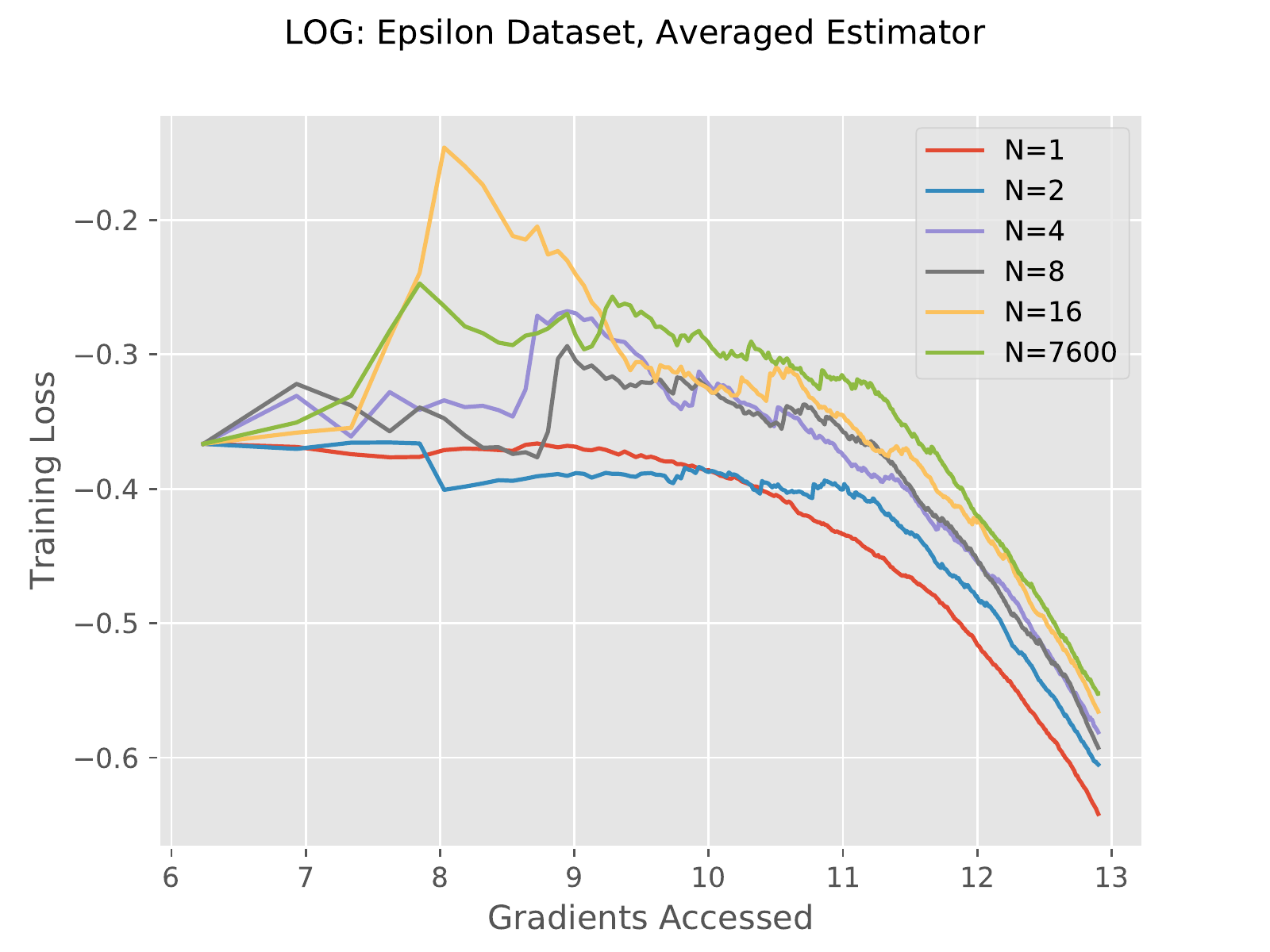}
	\end{minipage}\hfill
	\begin{minipage}{0.48\textwidth}
		\centering
		\includegraphics[width=\linewidth]{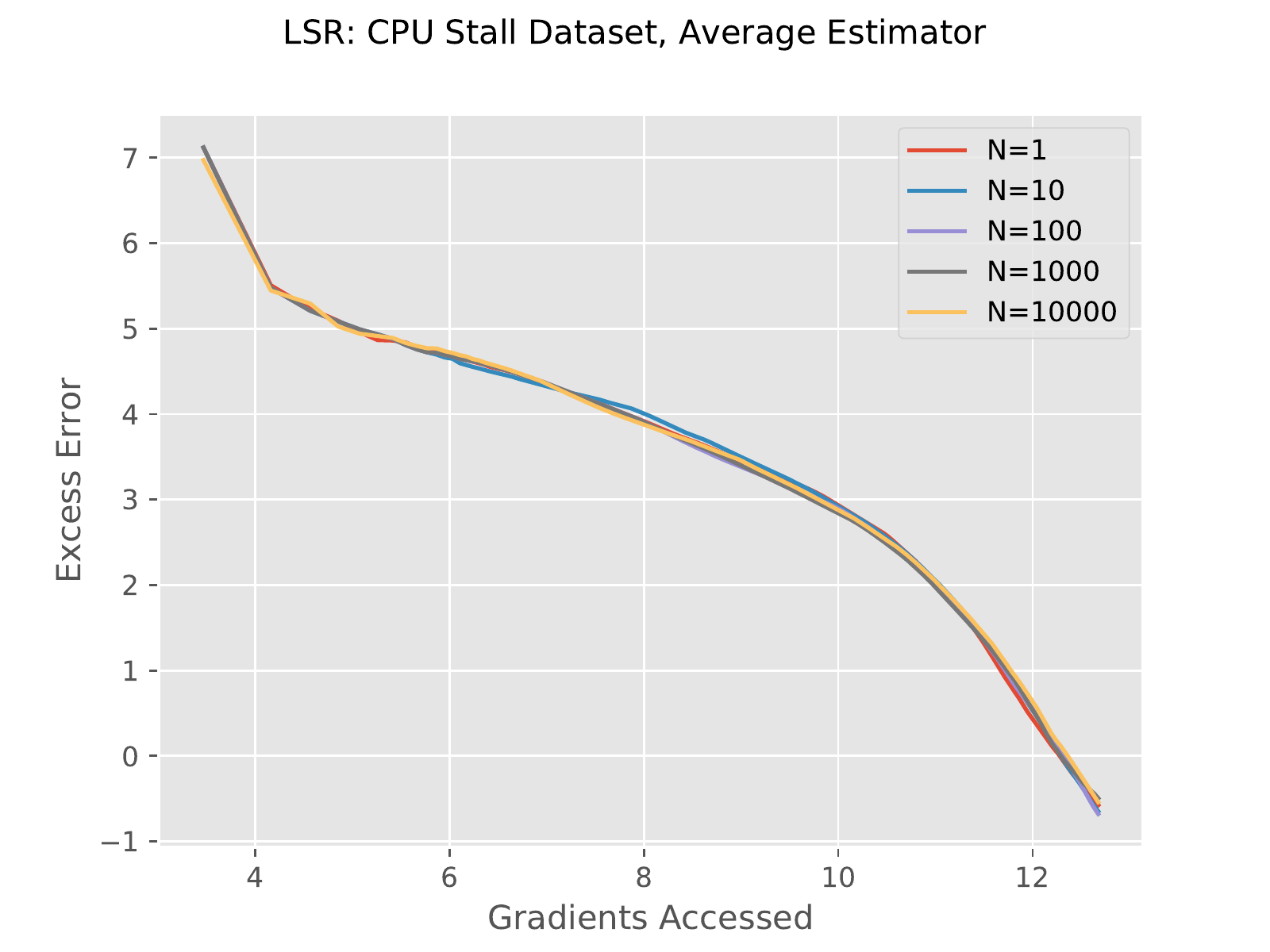}
	\end{minipage}
	\caption{Performance of Local SGD \label{Fig:gen_error}}
\end{figure}
We perform experiments for three different data-sets\footnote{\href{https://www.csie.ntu.edu.tw/~cjlin/libsvmtools/datasets/}{https://www.csie.ntu.edu.tw/~cjlin/libsvmtools/datasets/}}, two for least-squares regression and one for logistic regression \Cref{tab:exp_data}. For all the curves we use $\log (y)$ v/s $\log (x)$ plots unless explicitly mentioned. Moreover, to elucidate the theory we use the same learning rates for all the algorithms in an experiment. The number of workers is set to $P=32$ every where, and plots are labeled w.r.t. the number of local steps $N$ which we don't change along the different phases. We do the following experiments:

\begin{figure}[!htb]
	\begin{minipage}{0.48\textwidth}
		\centering
		\includegraphics[width=\linewidth]{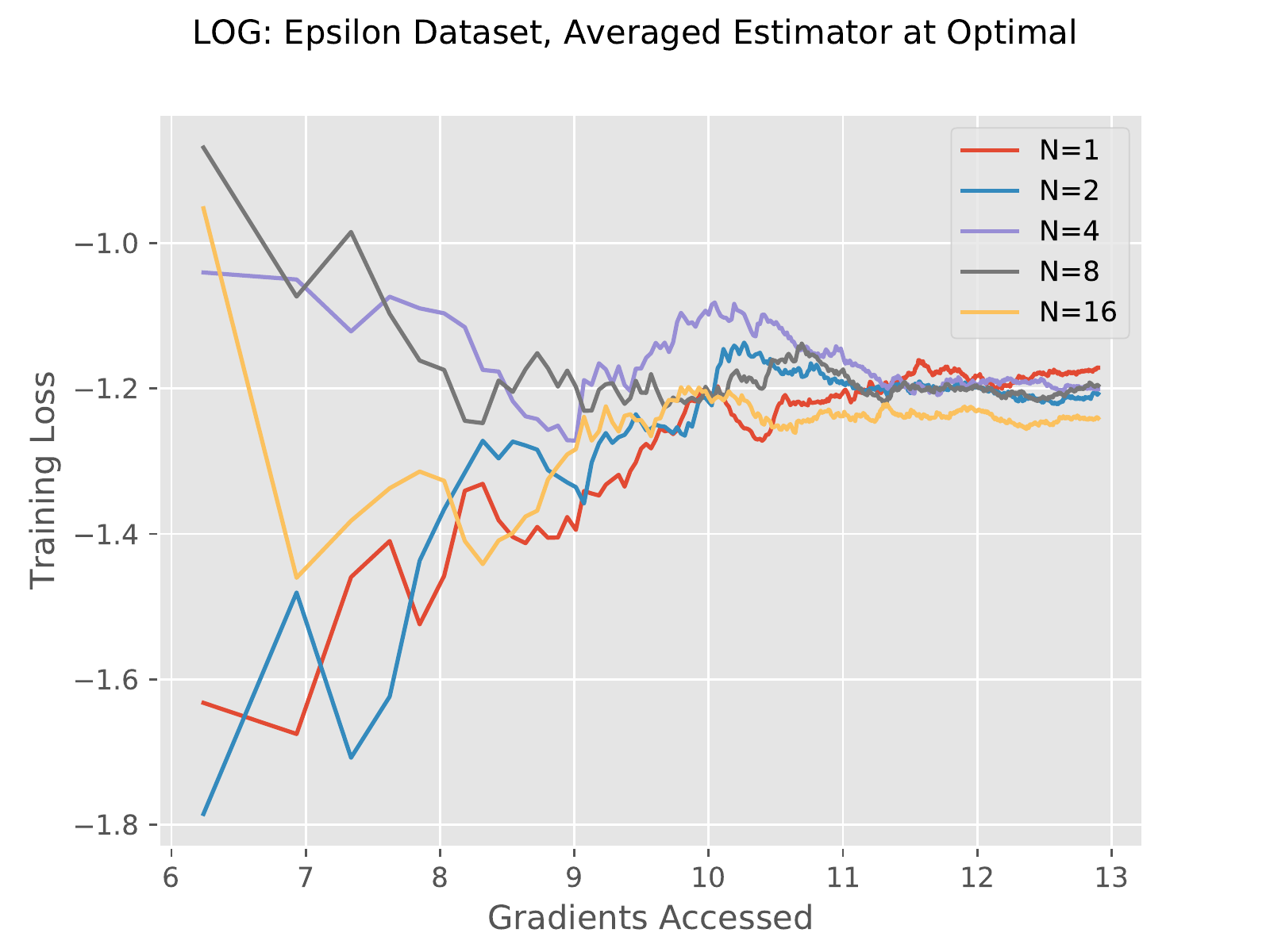}
		
	\end{minipage}\hfill
	\begin{minipage}{0.48\textwidth}
		\centering
		\includegraphics[width=\linewidth]{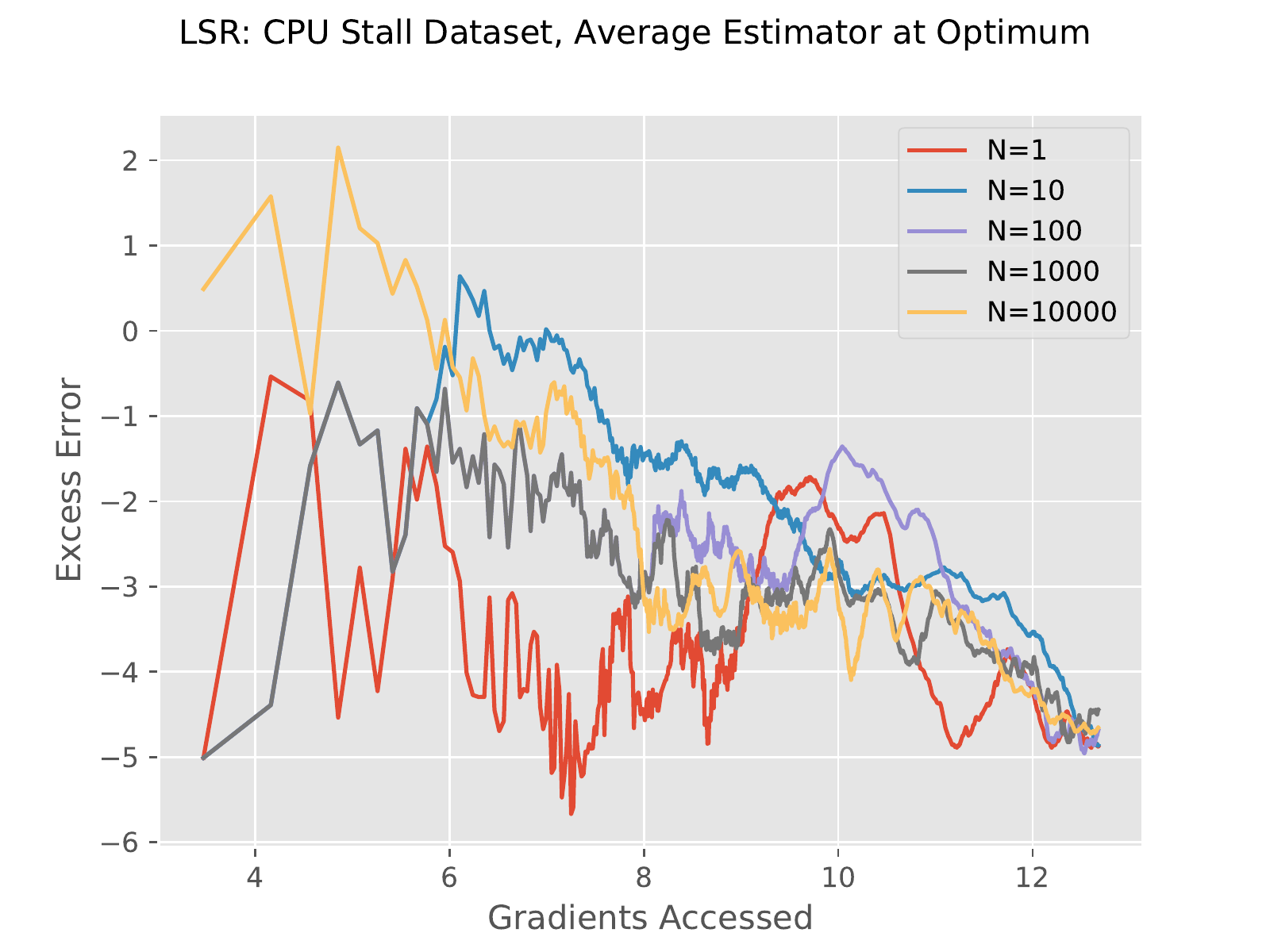}
	\end{minipage}
	\caption{Performance of Local SGD at the optimal \label{Fig:opt_error}}
\end{figure}

\begin{figure}[!htb]
	\begin{minipage}{0.48\textwidth}
		\centering
		\includegraphics[width=\linewidth]{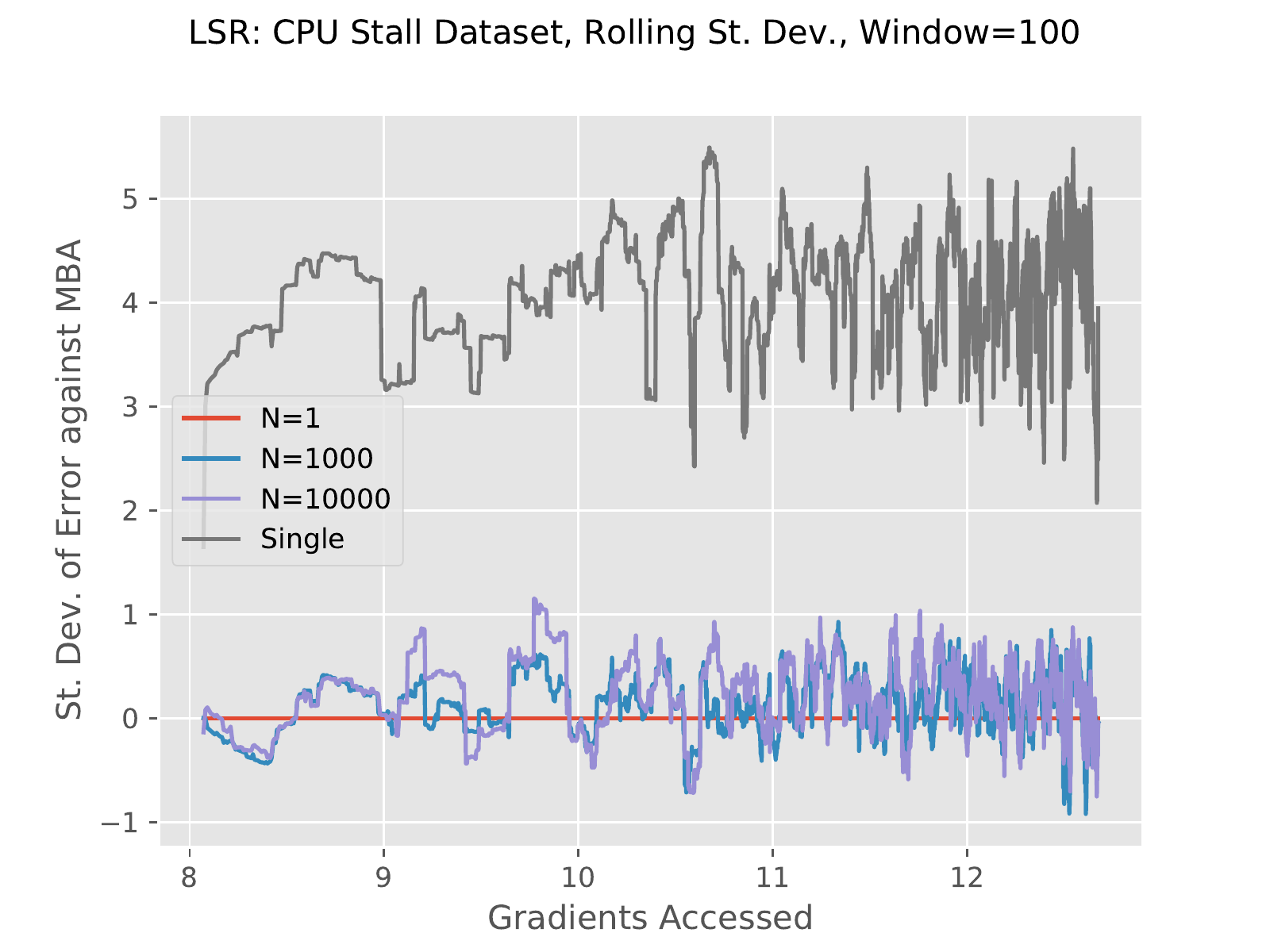}
	\end{minipage}\hfill
	\begin{minipage}{0.48\textwidth}
		\centering
		\includegraphics[width=\linewidth]{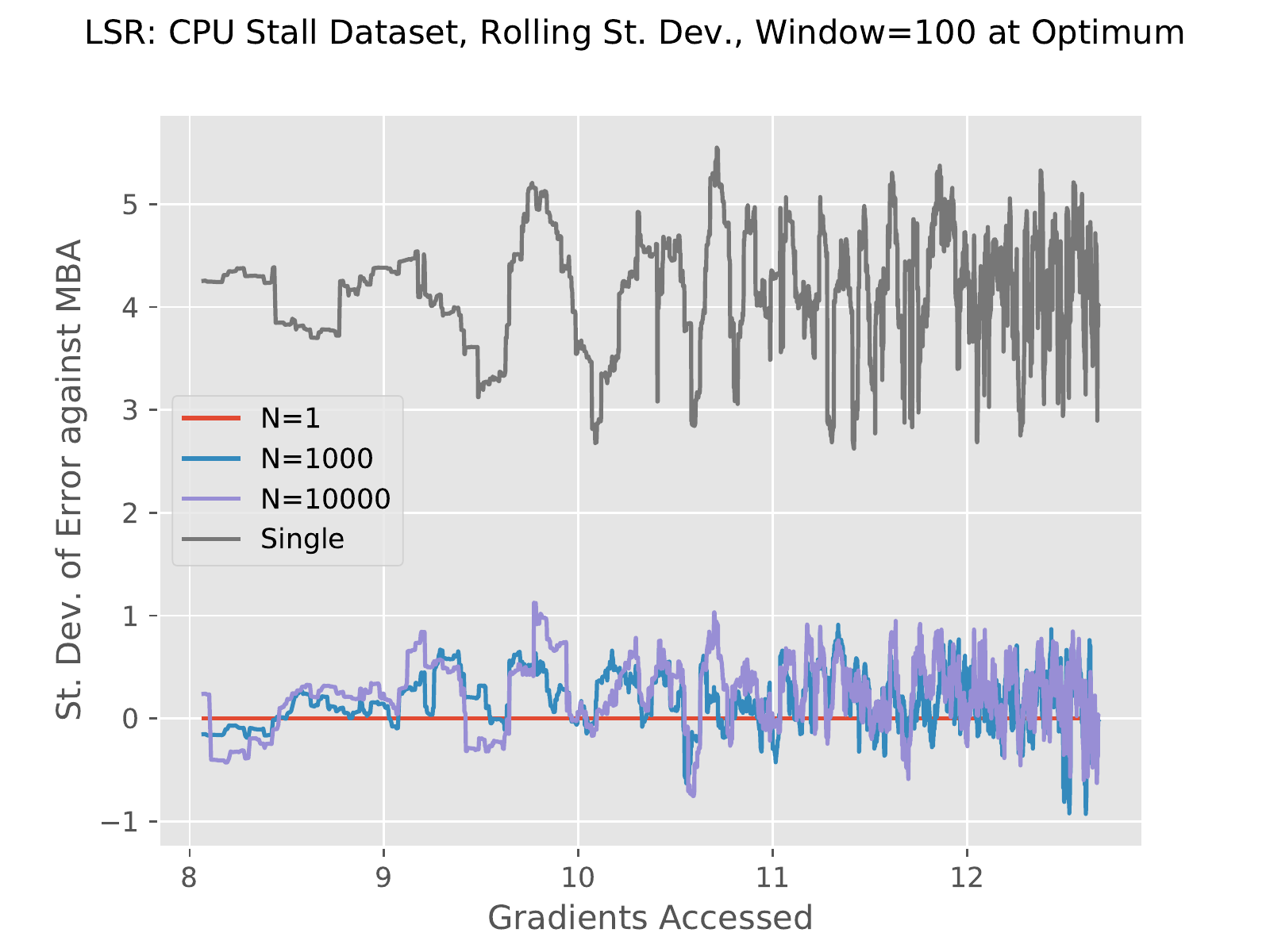}
	\end{minipage}
	\caption{Variance of the loss function compared to MBA \label{Fig:Var}}
\end{figure}

\begin{figure}[!htb]
	\begin{center} 
		\includegraphics[scale=0.5]{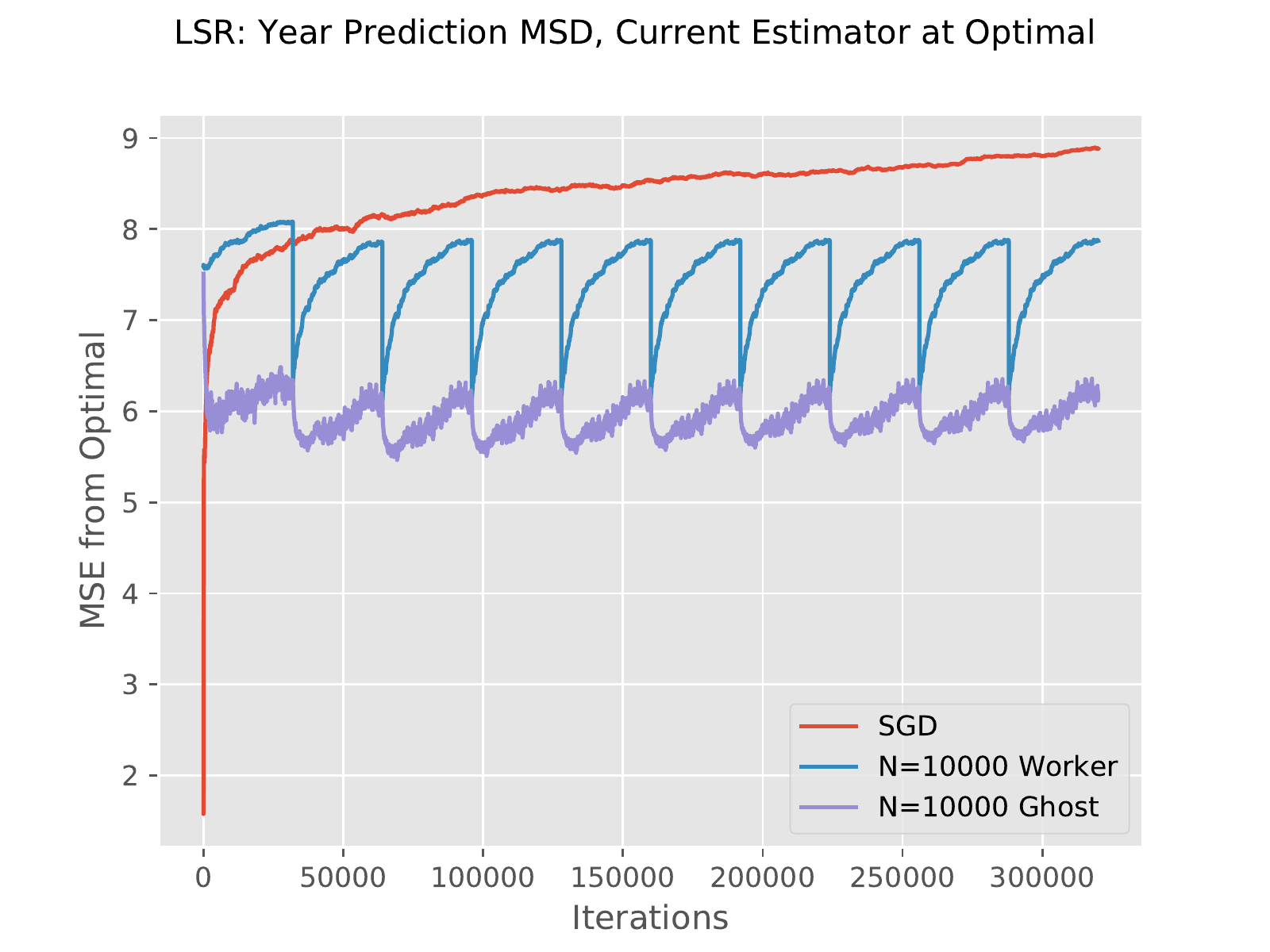}
		\caption{Iterate Convergence of a single process against SGD and the ghost process. \label{Fig:worker}}
	\end{center}
\end{figure}

\begin{enumerate}
	\item Performance of local SGD with different number of local steps spanning OSA to MBA (\Cref{Fig:gen_error}). We globally find MBA to perform the best. Besides, as we increase the number of local steps $N$ the performance gets closer to OSA. This observation aligns with our theoretical guarantees. We use the averaged iterate (i.e., $\overline{\overline{\ww}}$, the average over all the iterates till that point) for reporting the performance. The current iterate (i.e., $\breve{\ww}^t_k$, the ghost iterate for the current iteration) is omitted as the graphs are too noisy to be interpreted, and a variance of the loss is used instead.    
	\item Performance of local SGD with different number of local steps when started at the optimal point (\Cref{Fig:opt_error}). We expect that if we start at $\ws$ then the bias term goes to zero and the difference between the algorithms becomes sharper. This is because our results predict that for constant learning rate, the initial conditions are forgotten at the same rate. We see that mini-batch outperforms OSA no the first iterations, but not asymptotically.
	\item Variance of the estimators, for loss (\Cref{Fig:Var}) and iterate values (\Cref{Fig:worker}). We expect that a larger mini-batch size predicts a lower variance for these cases, and we observe the same through our experiments. In fact, the mean squared error of the parameters at the optimal  is observed to be following a periodic curve. The value on an individual worker rises until it communicates, but always remains lower than a single SGD process run for the same number of iterations. This, verifies our theory and results for iterate convergence. Moreover, the variance at the loss function follows a similar pattern which elucidates the fact the intuitions developed in the paper also hold for functional convergence.     
\end{enumerate}

\section{Some Additional Material}\label{app:material}
\subsection{Pseudo codes}
Pseudo codes of both algorithms are given in \Cref{fig:pseudocodes}.

\begin{figure}[H]
	\begin{minipage}[t]{7cm}
		\caption{Vanilla-SGD}
		\vspace{-0.2in}
		\label{alg:SGD}  
		\begin{algorithm}[H]
			\KwIn{$F:\R^d\rightarrow\R$}
			$\ites_0\gets \textbf{Initialize}$\\
			\For{t = 0,1,2,...T}{
				$\gg_t(\ites_{t-1})\gets \textbf{SFO}(F,\ites_{t-1})$\\
				$\ites_{t}\gets \ites_{t-1}-\eta_t\gg_t (\ites_{t-1})$
			}
			\KwOut{$S(\ites_{0}, \ites_{1},..,\ites_{T-1},\ites_{T})\in\R^d$}
		\end{algorithm}
	\end{minipage}
	\hspace{1em}
	\begin{minipage}[t]{7cm}
		\caption{Local-SGD}
		\vspace{-0.2in}
		\label{alg:locSGD}
		\begin{algorithm}[H]
			\KwIn{$F:\R^d\rightarrow\R$}
			$\hat{\ww}^0=\ww^0 \gets \textbf{Initialize}$\\
			\For{$t=1,2,...C $}{
				\ForPar{i=1,2,...P}{
					$\ww_{i,0}^t\gets \hat{\ww}^{t-1}$\\
					\For{k=0,1,2,...$N^t$}{
						$\gg_{i,k}^t(\ww_{i,k-1}^t)\gets \textbf{SFO}(F,\ww_{i,k-1}^t)$
						$\ww_{i,k}^t\gets\ww_{i,k-1}^t-\eta_{k}^t\gg_{i,k}^t(\ww_{i,k-1}^t)$
					}
					$\overline{\ww}_i^t \gets \frac{1}{N_t}\sum_{k=1}^{N_t}\ww_{i,k}^t$    
				}
				$\overline{\ww}^t \gets \frac{1}{P}\sum_{i=1}^{P}\overline{\ww}_i^t$
				$\hat{\ww}^t \gets \frac{1}{p}\sum_{i=1}^{P}\ww_{i,N_t}^t$
			}
			\KwOut{$\overline{\overline{\ww}}^T=\frac{1}{C}\sum_{t=1}^{C}\overline{\ww}^{t}\in\R^d$}
		\end{algorithm}
		
	\end{minipage}
	\caption{Serial and parallel SGD algorithms. \textbf{SFO} stands for the stochastic first order oracle. Note that every node has access to the full function i.e., the data is not distributed across nodes.}\label{fig:pseudocodes}
\end{figure}

\subsection{Summary of Results}In the table below, we specify for which algorithm our results apply (mini batch, one shot, or local SGD), under which assumptions they are proved and if they apply to the on-line setting(OL) or just the finite horizon(FH) case. \\
\begin{table}[ht]
	\begin{center}
		\begin{tabular}{ccccccccccc}
			\hline
			& \multicolumn{1}{c}{} & \multicolumn{7}{c}{Assumptions} & \multicolumn{2}{c}{Setting}\\
			\cmidrule(lr){3-9}\cmidrule(lr){10-11}
			Proposition & Algorithm & \Cref{hyp:strong_convex} & \Cref{hyp:regularity} & \Cref{hyp:quad} & \Cref{ass:def_filt} & \Cref{hyp:addit_noise} & \Cref{ass:lip_noisy_gradient_AS} & \Cref{hyp:variancebound}&FH & OL\\
			\hline 
			\Cref{prop:conver_Mini_Batch}       &Mini-Batch               &$\cm$&$\cm$&          &$\cm$&          &$\cm$&$\cm$&\cm&\\
			\Cref{prop:conver_One_Shot}         &One-shot averaging&$\cm$&$\cm$&          &$\cm$&         &$\cm$&$\cm$&\cm&          \\
			\Cref{prop:conver_Mini_Batch_DSS}       &Mini-Batch       \&OS        &$\cm$&$\cm$&          &$\cm$&          &$\cm$&$\cm$&&\cm\\
			\hline
			\Cref{prop:conv_quad_Simple}        &Local SGD               &         &         &$\cm$ &$\cm$&$\cm$&         &&\cm&           \\
			\Cref{prop:conv_quad}                   &Local SGD               &          &         &$\cm$&$\cm$&         &$\cm$&$\cm$&\cm&\cm\\
			\Cref{prop:conv_general_function} &Local SGD               &$\cm$&$\cm$&          &$\cm$&$\cm$&         &     &\cm&\cm\\
			\Cref{th:LocalSGDconv} 1. &Local SGD               &&&$\cm$          &$\cm$&&  $\cm$       & $\cm$    &$\cm$&\\
			\Cref{th:LocalSGDconv} 2. &Local SGD               &$\cm$&$\cm$&          &$\cm$&$\cm$&         &     &$\cm$&\\
			\hline
		\end{tabular}
	\end{center}
	\vspace{0.5em}
	\caption{Summary of results}\label{tab:sum}
\end{table}

\subsection{Example: Learning from \iid~observations}
\label{ex:iid_observation_learning}
Our main motivation comes
from machine learning; consider two sets $\mathcal{X}, \mathcal{Y}$
and a convex loss function
$\loss: \mathcal{X}\times \mathcal{Y} \times \rset^{d} \to \rset$. The generalization error is defined as
$F_\loss (\ite ) = \E_{X,Y} [\loss(X, Y, \ite) ]$, where $(X,Y)$ are
some random variables. Given \iid~observations
$ (X_k, Y_k)_{k \in \nsets } $ with the same distribution as $(X,Y)$,
for any $k \in \nsets$, we define $f_k(\cdot) = \loss(X_k,Y_k, \cdot)$
the loss with respect to observation $ k $. SGD can be used in two contexts:
\begin{enumerate} 
	\item \emph{Stochastic Approximation}: We use \emph{independent} observations at each iteration. The total number of iterations is thus at most the number of observations we access. SGD then corresponds to following the gradient of the loss  $f_{k}$ on a single independent observation $(X_k,Y_k)$. As the gradients we use are then unbiased gradients of the generalization error, this means that SGD directly minimizes this (unknown) function.
	\item \emph{Empirical Risk Minimization}: We define the empirical risk as  $\hat F_\loss (\ite ) = N^{-1} \sum_{k=1}^{N} [\loss(X_k, Y_k, \ite) ]$. At each step $ t $, we sample an index $ i_t $ \emph{uniformly on $ \un{N} $}, and use the gradient of the loss  $f_{i_t}$. Here the number of iterations is not limited, but the algorithm will converge to the minimum of the empirical risk. 
\end{enumerate}
In practice, this means that in the first situation, we want to optimize the precision of the algorithm for a limited number of oracle calls, while in the second situation one would rather optimize the number of outer iterations of the algorithm (\ie~its running time). In both these assumptions, Assumption~\Cref{ass:def_filt} is satisfied for the filtration generated by all the observations before time $ (t,k)$ (respectively all the indices sampled before time  $ (t,k)$).

Two typical situations regarding loss functions are worth mentioning. On the first hand,  in \emph{least-squares regression},  $\mathcal{X} =\rset^{d}$, $\mathcal{Y}=\R$, and the loss function is $\loss(X,Y, \ite) = (\langle X, \ite\rangle -Y)^{2}$. Then $F_\Sigma$ is the quadratic function
$\ite \mapsto \norm{\Sigma^{1/2} (\ite -\ws)}^2/2$, with  $\Sigma= \expeLine{XX^{\top}}$, which satisfies Assumption~{\Cref{hyp:quad}}. For any $\ite \in \rset^d$,
\begin{equation}
\label{eq:eps_reg_line}
f'_k(\ite) - F_\Sigma'(\ite) =  (X_k X_k^{\top}-\Sigma) (\ite-\ws) - (X_k^{\top} \ws - Y_k)X_k
\end{equation}
Then,  Assumption~{\Cref{ass:lip_noisy_gradient_AS}} and \Cref{hyp:variancebound}  are satisfied, if $ X $ is bounded and $ Y $ has finite variance. 

On the other hand, in \emph{logistic regression}, where
$\loss(X,Y, \ite) = \log (1+ \exp(-Y\langle X, \ite\rangle ))$. Assumptions {\Cref{hyp:regularity}} and  {\Cref{hyp:addit_noise}}  are then satisfied \cite{Bac_2014}, as  is
Assumption \Cref{hyp:strong_convex} under an additional restriction to a compact set or if an extra regularization is added.

SGD for least squares regression typically satisfies \Cref{hyp:quad}, \Cref{ass:def_filt}, \Cref{ass:lip_noisy_gradient_AS} and \Cref{hyp:variancebound}. On the other hand, SGD for logistic regression satisfies \Cref{hyp:strong_convex}, {\Cref{hyp:regularity}}, \Cref{ass:def_filt} and {\Cref{hyp:addit_noise}}.

\section{Convergence guaranties for the second order moment.} \label{app:auxiliary}
In this section, we prove several Lemmas that allow to control the second order moment for the iterate.
We first recall a few useful inequalities that will be used in the following. See for example~\cite{Nes_2004}.

If $ F $ is convex and smooth (\eg~satisfies~\Cref{hyp:regularity}), the gradient of $ F $ is cocoercive, thus for any $ \ite \in \rd $:
\begin{equation}\label{eq:gradcoco}
L \ps{F'(\ite)}{\ite-\ws} \geq  \mynorm{F'(\ite)}.
\end{equation}
If the function is strongly-convex (Assumption~\Cref{hyp:strong_convex}), then for any $ \ite \in \rd $:
\begin{equation}\label{eq:strong_conv}
\ps{F'(\ite)}{\ite-\ws} \geq \mu \mynorm[2]{\ite-\ws}.
\end{equation}

\subsection{Inner iteration Lemma}\label{sec:innercontro}
We first recall the proof of the convergence for inner iterates. This proof corresponds to what happens on one machine, and can be found in the literature~\cite{Bac_Mou_2011, Die_Dur_2018} for example. 

For any $ p\in \un{P},t\in\un{C},  k\in \un{\Nt}$, under Assumptions~\Cref{hyp:strong_convex},~\Cref{hyp:regularity},~\Cref{ass:def_filt},~\Cref{ass:lip_noisy_gradient_AS},~\Cref{hyp:variancebound}, we have
\begin{align}
\expe{\mynorm[2]{\iter{p}{k}{t}-\ws} } &\le \expe{\mynorm[2]{\iter{p}{k-1}{t}-\ws} } - \eta_k^t  \ps{F'(\iter{p}{k-1}{t})}{\iter{p}{k-1}{t}-\ws}+ 2 (\eta_k^t)^{2} \sigma^{2} \label{eq:bound_inner_product}\\
\expe{\mynorm[2]{\iter{p}{k}{t}-\ws} } &\le (1-\eta_k^t \mu ) \expe{\mynorm[2]{\iter{p}{k-1}{t}-\ws} } + 2 \eta_k^t \sigma^{2}. \nonumber
\end{align}
Using the second equation recursively results in:
\begin{align}
\expe{\mynorm[2]{\iter{p}{k}{t}-\ws} } &\le \prod_{m=1}^{k}(1-\eta_m^t \mu ) \expe{\mynorm[2]{\iter{p}{0}{t}-\ws} } + 2\sigma^{2} \sum_{m=1}^{k} (\eta_m^t)^{2} \prod_{l=m+1}^{k}(1-\eta_l^t \mu ) . \label{eq:boundinnerproc}
\end{align}

More precisely, for precise reference in the following proofs, we referenced this inequality with the following specific cases:

\begin{lemma}\label{lem:MB_bound}
	Under Assumptions~\Cref{hyp:strong_convex},~\Cref{hyp:regularity},~\Cref{ass:def_filt},~\Cref{ass:lip_noisy_gradient_AS},~\Cref{hyp:variancebound}, for mini-batch SGD with batch-size $P$ and step-size $\eta$ we have,
	\begin{align*}
	\e{\norm{\ww^t_{MB} -\ww^\star}^2} \leq \prod_{m=1}^{t}(1-\mu\eta)\e{\norm{\ww^0 -\ww^\star}^2} + \frac{2\sigma^2\eta^2}{P}\sum_{m=1}^{t}\prod_{l=m+1}^{t}(1-\mu\eta).
	\end{align*}
\end{lemma}

Such a result on reduced variance for mini-batch SGD ($\frac{\sigma^2}{P}$) can be found in many previous works like \cite{dekel2012optimal}. Since mini-batch SGD is trivial to parallelize, this result also holds for the averaged iterate for outer iteration $t$ while using mini-batch averaging. Similarly, for decaying step sizes,

\begin{lemma}\label{lem:MB_bound_DSS}
	Under Assumptions~\Cref{hyp:strong_convex},~\Cref{hyp:regularity},~\Cref{ass:def_filt},~\Cref{ass:lip_noisy_gradient_AS},~\Cref{hyp:variancebound}, and $\tilde{\eta}_{t}=\frac{c_\eta}{t^\alpha}$ for mini-batch SGD, for any $t \in \un{C} $ we have,
	\begin{align*}
	\e{\norm{\ww^t_{MB} -\ww^\star}^2} \leq \prod_{m=1}^{t}(1-\mu\tilde{\eta}_{m})\norm{\ww^0 -\ww^\star}^2 + 2\sigma^2\frac{1}{P}\sum_{m=1}^{t}(\tilde{\eta}_{m})^2\prod_{l=m+1}^{t}(1-\mu\tilde{\eta}_{l}).
	\end{align*}
\end{lemma} 

Similarly, in the case of one-shot averaging,
\begin{lemma}\label{lem:OS_bound}
	Under Assumptions~\Cref{hyp:strong_convex},~\Cref{hyp:regularity},~\Cref{ass:def_filt},~\Cref{ass:lip_noisy_gradient_AS},~\Cref{hyp:variancebound} and  a constant step-size $\eta$ using one-shot averaging, for any $K\in \un{N^1}$ and $i \in \un{P}$ we have,
	\begin{align*}
	\e{\norm{\ww_{i,K}^1-\ww^\star}^2} \leq \prod_{m=1}^{K}(1-\mu\eta) \norm{\ww^0-\ww^\star}^2 + 2\sigma^2\eta^2\sum_{m=1}^{K}\prod_{l=m+1}^{K}(1-\mu\eta_l^1).
	\end{align*}
\end{lemma}

\begin{lemma}\label{lem:OS_bound_DSS}
	Under Assumptions~\Cref{hyp:strong_convex},~\Cref{hyp:regularity},~\Cref{ass:def_filt},~\Cref{ass:lip_noisy_gradient_AS},~\Cref{hyp:variancebound}, and $\eta^1_k=\tilde{\eta}_{k}=\frac{c_\eta}{k^\alpha}$ using one-shot averaging for any $K\in \un{N^1}$ and $i \in \un{P}$ we have,
	\begin{align*}
	\e{\norm{\ww_{i,K}^1-\ww^\star}^2} \leq \prod_{m=1}^{K}(1-\mu\eta_m^1) \norm{\ww^0-\ww^\star}^2 + 2\sigma^2\sum_{m=1}^{K}(\eta_m^1)^2\prod_{l=m+1}^{K}(1-\mu\eta_l^1).
	\end{align*}
\end{lemma}

\subsection{Proof of \Cref{prop:conv_quad_Simple}}\label{sec:proofquadsimple}

In this Section we prove \Cref{prop:conv_quad_Simple}. In order to provide a bound on the mean squared distance to the optimum of the outer iterates, we introduce a \emph{ghost} sequence \cite{mania2015perturbed}, \ie, a sequence of iterates which is not actually computed. For any $ t\in \un{C}, k\in \un{\Nt} $, we define 
\begin{equation}\label{eq:def_ghostit}
\breve{\ite}^{t}_k:= \frac{1}{P}\sum_{i=1}^{P} \iter{i}{k}{t}. 
\end{equation}

We prove the following Lemma:
\begin{lemma}\label{lem:auxPropquadsimple}
	Under Assumptions~\Cref{hyp:quad}, \Cref{ass:def_filt} and \Cref{hyp:addit_noise}, for any $t\in \un{C}, K\in \un{\Nt} $, we have:
	\begin{equation}\label{key}
	\expe{\norm{\breve{\ww}_{K}^t-\ww^\star}^2} \leq \prod_{m=1}^{K}(1 -\mu \eta_{m}^t) \norm{\breve{\ww}_{0}^t-\ww^\star}^2 +\frac{  \siginf}{P} \sum_{m=1}^{K}  (\eta_{m}^t)^{2} \prod_{l=m+1}^{K}(1 -\mu \eta_{l}^t) \eqsp. 
	\end{equation}
\end{lemma}
Remarking that for any $ t\in\un{C} $, $ \breve{\ite}^{t}_{\Nt} =\hat \ite^t$ this implies the \emph{first inequality} of \Cref{prop:conv_quad_Simple}. Note that this Lemma is valid for both decaying steps and and a constant learning rate. Especially, for a constant step size $ \eta $, and $ K=\Nt $:
\begin{equation*}\label{key}
\expe{\norm{\hat{\ww}^t-\ww^\star}^2} \leq (1 -\mu \eta)^{\Nt} \norm{\hat{\ww}^{t-1}-\ww^\star}^2 +\frac{  \siginf}{P} \eta \frac{1-(1 -\mu \eta)^{\Nt}}{\mu} \eqsp. 
\end{equation*}
More generally, we also have the following corollary, if we denote $ (\tilde \eta_k)_{k\geq0} $ the sequence such that $ \eta_k^t= \tilde{\eta}_{\lbrace\sum_{t'=1}^{t-1}\Ntp+ k\rbrace} $ (this just corresponds to re-indexing the sequence):
\begin{corollary}\label{cor:cor1}
	Under Assumptions~\Cref{hyp:quad}, \Cref{ass:def_filt} and \Cref{hyp:addit_noise}, for any $T\in \un{C} $, we have:
	\begin{equation}\label{key}
	\expe{\norm{\hat{\ww}^T-\ww^\star}^2} \leq \prod_{k=1}^{\sum_{t=1}^{T}\Nt}(1 -\mu \tilde\eta_{k}) \norm{{\ww}_{0}-\ww^\star}^2 +\frac{  \siginf}{P} \sum_{t=1}^{\sum_{t=1}^{T}\Nt} \tilde \eta_{k}^{2} \prod_{j=k+1}^{\sum_{t=1}^{T}\Nt}(1 -\mu \tilde \eta_{j}) \eqsp. 
	\end{equation}
\end{corollary}
\begin{proof}[Proof of \Cref{cor:cor1}]
	By induction, \Cref{lem:auxPropquadsimple} implies that for any $T\in \un{C} $
	\begin{equation}\label{eq:cor1}
	\expe{\norm{\hat{\ww}^T-\ww^\star}^2} \leq \prod_{t=1}^{T}\prod_{k=1}^{\Nt}(1 -\mu \eta_{k}^t) \norm{{\ww}_{0}-\ww^\star}^2 +\frac{  \siginf}{P} \sum_{t=1}^{T}  \prod_{t'=t+1}^{T} \prod_{k=1}^{\Ntp}(1 -\mu \eta_{k}^t) \sum_{k=1}^{\Nt}  (\eta_{k}^t)^{2} \prod_{j=k+1}^{\Nt}(1 -\mu \eta_{j}^t) \eqsp. 
	\end{equation}
	Then using $ \eta_k^t= \tilde{\eta}_{\lbrace\sum_{t'=1}^{t-1}\Ntp+ k\rbrace}  $, the corollary is just re-writing of \Cref{eq:cor1}.
	
\end{proof} 
To prove the second inequality of \Cref{prop:conv_quad_Simple}, we combine \Cref{lem:auxPropquadsimple} and Equation~\eqref{eq:boundinnerproc}, using the fact that $ \iter{p}{0}{t}= \hat \ite^{t-1}  $. 

This results means that for a quadratic function with gradients having uniformly bounded variance, the outer iteration decay is the same as for mini-batch iterations (but for mini-batch, it is true under the weaker set of Assumptions~\Cref{hyp:strong_convex}, \Cref{hyp:regularity}, \Cref{ass:def_filt}, \Cref{ass:lip_noisy_gradient_AS}, \Cref{hyp:variancebound}). 

\subsubsection{Proof}
\begin{proof}[Proof of \Cref{lem:auxPropquadsimple}]
	By definition of $ 	\breve{\ww}_{k}^t $, we have for any $ t\in \un{C}, k\in \un{\Nt} $, using the linearity of $ F' $ (Assumption~\Cref{hyp:quad}):
	\begin{align}
	\frac{1}{P}\sum_{i=1}^{P}\ww_{i,k+1}^t &= \frac{1}{P}\sum_{i=1}^{P}\ww_{i,k}^t - \frac{1}{P}\sum_{i=1}^{P}\eta_{k+1}^tg_{i,k+1}^t(\ww_{i,k}^t)\nonumber\\
	\breve{\ww}_{k+1}^t -\ww^\star&= \breve{\ww}_{k}^t -\ww^\star - \frac{1}{P}\sum_{i=1}^{P}\eta_{k+1}^tg_{i,k+1}^t(\ww_{i,k}^t)\nonumber\\
	\expe{\norm{\breve{\ww}_{k+1}^t-\ww^\star}^2|\mathcal{H}_{k,t}} &\leq \norm{\breve{\ww}_{k}^t-\ww^\star}^2  -2\eta_{k+1}^t\inner{\breve{\ww}_k^t-\ww^\star}{F'(\breve{\ww}_k^t)}\nonumber\\
	&+ (\eta_{k+1}^t)^2\expe{\norm{\frac{1}{P}\sum_{i=1}^{P}g_{i,k+1}^t(\ww_{i,k}^t)}^2|\mathcal{H}_{k,t}} \label{eq:decQuad1}\eqsp .
	\end{align}
	Now analyzing just the last term,
	\begin{align}
	&(\eta_{k+1}^t)^2\expe{\norm{\frac{1}{P}\sum_{i=1}^{P}g_{i,k+1}^t(\ww_{i,k}^t)}^2|\mathcal{H}_{k,t}}\nonumber\\
	&= (\eta_{k+1}^t)^2\expe{\norm{\frac{1}{P}\sum_{i=1}^{P}\big(g_{i,k+1}^t(\ww_{i,k}^t)-F'(\ww_{i,k}^t)\big)}^2|\mathcal{H}_{k,t}} + (\eta_{k+1}^t)^2\norm{F'(\breve{\ww}_{k}^t)}^2\label{eq:decQuad2}\eqsp.
	\end{align}
	Under the independence of the noises (Assumption~\Cref{ass:def_filt}), then the uniform upper bound on the variance (Assumption~\Cref{hyp:addit_noise}),  we have the following upper bound :
	\begin{align*}
	\expe{\norm{\frac{1}{P}\sum_{i=1}^{P}\big(g_{i,k+1}^t(\ww_{i,k}^t)-F'(\ww_{i,k}^t)\big)}^2|\mathcal{H}_{k,t}} & = \frac{1}{P^2}\sum_{i=1}^{P}\expe{\norm{\big(g_{i,k+1}^t(\ww_{i,k}^t)-F'(\ww_{i,k}^t)\big)}^2|\mathcal{H}_{k,t}}  \\
	&\le \frac{1}{P} \siginf\eqsp.
	\end{align*}
	Under Assumption~\Cref{hyp:quad}, $ F' $ is co-coercive, thus using Equation~\eqref{eq:gradcoco}, we have the following upper bound:
	\begin{align*}
	\expe{\norm{\breve{\ww}_{k+1}^t-\ww^\star}^2|\mathcal{H}_{k,t}} &\leq \norm{\breve{\ww}_{k}^t-\ww^\star}^2  -2 \eta_{k+1}^t (1-\eta_{k+1}^t L) \inner{\breve{\ww}_k^t-\ww^\star}{F'(\breve{\ww}_k^t)}+\frac{ (\eta_{k+1}^t)^2 \siginf}{P} \eqsp.
	\end{align*}
	And using strong convexity (esp. Equation~\eqref{eq:strong_conv}), and the fact that $ \eta_{k+1}^{t}L \le \frac{1}{2} $:
	\begin{align}
	\expe{\norm{\breve{\ww}_{k+1}^t-\ww^\star}^2|\mathcal{H}_{k,t}} &\leq (1 -2\mu \eta_{k+1}^t (1-\eta_{k+1}^t L))\norm{\breve{\ww}_{k}^t-\ww^\star}^2  +\frac{ (\eta_{k+1}^t)^2 \siginf}{P} \eqsp \nonumber\\
	&\le (1 -\mu \eta_{k+1}^t)\norm{\breve{\ww}_{k}^t-\ww^\star}^2 +\frac{ (\eta_{k+1}^t)^2 \siginf}{P} \eqsp. \label{eq:unifboundedvar}
	\end{align}
	By recursion, we then have, for any $ K\in \un{\Nt} $:
	\begin{align*}
	\expe{\norm{\breve{\ww}_{K}^t-\ww^\star}^2} &\leq \prod_{k=1}^{K}(1 -\mu \eta_{k}^t) \norm{\breve{\ww}_{0}^t-\ww^\star}^2 +\frac{  \siginf}{P} \sum_{k=1}^{K}  (\eta_{k}^t)^{2} \prod_{j=k}^{K}(1 -\mu \eta_{j}^t) \eqsp. 
	\end{align*}
	This concludes the proof.	
\end{proof}

\subsection{Proof of \Cref{prop:conv_quad_DSS}}
In this Section we prove \Cref{prop:conv_quad_DSS}. 

\subsubsection{Statement of \Cref{prop:conv_quad_DSS}}
\begin{proposition}[Local-SGD: Quadratic Functions]\label{prop:conv_quad_DSS}
	Under Assumptions~\Cref{hyp:quad},\Cref{ass:def_filt},\Cref{ass:lip_noisy_gradient_AS},\Cref{hyp:variancebound}, we have the following bound for one shot averaging: $ p\in \un{P}, t\in \un{C}, k \in \un{\Nt} $, 
	\begin{align}
	\expe{\mynorm[2]{ \hat \ite^{t} -\ws}} &\le\kappa_2^{t}\prod_{k=1}^{\sum_{t'=1}^{t}\Ntp} (1-\mu\tilde \eta_{k})\norm{{\ww}_0-\ww^\star}^2 + 2\kappa_1^{t}\kappa_2^{t}\frac{\Lone}{P}\sum_{t=1}^{\sum_{k=1}^{t}\Nt} \tilde \eta_{k}^{2} \prod_{j=k+1}^{\sum_{t'=1}^{t}\Ntp}(1 -\mu \tilde \eta_{j}) \label{tralala2}\\
	\expe{\mynorm[2]{ \iter{p}{k}{t} -\ws}} &\le\kappa_2^{t}\prod_{k=1}^{\sum_{t'=1}^{t}\Ntp+k} (1-\mu\tilde \eta_{k})\norm{{\ww}_0-\ww^\star}^2 + 2\kappa_1^{t}\kappa_2^{t}\frac{\Lone}{P}\sum_{u=1}^{\sum_{t'=1}^{t}\Ntp} \tilde \eta_{u}^{2} \prod_{j=k+1}^{\sum_{t'=1}^{t}\Ntp+k}(1 -\mu \tilde \eta_{j})\nonumber\\
	& + 2 \frac{\Lone}{P}\sum_{u=\sum_{t'=1}^{t}\Ntp}^{\sum_{t'=1}^{t}\Ntp+k} \tilde \eta_{u}^{2} \prod_{j=u+1}^{\sum_{t'=1}^{t}\Ntp+k}(1 -\mu \tilde \eta_{j}),
	\end{align}
	with, for $ t\in \un{C} $,  $\kappa_1^t= \left (4+ \mu \sum_{k=1}^{\Nt}(\eta_{k}^t)^2 \right ) $, and $\kappa_2^t:=  \exp\left (\mu \sum_{t'=0}^{t} \sum_{k=1}^{\Nt}(\eta_{k}^t)^2 \right )$.
\end{proposition}

When considering a constant step size $ \eta $, we have the following corollary.
\begin{corollary}[Local-SGD: Quadratic Functions]\label{prop:conv_quad}
	Under Assumptions~\Cref{hyp:quad},\Cref{ass:def_filt},\Cref{ass:lip_noisy_gradient_AS},\Cref{hyp:variancebound}, we have the following bound for one shot averaging: $ p\in \un{P}, t\in \un{C}, k \in \un{\Nt} $, constant learning rate $\eta$, 
	\begin{align}
	\expe{\mynorm[2]{ \hat \ite^{t-1} -\ws}} &\le\tau^{t}_2 (1-\eta\mu)^{\boldsymbol{N}_1^{t-1}}\norm{{\ww}_0-\ww^\star}^2 + 2\tau^{t}_1\tau^{t}_2\frac{\Lone \eta}{P}\frac{1 - (1-\eta \mu)^{\boldsymbol{N}_1^{t-1}}}{\mu} \label{tralala}\\
	\expe{\mynorm[2]{ \iter{p}{k}{t} -\ws}} &\le\tau_2^{t} (1-\eta\mu)^{\boldsymbol{N}_1^{t-1}+k}\norm{{\ww}_0-\ww^\star}^2 \nonumber \\ &+ 2 \Lone\Bigg(\sup_{t'=1 \dots t} (\tau^{t}_1) \tau^{t}_2\frac{1 - (1-\eta \mu)^{\boldsymbol{N}_1^{t-1}}}{P\mu} +\frac{1 - (1-\eta \mu)^{k}}{\mu}\Bigg).
	\end{align}
\end{corollary}
Where we have $\tau^{t}_1 = 4 + \mu\boldsymbol{N}^{t}\eta^2$ and $\tau^{t}_2 = \exp\left (\mu \boldsymbol{N}_1^{t}\eta^2 \right )$. Under the latter requirement (for optimality) that for any $ t $, $  \boldsymbol{N}^{t} \mu P \eta \le 1  $, we have $ \mu \boldsymbol{N}_1^{t}\eta^2 \le C \eta P^{-1} $, thus this is generally a small constant. This result is a consequence of \Cref{lem:auxProp_quad}.

\paragraph{Interpretation.} As before, the first bound shows that the variance of the iterates \emph{after communication} is reduced by a factor of $ P $ \emph{w.r.t.}~ the serial case, thus almost as good as mini-batch averaging. However, the constants involved are worse than in the additive noise setting (\Cref{prop:conv_quad_Simple}).
Consequently, and similarly to \Cref{prop:conv_quad_Simple}, the bound for the current iterates is  composed of two terms for the variance: a ``reduced variance'' coming from the communication step, and a ``inner loop'' variance, that does not benefit from the number of machines.

Finally, we provide a convergence result in the most general case, removing the quadratic assumption. For the sake of concision, we skip the bound for the averaged iterate after a communication round, and directly give the result for  the inner process.

\subsubsection{Proof}

This result is a consequence of \Cref{lem:auxProp_quad}, which implies \Cref{tralala}. 
Indeed, using it  recursively, and using $ (1+x)\le \exp(x) $, we get:  
\begin{align*}
\expe{\mynorm[2]{ \hat \ite^{T} -\ws}} &\le\exp\left (\mu \sum_{t'=0}^{T} \sum_{k=1}^{\Nt}(\eta_{k}^t)^2 \right )\prod_{t'=1}^{T}\prod_{k=1}^{\Ntp} (1-\mu\eta^t_{k})\expe{\norm{{\ww}_0-\ww^\star}^2} \\
&+ 2\kappa_1\exp\left (\mu \sum_{t'=0}^{t} \sum_{k=1}^{\Nt}(\eta_{k}^t)^2 \right )\frac{\Lone}{P}\sum_{t=1}^{T}  \prod_{t'=t+1}^{T} \prod_{k=1}^{\Ntp}(1 -\mu \eta_{k}^t) \sum_{k=1}^{\Nt}  (\eta_{k}^t)^{2} \prod_{j=k+1}^{\Nt}(1 -\mu \eta_{j}^t)  
\end{align*}
With, for $ t\in \un{C} $,  $\kappa_1^t= \left (4+ \mu \sum_{k=1}^{\Nt}(\eta_{k}^t)^2 \right ) $, and $\kappa_2^t:=  \exp\left (\mu \sum_{t'=0}^{t} \sum_{k=1}^{\Nt}(\eta_{k}^t)^2 \right )$, and re-writing everything in terms of the sequence $ \tilde \eta_k $, it gives \Cref{tralala2}. The second inequality naturally follows.

\begin{lemma}\label{lem:auxProp_quad}
	Under Assumptions~\Cref{hyp:quad}, \Cref{ass:def_filt}, \Cref{ass:lip_noisy_gradient_AS}, \Cref{hyp:variancebound}, for any $t\in \un{C}, K\in \un{\Nt} $, we have: 
	\begin{align}
	\expe{\norm{\hat{\ww}^t-\ww^\star}^2} 
	&\leq \left (1+ \mu\sum_{k=1}^{\Nt}(\eta_{k}^t)^2 \right ) \prod_{k=1}^{\Nt}(1-\mu\eta_{k})\expe{\norm{\hat{\ww}^{t-1}-\ww^\star}^2} \\
	&+ 2\left (4+ \mu \sum_{k=1}^{\Nt}(\eta_{k}^t)^2 \right )\frac{\Lone}{P}\sum_{k=0}^{\Nt}(\eta_{k}^t)^2 \prod_{j=k+1}^{\Nt}(1-\mu\eta_{j}^t).
	\end{align}
\end{lemma}

The proof is a bit technical, so we summarize here the 2 main steps:
\begin{enumerate}
	\item We prove an inequality (namely \Cref{eq:firststep}) that is comparable to \Cref{eq:unifboundedvar}, but with an extra term.
	\item  We use the control on the inner process (\Cref{sec:innercontro}) to control the extra term.
\end{enumerate}

\begin{proof}
	We consider again the ghost process defined at Equation~\eqref{eq:def_ghostit}.  Equations~\eqref{eq:decQuad1} and~\eqref{eq:decQuad2} are still valid. We now use the following decomposition\footnote{In the following, $ \square, \diamond, \clubsuit $, etc.  are used as symbolic notations to ease presentation.}: 
	\begin{align*}
	\square&:=(\eta_{k+1}^t)^2\expe{\norm{\frac{1}{P}\sum_{i=1}^{P}g_{i,k+1}^t(\ww_{i,k}^t)}^2|\mathcal{H}_{k,t}}\\
	&= (\eta_{k+1}^t)^2\expe{\norm{\frac{1}{P}\sum_{i=1}^{P}\big(g_{i,k+1}^t(\ww_{i,k}^t)-F'(\ww_{i,k}^t)\big)}^2|\mathcal{H}_{k,t}} + (\eta_{k+1}^t)^2\norm{F'(\breve{\ww}_{k}^t)}^2\\
	&\leq 2(\eta_{k+1}^t)^2\expe{\norm{\frac{1}{P}\sum_{i=1}^{P}\big(g_{i,k+1}^t(\ww_{i,k}^t)-F'(\ww_{i,k}^t)-g_{i,k+1}^t(\ww^\star)\big)}^2|\mathcal{H}_{k,t}}\\
	&+2(\eta_{k+1}^t)^2\expe{\norm{\frac{1}{P}\sum_{i=1}^{P}g_{i,k+1}^t(\ww^\star)}^2| \mathcal{H}_{k,t}} + (\eta_{k+1}^t)^2\norm{F'(\breve{\ww}_{k}^t)}^2\eqsp.
	\end{align*}
	Using the  independence of the noises (Assumption~\Cref{ass:def_filt}) we have,
	\begin{align*}
	\square&\leq \frac{2(\eta_{k+1}^t)^2}{P^2}\sum_{i=1}^{P}\expe{\norm{\big(g_{i,k+1}^t(\ww_{i,k}^t)-F'(\ww_{i,k}^t)-g_{i,k+1}^t(\ww^\star)\big)}^2|\mathcal{H}_{k,t}} \\
	&+\frac{2(\eta_{k+1}^t)^2}{P}\expe{\norm{g_{i,k+1}^t(\ww^\star)}^2| \mathcal{H}_{k,t}}
	+ (\eta_{k+1}^t)^2\norm{F'(\breve{\ww}_{k}^t)}^2\\
	&\leq \frac{4(\eta_{k+1}^t)^2}{P^2}\sum_{i=1}^{P}\bigg(\expe{\norm{\big(g_{i,k+1}^t(\ww_{i,k}^t)-g_{i,k+1}^t(\ww^\star)\big)}^2|\mathcal{H}_{k,t}} + \expe{\norm{\big(F'(\ww_{i,k}^t)-F'(\ww^\star)\big)}^2|\mathcal{H}_{k,t}}\bigg)\\
	&+\frac{2(\eta_{k+1}^t)^2}{P}\expe{\norm{g_{i,k+1}^t(\ww^\star)}^2| \mathcal{H}_{k,t}} + (\eta_{k+1}^t)^2\norm{F'(\breve{\ww}_{k}^t)}^2\eqsp. 
	\end{align*}
	Using Assumption~\Cref{ass:lip_noisy_gradient_AS} (co-coercivity for $(g_{i,k}^t)$-s and $F$) we obtain,
	\begin{align}
	\square&\leq \frac{8\CA(\eta_{k+1}^t)^2}{P^2}\sum_{i=1}^{P}\inner{F'(\ww_{i,k}^t)- F'(\ww^\star)}{\ww_{i,k}^t-\ww^\star} +\frac{2(\eta_{k+1}^t)^2}{P}\expe{\norm{g_{i,k+1}^t(\ww^\star)}^2| \mathcal{H}_{k,t}}\nonumber\\
	&+ (\eta_{k+1}^t)^2\CA\inner{F'(\breve{\ww}_{k}^t)}{\breve{\ww}_k^t-\ww^\star}\eqsp. \label{eq:quad17}
	\end{align}
	This leads to, combining Equations~\eqref{eq:decQuad1} and ~\eqref{eq:quad17}, and the upper bound on the variance of the noise at the optimum (Assumption~\Cref{hyp:variancebound})
	\begin{align*}
	\diamond&:=\expe{\norm{\breve{\ww}_{k+1}^t-\ww^\star}^2|\mathcal{H}_{k,t}} \\
	&\leq \norm{\breve{\ww}_{k}^t-\ww^\star}^2  -2\eta_{k+1}^t\inner{\breve{\ww}_k^t-\ww^\star}{F'(\breve{\ww}_k^t)}+\frac{2(\eta_{k+1}^t)^2}{P}\expe{\norm{g_{i,k+1}^t(\ww^\star)}^2| \mathcal{H}_{k,t}}\\
	&+ \frac{8\CA(\eta_{k+1}^t)^2}{P^2}\sum_{i=1}^{P}\inner{F'(\ww_{i,k}^t)- F'(\ww^\star)}{\ww_{i,k}^t-\ww^\star}  +(\eta_{k+1}^t)^2\CA\inner{F'(\breve{\ww}_{k}^t)}{\breve{\ww}_k^t-\ww^\star}\\
	&\leq \norm{\breve{\ww}_{k}^t-\ww^\star}^2  -2\eta_{k+1}^t(1-\eta_{k+1}^t\CA)\inner{\breve{\ww}_k^t-\ww^\star}{F'(\breve{\ww}_k^t)} +2\frac{(\eta_{k+1}^t)^2}{P}\sigma^{2}\\
	&+ \frac{8\CA(\eta_{k+1}^t)^2}{P^2}\sum_{i=1}^{P}\inner{F'(\ww_{i,k}^t)- F'(\ww^\star)}{\ww_{i,k}^t-\ww^\star}\eqsp.
	\end{align*}
	Using $\CA\eta^t_{k+1}\leq\frac{1}{2}$, and strong-convexity (Assumption~\Cref{hyp:strong_convex}) 
	\begin{align}
	\expe{\norm{\breve{\ww}_{k+1}^t-\ww^\star}^2|\mathcal{H}_{k,t}} &\leq (1-\mu \eta_{k+1}^t)\norm{\breve{\ww}_{k}^t-\ww^\star}^2  +\frac{2(\eta_{k+1}^t)^2\Lone}{P}\nonumber\\
	&+ \frac{8\CA(\eta_{k+1}^t)^2}{P^2}\sum_{i=1}^{P}\inner{F'(\ww_{i,k}^t)- F'(\ww^\star)}{\ww_{i,k}^t-\ww^\star}\eqsp. \label{eq:firststep}
	\end{align}
	
	This inequality should be compared to~Equation~\eqref{eq:unifboundedvar}. It is interesting to remark that the last term is not an artifact of the proof: this is easy to check for least-squares regression.
	
	Using recursively the above inequality and using the  definition of $\breve{\ww}^t$, and taking expectation on the historical randomness we have, for any $ N\in \un{\Nt-1} $
	\begin{align*}
	\expe{\norm{\breve{\ww}_{N+1}^t-\ww^\star}^2} &\leq \prod_{k=0}^{N}(1-\mu\eta^{t}_{k+1})\expe{\norm{\breve{\ww}_{0}^{t}-\ww^\star}^2} + 2\frac{\Lone}{P}\sum_{k=0}^{N}(\eta_{k+1}^t)^2\prod_{j=k+1}^{N}(1-\mu\eta_{j+1}^t)\nonumber\\
	&+ \frac{8\CA}{P^2}\expe{\sum_{k=0}^{N}(\eta_{k+1}^t)^2\sum_{i=1}^{P}\inner{F'(\ww_{i,k}^t)- F'(\ww^\star)}{\ww_{i,k}^t-\ww^\star}\prod_{j=k+1}^{N}(1-\mu\eta_{j+1}^t)} .
	\end{align*}
	Especially, for $ N=\Nt-1 $, $  \breve{\ww}_{\Nt}^t = \hat \ite ^t$, and moreover $  \breve{\ww}_{0}^t = \hat \ite ^{t-1}$:
	\begin{align}
	\expe{\norm{\hat{\ww}^t-\ww^\star}^2} &\leq \prod_{k=0}^{\Nt-1}(1-\mu\eta^{t}_{k+1})\expe{\norm{\hat{\ww}^{t-1}-\ww^\star}^2} + 2\frac{\Lone}{P}\sum_{k=0}^{\Nt-1}(\eta_{k+1}^t)^2\prod_{j=k+1}^{\Nt-1}(1-\mu\eta_{j+1}^t)\nonumber\\
	&+ \frac{8\CA}{P^2}\expe{\sum_{k=0}^{\Nt-1}(\eta_{k+1}^t)^2\sum_{i=1}^{P}\inner{F'(\ww_{i,k}^t)- F'(\ww^\star)}{\ww_{i,k}^t-\ww^\star}\prod_{j=k+1}^{\Nt-1}(1-\mu\eta_{j+1}^t)}\eqsp.\label{eq:part1}
	\end{align}
	To upper bound the last term in the above equation, we use~\Cref{eq:bound_inner_product},
	\begin{align*}
	\clubsuit&:=\frac{8\CA}{P^2}\sum_{k=0}^{\Nt-1}(\eta_{k+1}^t)^2\sum_{i=1}^{P}\inner{F'(\ww_{i,k}^t)- F'(\ww^\star)}{\ww_{i,k}^t-\ww^\star}\prod_{j=k+1}^{\Nt-1}(1-\mu\eta_{j+1}^t)\nonumber\\
	&\leq \frac{8\CA}{P^2}\sum_{k=0}^{\Nt-1}\eta_{k+1}^t\sum_{i=1}^{P}\bigg(\expe{\norm{\ww_{i,k}^t-\ww^\star}^2}-\expe{\norm{\ww_{i,k+1}^t-\ww^\star}^2}\nonumber\\
	&+2(\eta_{k+1}^t)^2\Lone\bigg)\prod_{j=k+1}^{\Nt-1}(1-\mu\eta_{j+1}^t)\nonumber\\
	&\leq \frac{8\CA}{P^2}\sum_{k=0}^{\Nt-1}\eta_{k+1}^t\sum_{i=1}^{P}\bigg(\expe{\norm{\ww_{i,k}^t-\ww^\star}^2}-\expe{\norm{\ww_{i,k+1}^t-\ww^\star}^2}\bigg)\prod_{j=k+1}^{\Nt-1}(1-\mu\eta_{j+1}^t)\nonumber\\
	&+\frac{16\CA \Lone}{P}\sum_{k=0}^{\Nt-1}(\eta_{k+1}^t)^3 \prod_{j=k+1}^{\Nt-1}(1-\mu\eta_{j+1}^t)\eqsp.
	\end{align*}
	Note that since the mean squared distance doesn't depend on the machine, we  can assume to be working on machine $1$. This leads to,  using an Abel transform:
	\begin{align*}
	\clubsuit&\leq \frac{8\CA}{P}\sum_{k=0}^{\Nt-1}\bigg(\expe{\norm{\ww_{1,k}^t-\ww^\star}^2}-\expe{\norm{\ww_{1,k+1}^t-\ww^\star}^2}\bigg)\prod_{j=k+1}^{\Nt-1}(1-\mu\eta_{j+1}^t)\eta_{k+1}^t\nonumber\\
	&+\frac{16\CA \Lone}{P}\sum_{k=0}^{\Nt-1}(\eta_{k+1}^t)^3 \prod_{j=k+1}^{\Nt-1}(1-\mu\eta_{j+1}^t)\nonumber\\
	&\leq \frac{8\CA}{P}\Bigg(\sum_{k=0}^{\Nt-1}\expe{\norm{\ww_{1,k}^t-\ww^\star}^2}\bigg(\eta_{k+1}^t\prod_{j=k+1}^{\Nt-1}(1-\mu\eta^t_{j+1})-\eta_k^t\prod_{j=k}^{\Nt-1}(1-\mu\eta_{j+1}^t)\bigg)\nonumber\\
	&+\expe{\norm{\ww_{1,0}^t-\ww^\star}^{2}}\prod_{j=0}^{\Nt-1}(1-\mu \eta_{j+1}^t)\eta_0^t-\expe{\norm{\ww_{1,\Nt}^t-\ww^\star}^{2}}\eta_{\Nt}^t\Bigg)\nonumber\\
	&+\frac{16\CA \Lone}{P}\sum_{k=0}^{\Nt-1}(\eta_{k+1}^t)^3 \prod_{j=k+1}^{\Nt-1}(1-\mu\eta_{j+1}^t)\eqsp.
	\end{align*}
	Finally, using convexity, we have that $$  \expe{\norm{\breve{\ww}^t_{\Nt}-\ww^\star}^{2}} \le \frac{1}{P} \sum_{p=1}^{P}\expe{\norm{\ww_{p,\Nt}^t-\ww^\star}^{2} } = \expe{\norm{\ww_{1,\Nt}^t-\ww^\star}^{2} } .$$ 
	Thus:
	\begin{align}
	\clubsuit&\leq \frac{8\CA}{P}\sum_{k=0}^{\Nt-1}\expe{\norm{\ww_{1,k}^t-\ww^\star}^2}\prod_{j=k+1}^{\Nt-1}(1-\mu\eta_{j+1})\big(\eta_{k+1}^t-\eta_k^t(1-\mu\eta_{k+1}^t)\big)\nonumber\\
	&+\frac{8\CA}{P}\expe{\norm{\hat{\ww}^{t-1}-\ww^\star}^2}\prod_{j=0}^{\Nt-1}(1-\mu \eta_{j+1}^t)\eta_0^t -\frac{8\CA}{P}\expe{\norm{\hat{\ww}^t-\ww^\star}} \eta_{\Nt}^t\nonumber\\
	&+\frac{16\CA \Lone}{P}\sum_{k=0}^{\Nt-1}(\eta_{k+1}^t)^3 \prod_{j=k+1}^{\Nt-1}(1-\mu\eta_{j+1}^t)\eqsp.\label{eq:part2}
	\end{align}
	We now use \Cref{eq:boundinnerproc}. It leads to the following,
	\begin{align}
	&\frac{8\CA}{P}\sum_{k=0}^{\Nt-1}\expe{\norm{\ww_{1,k}^t-\ww^\star}^2}\prod_{j=k+1}^{\Nt-1}(1-\mu\eta_{j+1})\big(\eta_{k+1}^t-\eta_k^t(1-\mu\eta_{k+1}^t)\big)\nonumber\\
	&\leq \frac{8\CA}{P}\prod_{j=0}^{\Nt-1}(1-\mu\eta_{j+1})\expe{\norm{\hat{\ww}^{t-1}-\ww^\star}^2}\sum_{k=0}^{\Nt-1}\big(\eta_{k+1}^t-\eta_k^t(1-\mu\eta_{k+1}^t)\big)\nonumber\\
	&+ \frac{8\CA}{P}\sum_{k=0}^{\Nt-1}\big(2\Lone \sum_{l=1}^{k}(\eta_{l}^t)^2\prod_{m=l+1}^{k}(1-\mu\eta_{m}^t)\big)\prod_{j=k+1}^{\Nt-1}(1-\mu\eta_{j+1})\big(\eta_{k+1}^t-\eta_k^t(1-\mu\eta_{k+1}^t)\big)\nonumber\\
	&\leq \frac{8\CA}{P}\prod_{j=0}^{\Nt-1}(1-\mu\eta_{j+1})\expe{\norm{\hat{\ww}^{t-1}-\ww^\star}^2}\sum_{k=0}^{\Nt-1}\big(\eta_{k+1}^t-\eta_k^t(1-\mu\eta_{k+1}^t)\big)\nonumber\\
	&+ \frac{16 \Lone \CA}{P}\sum_{k=0}^{\Nt-1}\sum_{l=1}^{k}(\eta_{l}^t)^2\prod_{j=l+1}^{\Nt-1}(1-\mu\eta_{j+1})\big(\eta_{k+1}^t-\eta_k^t(1-\mu\eta_{k+1}^t)\big)\nonumber\\ 
	&\leq \frac{8\CA}{P}\prod_{j=0}^{\Nt-1}(1-\mu\eta_{j+1})\expe{\norm{\hat{\ww}^{t-1}-\ww^\star}^2}\big(\eta_{\Nt-1}^t-\eta_{0}^t+ \sum_{k=0}^{\Nt-1}\mu(\eta_k^t)^2\big)\nonumber\\
	&+ \frac{16 \Lone \CA}{P}\sum_{l=1}^{\Nt-1}\sum_{k=l}^{\Nt-1}(\eta_{l}^t)^2\prod_{j=l+1}^{\Nt-1}(1-\mu\eta_{j+1})\big(\eta_{k+1}^t-\eta_k^t(1-\mu\eta_{k+1}^t)\big)\nonumber\\ 
	&\leq \frac{8\CA}{P}\prod_{j=0}^{\Nt-1}(1-\mu\eta_{j+1})\expe{\norm{\hat{\ww}^{t-1}-\ww^\star}^2}\big(\eta_{\Nt-1}^t-\eta_{0}^t+ \sum_{k=0}^{\Nt-1}\mu(\eta_k^t)^2\big)\nonumber\\
	&+ \frac{16 \Lone \CA}{P}\sum_{l=1}^{\Nt-1}(\eta_{l}^t)^2\prod_{j=l+1}^{\Nt-1}(1-\mu\eta_{j+1})\sum_{k=0}^{\Nt-1}\big(\eta_{k+1}^t-\eta_k^t(1-\mu\eta_{k+1}^t)\big)\nonumber\\
	&\leq \frac{8\CA}{P}\prod_{j=0}^{\Nt-1}(1-\mu\eta_{j+1})\expe{\norm{\hat{\ww}^{t-1}-\ww^\star}^2}\big(\eta_{{\Nt}}^t-\eta_{0}^t+ \sum_{k=0}^{\Nt-1}\mu(\eta_{k+1}^t)^2\big)\nonumber\\
	&+ \frac{16 \Lone \CA}{P}\sum_{k=0}^{\Nt-1}(\eta_{k+1}^t)^2\prod_{j=k+1}^{\Nt-1}(1-\mu\eta_{j+1})\big(\eta_{\Nt}^t-\eta_{0}^t+ \sum_{k=0}^{\Nt-1}\mu(\eta_{k+1}^t)^2\big) \label{eq:part3}.
	\end{align} 
	Combining \Cref{eq:part1,eq:part2,eq:part3}, we get, denoting $ C_{\Nt}=\eta_{\Nt}^t+ \sum_{k=0}^{\Nt-1}\mu(\eta_{k+1}^t)^2 $:
	\begin{align*}
	\expe{\norm{\hat{\ww}^t-\ww^\star}^2} &\leq \prod_{k=0}^{\Nt-1}(1-\mu\eta_{k+1})\expe{\norm{\hat{\ww}^{t-1}-\ww^\star}^2} + 2\frac{\Lone}{P}\sum_{k=0}^{\Nt-1}(\eta_{k+1}^t)^2\prod_{j=k+1}^{\Nt-1}(1-\mu\eta_{j+1}^t)\nonumber\\
	&+\frac{8\CA}{P}\prod_{j=0}^{\Nt-1}(1-\mu\eta_{j+1})\expe{\norm{\hat{\ww}^{t-1}-\ww^\star}^2}(C_{\Nt}-\eta_0^t) -\frac{8\CA}{P}\expe{\norm{\hat{\ww}^t-\ww^\star}}\eta_{\Nt}^t\nonumber\\
	&+ \frac{16 \Lone \CA}{P}\sum_{k=0}^{\Nt-1}(\eta_{k+1}^t)^2\prod_{j=k+1}^{\Nt-1}(1-\mu\eta_{j+1})(C_{\Nt}-\eta_0^t)\nonumber\\
	&+ \frac{8\CA}{P}\expe{\norm{\hat{\ww}^{t-1}-\ww^\star}^2}\prod_{j=0}^{\Nt-1}(1-\mu \eta_{j+1}^t)\eta_0^t +\frac{16\CA \Lone}{P}\sum_{k=0}^{\Nt-1}(\eta_{k+1}^t)^3 \prod_{j=k+1}^{\Nt-1}(1-\mu\eta_{j+1}^t)\nonumber\eqsp.
	\end{align*}
	Thus, simplifying:
	\begin{align*}
	&\hspace{-4em}\left (1+ \frac{8\CA}{P}\eta_{\Nt}^t\right )\expe{\norm{\hat{\ww}^t-\ww^\star}^2} \\
	&\leq \left (1 +\frac{8\CA}{P} \eta_{\Nt}^t+ \sum_{k=0}^{\Nt-1}\mu(\eta_{k+1}^t)^2 \right ) \prod_{k=0}^{\Nt-1}(1-\mu\eta_{k+1})\expe{\norm{\hat{\ww}^{t-1}-\ww^\star}^2} \\
	&+ 2\frac{\Lone}{P}\sum_{k=0}^{\Nt-1}(\eta_{k+1}^t)^2 \left (1+\frac{8\CA}{P}\eta_{\Nt}^t+ \sum_{k=0}^{\Nt-1}\mu(\eta_{k+1}^t)^2 +L \eta_{k+1}^t\right )\prod_{j=k+1}^{\Nt-1}(1-\mu\eta_{j+1}^t)\nonumber.
	\end{align*}
	This concludes the proof of  the Lemma, using $ L\eta^t_k\le 1/2 $:
	\begin{align*}
	\expe{\norm{\hat{\ww}^t-\ww^\star}^2} 
	&\leq \left (1+ \mu\sum_{k=1}^{\Nt}(\eta_{k}^t)^2 \right ) \prod_{k=1}^{\Nt}(1-\mu\eta_{k})\expe{\norm{\hat{\ww}^{t-1}-\ww^\star}^2} \\
	&+ 2\left (4+ \mu \sum_{k=1}^{\Nt}(\eta_{k}^t)^2 \right )\frac{\Lone}{P}\sum_{k=0}^{\Nt}(\eta_{k}^t)^2 \prod_{j=k+1}^{\Nt}(1-\mu\eta_{j}^t)\nonumber.
	\end{align*}
\end{proof}

This result can be used recursively. It implies that if 
$
\mu \sum_{t=1}^{C}\sum_{k=1}^{\Nt}(\eta_{k}^t)^2 \le K
$, then the upper bound on the outer iterates is as good as the one for mini-batch, up to a constant.

\subsection{Proof of \Cref{prop:conv_general_function_DSS}}
In this Section we prove the first upper bound of \Cref{prop:conv_general_function}.

\subsubsection{Statement of \Cref{prop:conv_general_function_DSS}}

Finally, we provide a convergence result in the most general case, removing the quadratic assumption.

\begin{proposition}[Local-SGD: General Functions]\label{prop:conv_general_function_DSS}
	Under Assumptions~\Cref{hyp:strong_convex},~\Cref{hyp:regularity},~\Cref{ass:def_filt},~\Cref{hyp:addit_noise} we have:
	
	\begin{align}
	\expe{\mynorm[2]{ \iter{p}{k}{t} -\ws}} &\le\kappa_2\prod_{k=1}^{\sum_{t'=1}^{t}\Ntp+k} (1-\mu\tilde \eta_{k})\norm{{\ww}_0-\ww^\star}^2 + 2 \frac{\Lone}{P}\sum_{u=\sum_{t'=1}^{t}\Ntp}^{\sum_{t'=1}^{t}\Ntp+k} \tilde \eta_{u}^{2} \prod_{j=u+1}^{\sum_{t'=1}^{t}\Ntp+k}(1 -\mu \tilde \eta_{j}) \nonumber\\
	&+  (\sup_{t'=1 \dots t} C_{P,M,K,t'}) \frac{\Lone}{P}\sum_{u=1}^{\sum_{t'=1}^{t}\Ntp} \tilde \eta_{u}^{2} \prod_{j=k+1}^{\sum_{t'=1}^{t}\Ntp+k}(1 -\mu \tilde \eta_{j}),\nonumber	
	\end{align}
	with $  C_{P,M,K,t}=1 +M P \sum_{k=1}^{K} \eta_{k}^t\mynorm{\breve{\ww}_{k-1}^t-\ww^\star}  $.
\end{proposition}
\paragraph{Interpretation: } if  $ (\sup_{t'=1 \dots t}  C_{P,M,K,t} )  $ is uniformly bounded, we perform as well as minibatch SGD for the outer iterations (up to a constant).

For a constant step size $ \eta $, the proposition has the following corollary:
\begin{corollary}[Local-SGD: General Functions]\label{prop:conv_general_function}
	Under Assumptions~\Cref{hyp:strong_convex},~\Cref{hyp:regularity},~\Cref{ass:def_filt},~\Cref{hyp:addit_noise} we have:
	\begin{align*}
	\expe{\mynorm[2]{ \iter{p}{k}{t} -\ws}} &\le\tau_2^{t}(1-\eta\mu)^{\boldsymbol{N}_1^{t-1}+k}\norm{{\ww}_0-\ww^\star}^2\nonumber\\ & + {\siginf}\Bigg( \left (\sup_{t'=1 \dots t} C_{P,M,t'} \right )\frac{1 - (1-\eta \mu)^{\boldsymbol{N}_1^{t-1}}}{P\mu}+2\frac{1 - (1-\eta \mu)^{k}}{\mu} \Bigg) .
	\end{align*}
\end{corollary}
Where $  C_{P,M,t} = 1 +M P \eta \sum_{k=1}^{N^{t}} \expe{\mynorm{\breve{\ww}_{k-1}^t-\ww^\star}}  $.  We prove the on-line case of the result using \Cref{lem:lemme_aux_conv_inner_gen_case} in supplementary material.

\paragraph{Interpretation.} 
When communication occurs, averaging the different models over the machines results in a variance reduction, but at each phase, the variance accumulated within the phase is degraded with respect to the simplest setting by at most $  C_{P,M,t} $. This constant increases with the number of machines and the step size, and also depends on the mean distance $ \sum_{k=1}^{N^{t}} \expe{\mynorm{\breve{\ww}_{k-1}^t-\ww^\star}} $ during  phase $ t $.
As a consequence if $ C_{P,M,t}  $ is uniformly bounded, we perform as well as mini-batch SGD. If $ \expe{\mynorm{\breve{\ww}_{k-1}^t-\ww^\star} }$ is assumed to be decaying, this is true if for any $ t\in \un{T} $, $ \Nt \eta M P \expe{\norm{\hat{\ww}^t-\ww^\star}}   \le O(1) $.

In the following, we alternatively relax the  bounded variance assumption \Cref{hyp:addit_noise} and the quadratic assumption \Cref{hyp:quad}, and show similar results for local SGD. This allows us to successively cover the cases of least squares regression (LSR) and logistic regression (LR).

\subsubsection{Proof}
\Cref{prop:conv_general_function_DSS} follows from \Cref{lem:lemme_aux_conv_inner_gen_case}. We have for any $t\in \un{C}, K\in \un{\Nt}$, 
\begin{align*}
\E \norm{\breve{\ww}_{K}^t-\ww^\star}^2 
&\le \prod_{k=1}^{K}(1-  \mu \eta_k^{t}) \E \norm{\breve{\ww}_{0}^t-\ww^\star}^2 
+ C_{P,M,K,t}\frac{\siginf}{P}  \sum_{k=1}^{K}(\eta_{k}^t)^2 \prod_{j=k+1}^{K}(1-  \mu \eta_j^{t}), 
\end{align*}
with $  C_{P,M,K,t}=1 +M P \sum_{k=1}^{K} \eta_{k}^t\mynorm{\breve{\ww}_{k-1}^t-\ww^\star}  $.

As in the two previous sections, we first  focus on upper bounding $ \expe{\mynorm[2]{\breve{\ww}_k^t-\ws}}$. We prove the following Lemma:
\begin{lemma}\label{lem:lemme_aux_conv_inner_gen_case}
	For any $t\in \un{C}, K\in \un{\Nt}$, under Assumptions~\Cref{hyp:strong_convex},~\Cref{hyp:regularity},~\Cref{ass:def_filt},~\Cref{hyp:addit_noise} we have:
	\begin{align*}
	\E 	\norm{\breve{\ww}_{K}^t-\ww^\star}^2 
	&\le \prod_{k=1}^{K}(1-  \mu \eta_k^{t})\E  \norm{\breve{\ww}_{0}^t-\ww^\star}^2 
	+ C_{P,M,K,t}\frac{\siginf}{P}  \sum_{k=1}^{K}(\eta_{k}^t)^2 \prod_{j=k+1}^{K}(1-  \mu \eta_j^{t}), 
	\end{align*}
	with $  C_{P,M,K,t}=1 +M P \sum_{k=1}^{K} \eta_{k}^t\expe{\mynorm{\breve{\ww}_{k-1}^t-\ww^\star} } $.  	
\end{lemma}
This means, if we have consider an weak upper bound on $ \expe{\mynorm{\breve{\ww}_k^t-\ww^\star}} \le R $  that the inner loops keeps the same variance as the mini-batch case if $ MP \sum_{k=1}^{K} \eta_{k}^t =O(1)$. 
For example, for a constant step size $ \eta $, it results in $ P \Nt \eta\le 1 $, \ie~$ \Nt\le \frac{1}{P\eta} $.
Note that the number of inner steps one can make increases with the phases, as $ \expe{\mynorm{\hat w^t-\ws}} $ decreases.

\subsubsection{Proof of \Cref{lem:lemme_aux_conv_inner_gen_case}}\label{subsec:prooflemme_aux_conv_inner_gen_case}
We rely on the following decomposition. Almost surely, we have:
\begin{align}
\expe{\norm{\breve{\ww}_{k+1}^t-\ww^\star}^2 |\h{k}{t}}&\le \norm{\breve{\ww}_{k}^t-\ww^\star}^2  -2\eta_{k+1}^t\inner{\breve{\ww}_k^t-\ww^\star}{F'(\breve{\ww}_k^t)}\nonumber\\
&+ (\eta_{k+1}^t)^2\expe{\norm{\frac{1}{P}\sum_{i=1}^{P}g_{i,k+1}^t(\ww_{i,k}^t)}^2|\mathcal{H}_{k,t}}\nonumber\\
&+2\eta_{k+1}^t\inner{\breve{\ww}_k^t-\ww^\star}{F'(\breve{\ww}_k^t)-\frac{1}{P}\sum_{p=1}^{P} F'(\iter{p}{k}{t})}\label{eq:dec1}.
\end{align}
The first two lines correspond to the quadratic case (\Cref{eq:decQuad1}), that has been analyzed in \Cref{lem:auxProp_quad}. The third term accounts for the difference between the mean gradient and the gradient at the mean point. We use Assumption~\Cref{hyp:regularity} to control this term.

We then use the following Lemma, which control how the inner iterates $ \iter{p}{k}{t} $ deviate from their average $ \breve{\ww}_k^t $:
\begin{lemma}\label{lem:upperboundondivergence}
	For any $ \fortk $, under Assumptions~\Cref{hyp:strong_convex},~\Cref{hyp:regularity},~\Cref{ass:def_filt},~\Cref{hyp:addit_noise} we have a.s.:
	\begin{align*}
	\frac{1}{P}\sum_{p=1}^{P}\expe{\mynorm[2]{\iter{p}{k}{t}-\breve{\ww}_k^t}}
	&\le   \siginf \sum_{j=1}^{k} (\eta_j^t)^{2}  \prod_{s=j+1}^{k} (1-\eta_s^t \mu).
	\end{align*}
\end{lemma}
The proof of this Lemma is postponed to \Cref{subsec:prooflemma_dev}.

Using Cauchy-Schwarz inequality and the bound on the third order derivative of $ F $, we have:
\begin{align}
2\eta_{k+1}^t\inner{\breve{\ww}_k^t-\ww^\star}{F'(\breve{\ww}_k^t)-\frac{1}{P}\sum_{p=1}^{P} F'(\iter{p}{k}{t})} &\le  2\eta_{k+1}^t\mynorm{\breve{\ww}_k^t-\ww^\star}\mynorm{F'(\breve{\ww}_k^t)-\frac{1}{P}\sum_{p=1}^{P} F'(\iter{p}{k}{t})}, \label{eq:dec2}
\end{align}
and, using a second order expansion of the gradient at $\breve{\ww}_k^t  $ together with Assumption~\Cref{hyp:regularity} we have:
\begin{align}
\mynorm{F'(\breve{\ww}_k^t)-\frac{1}{P}\sum_{p=1}^{P} F'(\iter{p}{k}{t})}
&\le  \frac{M}{P}\sum_{p=1}^{P}\mynorm[2]{\iter{p}{k}{t}-\breve{\ww}_k^t}.\label{eq:boundwithM}
\end{align}

Using the proof of \Cref{eq:unifboundedvar}, and combining \Cref{eq:dec1,eq:boundwithM,eq:dec2} and \Cref{lem:upperboundondivergence}, we have, for any $ \fortk $:
\begin{align}
\vartriangle&:=\expe{\norm{\breve{\ww}_{k+1}^t-\ww^\star}^2 |\h{k}{t}}\nonumber\\
\vartriangle&\le \norm{\breve{\ww}_{k}^t-\ww^\star}^2  -2\eta_{k+1}^t\inner{\breve{\ww}_k^t-\ww^\star}{F'(\breve{\ww}_k^t)}+ (\eta_{k+1}^t)^2\expe{\norm{\frac{1}{P}\sum_{i=1}^{P}g_{i,k+1}^t(\ww_{i,k}^t)}^2|\mathcal{H}_{k,t}}\nonumber\\
&+2\eta_{k+1}^t\inner{\breve{\ww}_k^t-\ww^\star}{F'(\breve{\ww}_k^t)-\frac{1}{P}\sum_{p=1}^{P} F'(\iter{p}{k}{t})}\nonumber\\
\E[\vartriangle]&\le (1-  \mu \eta_{k+1}^{t}) \expe{\norm{\breve{\ww}_{k}^t-\ww^\star}^2 }
+ (\eta_{k+1}^t)^2 \frac{1}{P}\siginf \nonumber\\
& +2\eta_{k+1}^t\expe{\mynorm{\breve{\ww}_k^t-\ww^\star}}M \sum_{j=1}^{k} (\eta_j^t)^{2} \siginf \prod_{s=j+1}^{k} (1-\eta_s^t \mu)  \label{eq:equivmom4}.
\end{align}
Thus by induction, for any $ t\in \un{C}, K\in \un{\Nt} $:
\begin{align*}
\expe{\norm{\breve{\ww}_{K}^t-\ww^\star}^2 }
&\le \prod_{k=1}^{K}(1-  \mu \eta_k^{t}) \expe{\norm{\breve{\ww}_{0}^t-\ww^\star}^2} 
+ \frac{1}{P}\siginf  \sum_{k=1}^{K}(\eta_{k}^t)^2 \prod_{j=k+1}^{K}(1-  \mu \eta_j^{t}) \\
& + 2 \siginf M\sum_{k=1}^{K} \eta_{k}^t\expe{\mynorm{\breve{\ww}_{k-1}^t-\ww^\star}} \sum_{j=1}^{k} (\eta_j^t)^{2}  \prod_{s=j+1}^{k} (1-\eta_s^t \mu)\prod_{j=k+1}^{K}(1-  \mu \eta_j^{t}) \nonumber\\
&= \prod_{k=1}^{K}(1-  \mu \eta_k^{t})\expe{\norm{\breve{\ww}_{0}^t-\ww^\star}^2} 
+ \frac{1}{P}\siginf  \sum_{k=1}^{K}(\eta_{k}^t)^2 \prod_{j=k+1}^{K}(1-  \mu \eta_j^{t}) \\
& + 2 M \siginf \sum_{j=1}^{K} (\eta_j^t)^{2}   \prod_{s=j+1}^{K} (1-  \mu \eta_j^{t}) \sum_{k=j}^{K} \eta_{k}^t\expe{\mynorm{\breve{\ww}_{k-1}^t-\ww^\star}  }\nonumber\\
&= \prod_{k=1}^{K}(1-  \mu \eta_k^{t}) \expe{\norm{\breve{\ww}_{0}^t-\ww^\star}^2} 
+ C_{P,M,K,t}\frac{\siginf}{P}  \sum_{k=1}^{K}(\eta_{k}^t)^2 \prod_{j=k+1}^{K}(1-  \mu \eta_j^{t}), 
\end{align*}
with $  C_{P,M,K,t}=1 +M P \sum_{k=1}^{K} \eta_{k}^t\expe{\mynorm{\breve{\ww}_{k-1}^t-\ww^\star} } $.  
This concludes the proof.

In the following section, we proved the auxiliary Lemma that was used in the proof.

\subsubsection{Proof of \Cref{lem:upperboundondivergence}}\label{subsec:prooflemma_dev}
We now study $ \frac{1}{P}\sum_{p=1}^{P}\mynorm[2]{\iter{p}{k}{t}-\breve{\ww}_k^t} $ as $ k  $ increases. Note that initially ($ k=0 $), this quantity is 0. For any $ k\in \un{\Nt} , p\in \un{P}$: 
\begin{align*}
\mynorm[2]{\iter{p}{k}{t}-\breve{\ww}_k^t} &=\mynorm[2]{\iter{p}{k-1}{t}-\eta^t_k g^t_{p,k}(\iter{p}{k-1}{t}) - \breve{\ww}_{k-1}^t +\eta^t_k \frac{1}{P} \sum_{i=1}^{P} g^t_{i,k}(\iter{i}{k-1}{t}) }\\
&=\mynorm[2]{\iter{p}{k-1}{t}- \breve{\ww}_{k-1}^t }- 2\eta_k^t \ps{ \iter{p}{k-1}{t}- \breve{\ww}_{k-1}^t}{ g^t_{p,k}(\iter{p}{k-1}{t}) -\frac{1}{P} \sum_{i=1}^{P} g^t_{i,k}(\iter{i}{k-1}{t}) }\\
&+ (\eta_k^t)^{2} \mynorm[2]{ g^t_{p,k}(\iter{p}{k-1}{t}) -\frac{1}{P} \sum_{i=1}^{P} g^t_{i,k}(\iter{i}{k-1}{t}) }.
\end{align*}
Thus, expanding and using cocoercivity Assumption:
\begin{align*}
\expe{\mynorm[2]{\iter{p}{k}{t}-\breve{\ww}_k^t}|\h{t}{k-1}}
&=\mynorm[2]{\iter{p}{k-1}{t}- \breve{\ww}_{k-1}^t }\\
&- 2\eta_k^t \ps{ \iter{p}{k-1}{t}- \breve{\ww}_{k-1}^t}{ F'(\iter{p}{k-1}{t}) -\frac{1}{P} \sum_{i=1}^{P} F'(\iter{i}{k-1}{t}) }\\
&+ \expe{(\eta_k^t)^{2} \mynorm[2]{ g^t_{p,k}(\iter{p}{k-1}{t}) -\frac{1}{P} \sum_{i=1}^{P} g^t_{i,k}(\iter{i}{k-1}{t}) }|\h{t}{k-1}}\\
&=\mynorm[2]{\iter{p}{k-1}{t}- \breve{\ww}_{k-1}^t }- 2\eta_k^t \ps{ \iter{p}{k-1}{t}- \breve{\ww}_{k-1}^t}{ F'(\iter{p}{k-1}{t}) - F'(\breve{\ww}_{k-1}^t)}\\&+2\eta_k^t\ps{ \iter{p}{k-1}{t}- \breve{\ww}_{k-1}^t}{F'(\breve{\ww}_{k-1}^t) - \frac{1}{P} \sum_{i=1}^{P} F'(\iter{i}{k-1}{t}) }\\
&+ \expe{(\eta_k^t)^{2} \mynorm[2]{ g^t_{p,k}(\iter{p}{k-1}{t}) -\frac{1}{P} \sum_{i=1}^{P} g^t_{i,k}(\iter{i}{k-1}{t}) }|\h{t}{k-1}}\\
&\le (1-2\eta_k^t \mu(1-\eta_k^t L))  \mynorm[2]{\iter{p}{k-1}{t}- \breve{\ww}_{k-1}^t }\\&+2\eta_k^t\ps{ \iter{p}{k-1}{t}- \breve{\ww}_{k-1}^t}{F'(\breve{\ww}_{k-1}^t) - \frac{1}{P} \sum_{i=1}^{P} F'(\iter{i}{k-1}{t}) }\\
&+ \expe{(\eta_k^t)^2 \mynorm[2]{ (g^t_{p,k}-F')(\iter{p}{k-1}{t}) -\frac{1}{P} \sum_{i=1}^{P} (g^t_{i,k}-F')(\iter{i}{k-1}{t}) }|\h{t}{k-1}}.
\end{align*}
Summing over $ p\in\un{P} $:
\begin{align*}
\sum_{p=1}^{P}\expe{\mynorm[2]{\iter{p}{k}{t}-\breve{\ww}_k^t}|\h{t}{k-1}}
&\le (1-\eta_k^t \mu) \sum_{p=1}^{P} \mynorm[2]{\iter{p}{k-1}{t}- \breve{\ww}_{k-1}^t }\\&+2\eta_k^t\ps{\underbrace{\sum_{p=1}^{P} (\iter{p}{k-1}{t}- \breve{\ww}_{k-1}^t)}_{=0}}{F'(\breve{\ww}_{k-1}^t) - \frac{1}{P} \sum_{i=1}^{P} F'(\iter{i}{k-1}{t}) }\\
&+\sum_{p=1}^{P} \expe{(\eta_k^t)^2 \mynorm[2]{ (g^t_{p,k}-F')(\iter{p}{k-1}{t}) -\frac{1}{P} \sum_{i=1}^{P} (g^t_{i,k}-F')(\iter{i}{k-1}{t}) }|\h{t}{k-1}}.
\end{align*}
If we denote $ \delta_k^t= \frac{1}{P}\sum_{p=1}^{P}\expe{\mynorm[2]{\iter{p}{k}{t}-\breve{\ww}_k^t}}$, we thus have $ \delta_0=0 $ and
\begin{align*}
\delta_k^t
&\le (1-\eta_k^t \mu)\delta_{k-1}^t+\frac{1}{P} \sum_{p=1}^{P} \expe{(\eta_k^t)^2 \mynorm[2]{ (g^t_{p,k}-F')(\iter{p}{k-1}{t}) -\frac{1}{P} \sum_{i=1}^{P} (g^t_{i,k}-F')(\iter{i}{k-1}{t}) }|\h{t}{k-1}}.\\
&\le \frac{1}{P} \sum_{p=1}^{P}  \sum_{j=1}^{k} \expe{(\eta_j^t)^2 \mynorm[2]{( g^t_{p,j}-F')(\iter{p}{j-1}{t}) -\frac{1}{P} \sum_{i=1}^{P}( g^t_{i,j}-F')(\iter{i}{j-1}{t}) }} \prod_{s=j+1}^{k} (1-\eta_s^t \mu)\\
&\le   \sum_{j=1}^{k} \expe{(\eta_j^t)^2 \mynorm[2]{( g^t_{1,j}-F')(\iter{1}{j-1}{t}) -\frac{1}{P} \sum_{i=1}^{P}( g^t_{i,j}-F')(\iter{i}{j-1}{t}) }} \prod_{s=j+1}^{k} (1-\eta_s^t \mu)\\
&\le   \sum_{j=1}^{k} \expe{(\eta_j^t)^2 \mynorm[2]{( g^t_{1,j}-F')(\iter{1}{j-1}{t}) }} \prod_{s=j+1}^{k} (1-\eta_s^t \mu).
\end{align*}
Note that everything is tight until the last line for $ P=1 $ ( then for all $ k $, $ \delta^t_k=0 $). Under Assumption~\Cref{hyp:addit_noise}, we thus have:
\begin{align*}
\delta_k^t
&\le   \sum_{j=1}^{k} (\eta_j^t)^{2} \siginf \prod_{s=j+1}^{k} (1-\eta_s^t \mu).
\end{align*}
This concludes the proof.

\section{Convergence guaranties for the fourth order moment.} \label{app:aux4}
In this section, we prove several Lemmas that allow to control the fourth order moment of the iterate.
	While controlling the second order moment is sufficient for quadratic functions as no ``residual'' term appears in \Cref{eq:dec_Pol_Jud} (the ``residual'' corresponds to the rest of a linear expansion of the gradient, which is thus exact for a quadratic function), in the general case, we also need to control the 4th order moment.

We first give guarantees for the inner iterates (within a phase) in \Cref{secmom4inner}, then in the local SGD framework in \Cref{sec:moment4gen}. 
\subsection{Inner Iteration Lemmas}\label{secmom4inner}
Here, we can use the following Lemma from \cite{Die_Dur_2018}, that gives a recursion for the 4th order moment.

\begin{lemma}\label{lem:moment_bound}
	Under the Assumptions ~\Cref{hyp:strong_convex},~\Cref{hyp:regularity},~\Cref{ass:def_filt},~\Cref{ass:lip_noisy_gradient_AS} for th $4^{th}$-order moment, assuming  $\eta\le \frac{1}{18 L}$ we have,
	\begin{align*}
	\e{(\mynorm{\iter{i}{k}{t}-\ws})^{4}}^{1/2} &\leq \big(1-\eta\mu\big)\e{\mynorm{\iter{i}{k-1}{t}-\ws}^4}^{1/2} +  20\eta^2 \sigma^{2} \\
	\e{\mynorm{\iter{i}{k}{t}-\ws}^{4}}^{1/2} &\leq (1-\eta\mu)^{k}  \e{{\mynorm{\iter{i}{0}{t}-\ws}}^{4}}^{1/2} +  \frac{20\eta \sigma^2}{\mu}.
	\end{align*}
\end{lemma}

In the mini-batch setting, we have of course the same result with a variance reduction:
\begin{lemma}\label{lem:MB_moment_bound}
	Under the Assumptions ~\Cref{hyp:strong_convex},~\Cref{hyp:regularity},~\Cref{ass:def_filt},~\Cref{ass:lip_noisy_gradient_AS} for th $4^{th}$-order moment for mini-batch averaging we have, assuming  $\eta\le \frac{1}{18 L}$ we have, 
	\begin{align*}
	\e{\norm{\hat{\ww}^t-\ws}^{4}}^{1/2} &\leq \big(1-\eta\mu\big)\e{\norm{\hat{\ww}^{t-1}-\ws}^{4}}^{1/2} +  \frac{20\eta^2}{P} \sigma^{2}\\
	\e{\norm{\hat{\ww}^t-\ws}^{4}}^{1/2} &\leq \big(1-\eta\mu\big)^t\norm{\ww^0-\ws}^2 +  \frac{20\eta}{P\mu} \sigma^{2}.
	\end{align*}
\end{lemma}

Analogous to \Cref{lem:moment_bound} we have the following result for fourth order moments,
\begin{lemma}\label{lem:moment_bound_DSS}
	Under the Assumptions ~\Cref{hyp:strong_convex},~\Cref{hyp:regularity},~\Cref{ass:def_filt},~\Cref{ass:lip_noisy_gradient_AS} for th $4^{th}$-order moment, assuming  $\eta\le \frac{1}{18 L}$ we have,
	\begin{align*}
	\e{\norm{\ww_{i,k}^t-\ws}^{4}}^{1/2} &\leq \big(1-\eta_k^t\mu\big)\e{\norm{\ww_{i,k-1}^t-\ws}^{4}}^{1/2} +  20\eta^2 \sigma^{2} \\
	\e{\norm{\ww_{i,k}^t-\ws}^{4}}^{1/2} &\leq \prod_{j=1}^{k}(1-\eta_j^t\mu)\norm{\ww^0-\ws}^{2} + 20 \sigma^{2}\sum_{j=1}^{k}\prod_{l=j+1}^{k}(1-\mu\eta_{l}^t)(\eta_j^t)^2.
	\end{align*}
\end{lemma}
Similarly for mini-batch analogous to \Cref{lem:MB_moment_bound},
\begin{lemma}\label{lem:MB_moment_bound_DSS}
	Under the Assumptions ~\Cref{hyp:strong_convex},~\Cref{hyp:regularity},~\Cref{ass:def_filt},~\Cref{ass:lip_noisy_gradient_AS} for th $4^{th}$-order moment for mini-batch averaging  and decreasing step size we have, assuming  $\eta\le \frac{1}{18 L}$ we have, 
	\begin{align*}
	\e{\norm{\hat{\ww}^t-\ws}^{4}}^{1/2} &\leq \big(1-\eta^t\mu\big)\e{\norm{\hat{\ww}^{t-1}-\ws}^{4}}^{1/2} +  \frac{20\eta^2}{P} \sigma^{2}\\
	\e{\norm{\hat{\ww}^t-\ws}^{4}}^{1/2} &\leq \prod_{j=1}^{t}\big(1-\eta^j\mu\big)\norm{\hat{\ww}^0-\ws}^{2} +  \frac{20\sigma^{2}}{P}\sum_{j=1}^{t}\prod_{l=j+1}^{t}(1-\mu\eta^l)(\eta^j)^2 .
	\end{align*}
\end{lemma}

The proof is included for completeness and because the same proof technique is used afterwards in \Cref{sec:moment4gen}.

\begin{proof}
	For $i\ \in [P] $, $k\ \in [N_t]$ and $t\in [C]$ we define the notation $ \phi_{i,k}^t = \mynorm{\iter{i}{k}{t}-\ws}$.
	We have that,
	\begin{align*}
	(\phi_{i,k}^t)^{4}&= \big(\norm{\ww_{i,k-1}^t -\ws}^2  -2\eta\inner{\gg_{i,k}^t(\iter{i}{k-1}{t})} {\iter{i}{k-1}{t} -\ws}+\eta^2   \norm{\gg_{i,k}^t(\iter{i}{k-1}{t})}^{2}\big)^2 \\
	&=\big((\phi_{i,k-1}^t)^{2}  -2\eta\inner{\gg_{i,k}^t(\iter{i}{k-1}{t})}{\iter{i}{k-1}{t} -\ws}+\eta^{2 }  \norm{g_{i,k}^{t}(\iter{i}{k-1}{t})}^{2} \big)^{2}\\
	&=(\phi_{i,k-1}^t)^4  -4\eta  (\phi_{i,k-1}^t)^2 \inner{\gg_{i,k}^t(\iter{i}{k-1}{t})}{\iter{i}{k-1}{t} -\ws}\\
	& +4\eta^{2} \inner {g_{i,k}^t(\iter{i}{k-1}{t})}{\iter{i}{k-1}{t} -\ws}^2
	+2\eta^2(\phi_{i,k-1}^t)^2\norm{\gg_{i,k}^t(\iter{i}{k-1}{t})}^{2} \\
	&-4 \eta^3 \inner{\gg_{i,k}^t(\iter{i}{k-1}{t})}{\iter{i}{k-1}{t} -\ws}\norm{\gg_{i,k}^t(\iter{i}{k-1}{t})}^2  + \eta^4 \norm{\gg_{i,k}^t(\iter{i}{k-1}{t})}^{4}.
	\end{align*}
	Moreover, 
	\begin{align}
	\e{\norm{\gg_{i,k}^t(\iter{i}{k-1}{t})}^{p} | \mathbb{H}_{k-1}^t} &\leq 2^{p-1} \big(\e{\norm{ \gg_{i,k}^t(\iter{i}{k-1}{t})- \gg_{i,k}^t(\ws)}^{p} | \mathbb{H}_{k-1}^t}  + \e{\norm{\gg_{i,k}^t(\ws)}^{p} | \mathbb{H}_{k-1}^t }\big)\nonumber\\
	&\leq  2^{p-1} \big(\e{\norm{\gg_{i,k}^t(\iter{i}{k-1}{t})- \gg_{i,k}^t(\ws)}^{p}}   + \e{\norm{\gg_{i,k}^t(\ws)}^{p}| \mathbb{H}_{k-1}^t }\big) \nonumber\\
	&\leq  2^{p-1} \big( \norm{\gg_{i,k}^t(\iter{i}{k-1}{t})- \gg_{i,k}^t(\ws)}^{p} + \sigma^{p}\big) , \label{eq:momentp_gradlip}
	\end{align}
	Where we have used at the first line Minkowski's inequality and the fact that $x\mapsto x^p$ is convex on $\R^{+}$ for $p=1,\dots, 4$ thus $(x+y)^{p}\le 2^{p-1} (x^p+ y^p)$, and at the last line the Assumption~\Cref{ass:lip_noisy_gradient_AS} on the noise : $\e{\norm{f_{i,k}^t(\ws)}^{p}| \mathbb{H}_{k-1}^t} \leq \sigma^{p}$. 
	
	This leads to 
	\begin{align*}
	\blacktriangle&:=	\e{(\phi_{i,k}^t)^{4}|\mathbb{H}_{k-1}^t}  \\
	&\leq (\phi_{i,k-1}^t)^{4}  -4\eta(\phi_{i,k-1}^t)^{2}\e{ \inner{ \gg_{i,k}^t(\iter{i}{k-1}{t})}{ \iter{i}{k-1}{t} -\ws} | \mathbb{H}_{k-1}^t}\\ 
	&+ 4 \eta^{2}  \e{\inner{ \gg_{i,k}^t(\iter{i}{k-1}{t})}{ \iter{i}{k-1}{t} -\ws}^{2}|\mathbb{H}_{k-1}^t} + 2 \eta^{2}(\phi_{i,k-1}^t)^{2}\e{ \norm{\gg_{i,k}^t(\iter{i}{k-1}{t})}^{2} |\mathbb{H}_{k-1}^t}\\ 
	&-4 \eta^3  \e{ \inner{ \gg_{i,k}^t(\iter{i}{k-1}{t})}{ \iter{i}{k-1}{t} -\ws} \norm{\gg_{i,k}^t(\iter{i}{k-1}{t})}^{2} |\mathbb{H}_{k-1}^t]} + \eta^4 \e{ \norm{\gg_{i,k}^t(\iter{i}{k-1}{t})}^{4} |\mathbb{H}_{k-1}^t} \\
	&\leq (\phi_{i,k-1}^t)^{4}  -4  \eta  (\phi_{i,k-1}^t)^{2}  \inner{ F'(\iter{i}{k-1}{t})}{ \iter{i}{k-1}{t} -\ws} + 4 \eta^{2}  \e{\norm{\gg_{i,k}^t(\iter{i}{k-1}{t})}^{2} (\phi_{i,k-1}^t)^{2}|\mathbb{H}_{k-1}^t}\\
	&+2 \eta^{2}  (\phi_{i,k-1}^t)^{2}  \e{\norm{\gg_{i,k}^t(\iter{i}{k-1}{t})}^{2} |\mathbb{H}_{k-1}^t} +4 \eta^3  \phi_{i,k-1}^t  \e{\norm{\gg_{i,k}^t(\iter{i}{k-1}{t})}^{3} |\mathbb{H}_{k-1}^t}\\
	&+ \eta^4 \e{\norm{\gg_{i,k}^t(\iter{i}{k-1}{t})}^{4} |\mathbb{H}_{k-1}^t} \\
	&\leq (\phi_{i,k-1}^t)^{4}  -4  \eta  (\phi_{i,k-1}^t)^{2}  \inner{ F'(\iter{i}{k-1}{t})}{ \iter{i}{k-1}{t}-\ws} + 12\eta^{2} \sigma^{2} (\phi_{i,k-1}^t)^{2} + 16 \eta^3  \phi_{i,k-1}^t \sigma^3 + 8 \eta^4 \sigma^{4}\\
	&+ 12 \eta^{2}  (\phi_{i,k-1}^t)^{2}  \e{  \norm{ \gg_{i,k}^t(\iter{i}{k-1}{t})- \gg_{i,k}^t(\ws)}^{2} |\mathbb{H}_{k-1}^t}\\
	&+ 16 \eta^3  \phi_{i,k-1}^t  \e{ \norm{ \gg_{i,k}^t(\iter{i}{k-1}{t})- \gg_{i,k}^t(\ws)}^{3} |\mathbb{H}_{k-1}^t}  +  8\eta^4 \e{  \norm{ \gg_{i,k}^t(\iter{i}{k-1}{t})- \gg_{i,k}^t(\ws)}^{4} |\mathbb{H}_{k-1}^t} .
	\end{align*}
	Above we have used Cauchy Schwartz inequality several times for the second inequality and equation~\eqref{eq:momentp_gradlip} for the third one. 
	\begin{align*}
	\bigstar	&:=\e{ (\phi_{i,k}^t)^{4} |\mathbb{H}_{k-1}^t} \nonumber\\
	&\leq (\phi_{i,k-1}^t)^{4}  -4  \eta  (\phi_{i,k-1}^t)^{2}  \inner{ F'(\iter{i}{k-1}{t})}{ \iter{i}{k-1}{t} -\ws} + 12 \eta^{2} L (\phi_{i,k-1}^t)^{2} \inner{ F'(\iter{i}{k-1}{t})} {\iter{i}{k-1}{t} -\ws} \nonumber\\
	&+16 \eta^3 L^{2} (\phi_{i,k-1}^t)^2  \inner{ F'(\iter{i}{k-1}{t})}{\iter{i}{k-1}{t} -\ws}   + 8 \eta^4 L^{3 }(\phi_{i,k-1}^t)^2  \inner{ F'(\iter{i}{k-1}{t})}{ \iter{i}{k-1}{t} -\ws} \nonumber\\
	& + 12 \eta \sigma^{2} (\phi_{i,k-1}^t)^{2} + 8 \eta^2 \sigma^2  (\phi_{i,k-1}^t)^2 + 8 \eta^4 \sigma^4 + 8 \eta^4 \sigma^{4}\nonumber\\
	&= (\phi_{i,k-1}^t)^{4} +( -4  \eta + 12 \eta^{2} L+ 16 \eta^3 L^{2}+ 8 \eta^4 L^{3}) (\phi_{i,k-1}^t)^{2}  \inner{ F'(\iter{i}{k-1}{t})}{\iter{i}{k-1}{t} -\ws}    \nonumber\\
	& + (12\eta^2 \sigma^{2} +8 \eta^2 \sigma^{2})  (\phi_{i,k-1}^t)^{2} +   16 \eta^4 \sigma^{4}\nonumber\\
	&\leq (\phi_{i,k-1}^t)^{4}  - 4  \eta  (1- 9 \eta  L ) (\phi_{i,k-1}^t)^{2}  \inner{ F'(\iter{i}{k-1}{t})}{\iter{i}{k-1}{t} -\ws}   + 20\eta^2 \sigma^{2} (\phi_{i,k-1}^t)^{2} +   16 \eta^4 \sigma^{4} \label{eq:expansion_mom4_classique}.
	\end{align*}
	Above we used $\eta L \leq 1$ in the last line. Finally, using strong convexity, we have:
	\begin{align*}
	\e{(\phi_{i,k}^t)^{4}|\mathbb{H}_{k-1}^t}  &\leq   \big(1 - 4\eta\mu(1-9\eta L)\big) (\phi_{i,k-1}^t)^{4} + 20\eta^2 \sigma^{2} (\phi_{i,k-1}^t)^{2} + 16 \eta^4 \sigma^{4},
	\end{align*}
	Now $\e{(\phi_{i,k-1}^t)^{2}}\leq \e{(\phi_{i,k-1}^t)^{4}}^{1/2}$ by Jensen's inequality. Also since we assume $\eta \leq \frac{1}{9L}$ and $\frac{\mu}{L} \leq 1$ we can obtain $(1 - 4\eta\mu(1-9\eta L))^{1/2} \geq (1 - 4\eta\mu)^{1/2} \geq (1 - \frac{4\mu}{9L})^{1/2} \geq (1 - \frac{4}{9})^{1/2} \geq 1/2$. This finally leads to $ 20\eta^2 \sigma^{2} \e{(\phi_{i,k-1}^t)^{2}} \leq (1 - 4\eta\mu(1-9\eta L))^{1/2} \e{(\phi_{i,k-1}^t)^{4}}^{1/2} 40 \eta^2 \sigma^{2} $, which can be used below to obtain  
	\begin{align*}
	\e{(\phi_{i,k}^t)^{4}|\mathbb{H}_{k-1}^t} &\leq \big(1 - 4  \eta  \mu  (1-9\eta L)\big) \e{(\phi_{i,k-1}^t)^{4}} + 20\eta^2 \sigma^{2} \e{(\phi_{i,k-1}^t)^{2}} +  16 \eta^4 \sigma^{4} \\
	&\leq \bigg(\big(1-4\eta\mu(1-9\eta L)\big)^{1/2}\e{(\phi_{i,k-1}^t)^{4}}^{1/2} +  20 \eta^2 \sigma^{2} \bigg)^{2}\\
	\e{(\phi_{i,k}^t)^{4}}^{1/2} &\leq \big(1 - 2\eta\mu(1-9\eta L)\big) \e{(\phi_{i,k-1}^t)^{4}}^{1/2} +  20\eta^2\sigma^{2}.
	\end{align*}
	This Concludes the proof. 
\end{proof}

\subsection{Proof of ~\Cref{lem:moment4genF}}\label{sec:moment4gen}
In this section, we prove the following Lemma, which is necessary to conclude the proof for the second set of Assumptions in \Cref{th:LocalSGDconv}. Indeed, we need to control the moment of order 4 to be able to control the residual term that arises from linear expansion of the gradient around $ \ws $.

\begin{lemma}\label{lem:moment4genF}
	There exist absolute constants $ C_4, D_4, E_4 $, such that if $ \eta_k^t L\le \frac{1}{C_4} $:
	\begin{align}
	\expe{ \mynorm{\breve{\ww}_{k+1}^t-\ws} ^4}^{1/2}&\le
	(1- \eta_k^t\mu )\expe{ \mynorm{\breve{\ww}_{k}^t-\ws} ^4}^{1/2}  + D_4 (\eta_k^t)^{2}\frac{\siginf}{P}\nonumber\\
	&+ E_4  \eta_{k+1}^t  \mynorm{\breve{\ww}_{k}^t-\ws}  \mynorm{F'(\breve{\ww}_k^t)-\frac{1}{P}\sum_{p=1}^{P} F'(\iter{p}{k}{t})}.
	\end{align}
\end{lemma}
In other words, $ \expe{ \mynorm{\breve{\ww}_{k+1}^t-\ws} ^4}^{1/2} $ satisfies the same recursion as $ \expe{ \mynorm{\breve{\ww}_{k+1}^t-\ws} ^2}$, as this equation is the same as \Cref{eq:equivmom4} (up to absolute constants).

\begin{proof}
	This proof combines element from the classical bound for the fourth order moment, and from the proof of \Cref{lem:lemme_aux_conv_inner_gen_case}, which addresses the similar setting but only for the second order moment.
	We start from the definition of $\breve{\ww}_{k+1}^t  $:
	\begin{align}
	{\norm{\breve{\ww}_{k+1}^t-\ww^\star}^2}&\le
	\norm{\breve{\ww}_{k}^t-\ww^\star}^2  -2\eta_{k+1}^t\inner{\breve{\ww}_k^t-\ww^\star}{\frac{1}{P}\sum_{i=1}^{P}g_{i,k+1}^t(\breve{\ww}_k^t)}\nonumber\\
	&+ (\eta_{k+1}^t)^2{\norm{\frac{1}{P}\sum_{i=1}^{P}g_{i,k+1}^t(\ww_{i,k}^t)}^2}\nonumber\\
	&+2\eta_{k+1}^t\inner{\breve{\ww}_k^t-\ww^\star}{\frac{1}{P}\sum_{i=1}^{P}g_{i,k+1}^t(\breve{\ww}_k^t)-\frac{1}{P}\sum_{p=1}^{P} F'(\iter{p}{k}{t})}.
	\end{align}
	Thus, squaring this equation we get, denoting $  \breve\phi_{k}^t = \mynorm{\breve{\ww}_{k}^t-\ws} $:
	\begin{align*}
	(\breve\phi_{k+1}^t) ^4&\le
	(\breve\phi_{k}^t )^4  
	-4 (\breve\phi_{k}^t)^2\eta_{k+1}^t\inner{\breve{\ww}_k^t-\ww^\star}{\frac{1}{P}\sum_{i=1}^{P}g_{i,k+1}^t(\breve{\ww}_k^t)}\nonumber\\
	&+2 (\breve\phi_{k}^t )^2  (\eta_{k+1}^t)^2{\norm{\frac{1}{P}\sum_{i=1}^{P}g_{i,k+1}^t(\ww_{i,k}^t)}^2}\nonumber\\
	&+4 (\breve\phi_{k}^t )^2 \eta_{k+1}^t\inner{\breve{\ww}_k^t-\ww^\star}{\frac{1}{P}\sum_{i=1}^{P}g_{i,k+1}^t(\breve{\ww}_k^t)-\frac{1}{P}\sum_{p=1}^{P} F'(\iter{p}{k}{t})}\\
	& + 3 (\eta_{k+1}^t)^{2}\inner{\breve{\ww}_k^t-\ww^\star}{\frac{1}{P}\sum_{i=1}^{P}g_{i,k+1}^t(\breve{\ww}_k^t)}^{2}\\
	&+3  (\eta_{k+1}^t)^4{\norm{\frac{1}{P}\sum_{i=1}^{P}g_{i,k+1}^t(\ww_{i,k}^t)}^4}\\
	&+ 3 (2\eta_{k+1}^t)^{2}\inner{\breve{\ww}_k^t-\ww^\star}{\frac{1}{P}\sum_{i=1}^{P}g_{i,k+1}^t(\breve{\ww}_k^t)-\frac{1}{P}\sum_{p=1}^{P} F'(\iter{p}{k}{t})}^{2},
	\end{align*}
	formally, we have used $ (a+b+c+d)^{2}\le a^2+ 2ab+2ac+2ad+ 3 b^2+ 3 c^2+ 3 d^2 $.
	
	That is, conditioning on the past, and using Assumption~\Cref{ass:lip_noisy_gradient_AS} (cocoercivity and the fact that $ g_k^t $ is a.s.~$ L$-Lipshitz):
	\begin{align}
	\expe{(\breve\phi_{k+1}^t) ^4|\h{t}{k}}&\le
	(\breve\phi_{k}^t )^4  
	-4 (\breve\phi_{k}^t)^2\eta_{k+1}^t (1-\eta_k^t L)\inner{\breve{\ww}_k^t-\ww^\star}{F'(\breve{\ww}_k^t)}\nonumber\\
	&+2 (\breve\phi_{k}^t )^2  (\eta_{k+1}^t)^2 \expe{\norm{\frac{1}{P}\sum_{i=1}^{P}g_{i,k+1}^t(\ww_{i,k}^t)-F'(\ww_{i,k}^t)}^2|\h{t}{k}}\nonumber\\
	&+4 (\breve\phi_{k}^t )^2 \eta_{k+1}^t\inner{\breve{\ww}_k^t-\ww^\star}{F'(\breve{\ww}_k^t)-\frac{1}{P}\sum_{p=1}^{P} F'(\iter{p}{k}{t})}\nonumber\\
	& + 3 (\eta_{k+1}^t)^{2}\inner{\breve{\ww}_k^t-\ww^\star}{\frac{1}{P}\sum_{i=1}^{P}F'(\breve{\ww}_k^t)} L (\breve\phi_{k}^t)^2\nonumber\\
	&+6  (\eta_{k+1}^t)^4\expe{\norm{\frac{1}{P}\sum_{i=1}^{P}g_{i,k+1}^t(\ww_{i,k}^t)-F'(\ww_{i,k}^t)}^4|\h{t}{k}}\nonumber\\
	&+ 6  (\eta_{k+1}^t)^4 L^{2}(\breve\phi_{k}^t)^2 \inner{\breve{\ww}_k^t-\ww^\star}{F'(\breve{\ww}_k^t)}\nonumber\\
	&+ 3 (2\eta_{k+1}^t)^{2}\inner{\breve{\ww}_k^t-\ww^\star}{\frac{1}{P}\sum_{i=1}^{P}F'(\breve{\ww}_k^t)-\frac{1}{P}\sum_{p=1}^{P} F'(\iter{p}{k}{t})}^{2}\nonumber.
	\end{align}
	Rearranging terms and using the uniform upper bound on the 4-th moment of the noise \Cref{hyp:variancebound}, we have:
	\begin{align}
	\expe{(\breve\phi_{k+1}^t) ^4|\h{t}{k}}&\le
	(\breve\phi_{k}^t )^4  
	-4 (\breve\phi_{k}^t)^2\eta_{k+1}^t (1-\eta_k^t L -3\eta_k^t L  - 6  (\eta_{k+1}^t)^4 L^{2} )\inner{\breve{\ww}_k^t-\ww^\star}{F'(\breve{\ww}_k^t)}\nonumber\\
	&+2 (\breve\phi_{k}^t )^2  (\eta_{k+1}^t)^2\frac{\siginf}{P}+6  (\eta_{k+1}^t)^4 \frac{\sigma_\infty^{4}}{P^2}\nonumber\\
	&+4 (\breve\phi_{k}^t )^2 \eta_{k+1}^t\inner{\breve{\ww}_k^t-\ww^\star}{F'(\breve{\ww}_k^t)-\frac{1}{P}\sum_{p=1}^{P} F'(\iter{p}{k}{t})}\nonumber\\
	&+ 3 (2\eta_{k+1}^t)^{2}\expe{\inner{\breve{\ww}_k^t-\ww^\star}{\frac{1}{P}\sum_{i=1}^{P}g_{i,k+1}^t(\breve{\ww}_k^t)-\frac{1}{P}\sum_{p=1}^{P} F'(\iter{p}{k}{t})}^{2}|\h{t}{k}}\label{eq:moment4exp}.
	\end{align}
	The first 2 lines of \Cref{eq:moment4exp} correspond to the expansion in \Cref{eq:expansion_mom4_classique} (the constants are slightly different because we use a uniform bound on the gradient instead of co-coercivity). The last two lines correspond to the residual term, for which we will use \Cref{lem:upperboundondivergence}.
	
	We have:
	\begin{align*}
	& \hspace{-5em}4 (\breve\phi_{k}^t )^2 \eta_{k+1}^t\inner{\breve{\ww}_k^t-\ww^\star}{F'(\breve{\ww}_k^t)-\frac{1}{P}\sum_{p=1}^{P} F'(\iter{p}{k}{t})}\nonumber\\
	& \hspace{-5em}+6 (2\eta_{k+1}^t)^{2}\expe{\inner{\breve{\ww}_k^t-\ww^\star}{\frac{1}{P}\sum_{i=1}^{P}g_{i,k+1}^t(\breve{\ww}_k^t)-\frac{1}{P}\sum_{p=1}^{P} F'(\iter{p}{k}{t})}^{2}|\h{t}{k}} \\
	& \le 4 (\breve\phi_{k}^t )^3 \eta_{k+1}^t\mynorm{F'(\breve{\ww}_k^t)-\frac{1}{P}\sum_{p=1}^{P} F'(\iter{p}{k}{t})}\\
	&+ 6 (2\eta_{k+1}^t)^{2} L \mynorm[3]{\breve{\ww}_k^t-\ww^\star}\mynorm{\frac{1}{P}\sum_{i=1}^{P}F'(\breve{\ww}_k^t)-\frac{1}{P}\sum_{p=1}^{P} F'(\iter{p}{k}{t})}\\
	&=  (\breve\phi_{k}^t )^3 \eta_{k+1}^t (4+24\eta_k^t L)\mynorm{F'(\breve{\ww}_k^t)-\frac{1}{P}\sum_{p=1}^{P} F'(\iter{p}{k}{t})}.
	\end{align*}
	As a result, there exist absolute constants (``numbers'') $ C_4, D_4, E_4 $, such that if $ \eta_k^t L\le \frac{1}{C_4} $:
	\begin{align}
	\expe{(\breve\phi_{k+1}^t) ^4}^{1/2}&\le
	(1- \eta_k^t\mu )\expe{(\breve\phi_{k}^t )^4}^{1/2}  + D_4 (\eta_k^t)^{2}\frac{\siginf}{P}\nonumber\\
	&+ E_4  \eta_{k+1}^t \expe{ (\breve\phi_{k}^t )\mynorm{F'(\breve{\ww}_k^t)-\frac{1}{P}\sum_{p=1}^{P} F'(\iter{p}{k}{t})}}.
	\end{align}
	This is the result of the Lemma.
\end{proof}

\section{Main error decomposition}\label{app:errdec}
\subsection{General decomposition}
In this section, we prove the following decomposition for the on-line setting.

\begin{lemma}{\label{lem:parallel_decomposition}}
	Under the differentiability of \Cref{hyp:regularity} we have\footnote{Note that after the final iteration of the phase the learning rate (which the algorithm uses nowhere) corresponds to the first learning rate for the next phase. This anomaly in notation is a direct result of us considering the ghost process, which runs continuously till the end.}, 
	\begin{align*}
	F''(\ww^\star)(\overline{\overline{\ww}}^C - \ws)  &= \frac{P\left(\ww^0-\ws\right)}{T\eta_1^1} - \frac{P\left(\hat{\ww}^C-\ws\right)}{T\eta_{N^C+1}^C} -\frac{1}{T}\sum_{t=1}^{C}\sum_{k=1}^{N^t}\sum_{i=1}^{P}\left(\ww_{i,k}^t -\ws \right)\left(\frac{1}{\eta_k^t} - \frac{1}{\eta_{k+1}^t} \right)\\
	&+ \frac{1}{T}\sum_{t=1}^{C}\sum_{k=1}^{N^t}\sum_{i=1}^{P}\delta_{i,k}^t + \frac{1}{T}\sum_{t=1}^{C}\sum_{k=1}^{N^t}\sum_{i=1}^{P}\xi_{i,k}^t,
	\end{align*}
	where $\xi_{i,k}^t=F'(\ww_{i,k-1}^t)-\gg_{i,k}^t(\ww_{i,k-1}^t)$ and $\delta_{i,k}^t=F''(\ww^\star)(\ww_{i,k-1}^t-\ww^\star)-F'(\ww_{i,k-1}^t)$.
\end{lemma}
\begin{proof}
	Below, we have $\gg_{i,k}^t(\ww_{i,k-1}^t)$ as the stochastic gradient at step $k$ on machine $i$ for communication phase $t$. After adding and subtracting few quantities and rearranging we have,
	\begin{align*}
	&\ww_{i,k}^t = \ww_{i,k-1}^t - \eta_{k}^t \gg_{i,k}^t(\ww_{i,k-1}^t)\\
	&\ww_{i,k}^t = \ww_{i,k-1}^t - \eta_{k}^t F'(\ww_{i,k-1}^t) + \eta_{k}^t\big(F'(\ww_{i,k-1}^t)-\gg_{i,k}^t(\ww_{i,k-1}^t)\big)\\
	&\ww_{i,k}^t = \ww_{i,k-1}^t - \eta_{k}^t F'(\ww_{i,k-1}^t) + \eta_{k}^t\delta_{i,k}^t +\eta F''(\ww^\star)(\ww_{i,k-1}^t-\ww^\star)-\eta_{k}^t F''(\ww^\star)(\ww_{i,k-1}^t-\ww^\star)\\
	&\ww_{i,k}^t = \ww_{i,k-1}^t + \eta_{k}^t\xi_{i,k}^t +\eta_{k}^t \delta_{i,k}^t -\eta_{k}^t F''(\ww^\star)(\ww_{i,k-1}^t-\ww^\star).
	\end{align*}
	where $\xi_{i,k}^t$ and $\delta_{i,k}^t$ are respectively terms related to stochastic noise and quadratic residual. Obtaining the horizontal average over all the machines and recalling the definition of the ghost process $\breve{\ww}_k^t$ as defined above we have,
	\begin{align*}
	\frac{1}{P}\sum_{i=1}^{P}F''(\ww^\star)(\ww_{i,k-1}^t-\ww^\star)  &= \frac{1}{P}\sum_{i=1}^{P}\frac{1}{\eta_k^t}\big(\ww_{i,k-1}^t-\ww_{i,k}^t\big) +\frac{1}{P}\sum_{i=1}^{P}\delta_{i,k}^t +\frac{1}{P}\sum_{i=1}^{P}\xi_{i,k}^t\\
	F''(\ww^\star)(\breve{\ww}_{k-1}^t-\ww^\star)  &= \frac{\breve{\ww}_{k-1}^t-\breve{\ww}_{k}^t}{\eta_k^t}\ +\frac{1}{P}\sum_{i=1}^{P}\delta_{i,k}^t +\frac{1}{P}\sum_{i=1}^{P}\xi_{i,k}^t.
	\end{align*}
	Obtaining the vertical average over all the machines first within a communication phase and then among different phases we have,
	\begin{align*}
	\frac{1}{N^t}\sum_{k=1}^{N^t} F''(\ww^\star)(\breve{\ww}_{k-1}^t-\ww^\star)  &= \frac{1}{N^t}\sum_{k=1}^{N^t}\frac{\breve{\ww}_{k-1}^t-\breve{\ww}_{k}^t}{\eta_k^t} +\frac{1}{N^tP}\sum_{k=1}^{N^t}\sum_{i=1}^{P}\delta_{i,k}^t + \frac{1}{N^tP}\sum_{k=1}^{N^t}\sum_{i=1}^{P}\xi_{i,k}^t\\
	\frac{1}{\sum_{t=1}^{C}N^t}\sum_{t=1}^{C}\sum_{k=1}^{N^t} F''(\ww^\star)(\breve{\ww}_{k-1}^t-\ww^\star)  &= \frac{1}{\sum_{t=1}^{C}N^t}\sum_{t=1}^{C}\sum_{k=1}^{N^t}\frac{\breve{\ww}_{k-1}^t-\breve{\ww}_{k}^t}{\eta_k^t} +\frac{1}{P\sum_{t=1}^{C}N^t}\sum_{t=1}^{C}\sum_{k=1}^{N^t}\sum_{i=1}^{P}\delta_{i,k}^t\\
	&+ \frac{1}{P\sum_{t=1}^{C}N^t}\sum_{t=1}^{C}\sum_{k=1}^{N^t}\sum_{i=1}^{P}\xi_{i,k}^t.
	\end{align*}
	Now recalling the definitions for the overall iterate $\overline{\overline{\ww}}^C = \frac{1}{\sum_{t=1}^{C}N^t}\sum_{t=1}^{C}\sum_{k=1}^{N^t}\breve{\ww}_k^t$, $\hat{\ww}^t = \breve{\ww}_{N^t}^t$, the initial point $\hat{\ww}^0 = \ww^0$, and the total number of gradients $T=P\sum_{t=1}^{C}N^t$ as we have defined above. After making these changes and on rearranging we obtain,  
	\begin{align*}
	F''(\ww^\star)(\overline{\overline{\ww}}^C - \ws)  &= \frac{P}{T}\sum_{t=1}^{C}\sum_{k=1}^{N^t}\frac{\breve{\ww}_{k-1}^t-\breve{\ww}_{k}^t}{\eta_k^t} +\frac{1}{T}\sum_{t=1}^{C}\sum_{k=1}^{N^t}\sum_{i=1}^{P}\delta_{i,k}^t + \frac{1}{T}\sum_{t=1}^{C}\sum_{k=1}^{N^t}\sum_{i=1}^{P}\xi_{i,k}^t\\
	F''(\ww^\star)(\overline{\overline{\ww}}^C - \ws)  &= \frac{P\left(\ww^0-\ws\right)}{T\eta_1^1} - \frac{P\left(\hat{\ww}^C-\ws\right)}{T\eta_{N^C+1}^C} -\frac{P}{T}\sum_{t=1}^{C}\sum_{k=1}^{N^t}\left( \breve{\ww}_k^t -\ws \right)\left(\frac{1}{\eta_k^t} - \frac{1}{\eta_{k + 1}^t} \right)\\
	&+ \frac{1}{T}\sum_{t=1}^{C}\sum_{k=1}^{N^t}\sum_{i=1}^{P}\delta_{i,k}^t + \frac{1}{T}\sum_{t=1}^{C}\sum_{k=1}^{N^t}\sum_{i=1}^{P}\xi_{i,k}^t.
	\end{align*}
	Thus we have obtained the required result as,
	\begin{align*}
	F''(\ww^\star)(\overline{\overline{\ww}}^C - \ws)  &= \frac{P\left(\ww^0-\ws\right)}{T\eta_1^1} - \frac{P\left(\hat{\ww}^C-\ws\right)}{T\eta_{N^C+1}^C} -\frac{1}{T}\sum_{t=1}^{C}\sum_{k=1}^{N^t}\sum_{i=1}^{P}\left(\ww_{i,k}^t -\ws \right)\left(\frac{1}{\eta_k^t} - \frac{1}{\eta_{k+1}^t} \right)\\
	&+ \frac{1}{T}\sum_{t=1}^{C}\sum_{k=1}^{N^t}\sum_{i=1}^{P}\delta_{i,k}^t + \frac{1}{T}\sum_{t=1}^{C}\sum_{k=1}^{N^t}\sum_{i=1}^{P}\xi_{i,k}^t.
	\end{align*}
\end{proof}

\subsection{Bounding the noise term}
The stochastic noise term which appears above can be bounded using the following lemma,
\begin{lemma}\label{lem:xi_bound}
	Under the Assumptions ~\Cref{ass:def_filt}, ~\Cref{ass:lip_noisy_gradient_AS}, ~\Cref{hyp:variancebound} we have
	\begin{align*}
	\e{\norm{\xi_{i,k}^t}^2} \leq 2L^2\e{\norm{\ww_{i,k-1}^t\ww^\star}^2} + 2\sigma^2.
	\end{align*}
	\begin{proof}
		Using Assumptions ~\Cref{ass:def_filt}, ~\Cref{ass:lip_noisy_gradient_AS}, ~\Cref{hyp:variancebound} respectively we prove the result
		\begin{align*}
		\e{\norm{\xi_{i,k}^t}^2} &= \e{\norm{F'(\ww_{i,k-1}^t)-\gg_{i,k}^t(\ww_{i,k-1}^t)}^2} \leq \e{\norm{\gg_{i,k}^t(\ww_{i,k-1}^t)}^2} -\norm{F'(\ww_{i,k-1}^t)}^2\\
		&\leq 2\e{\norm{\gg_{i,k}^t(\ww_{i,k-1}^t)-\gg_{i,k}^t(\ww^\star)}^2} + 2\e{\norm{\gg_{i,k}^t(\ww^\star)}^2}
		\\ & \leq 2L^2\e{\norm{\ww_{i,k-1}^t-\ww^\star}^2} + 2\sigma^2.
		\end{align*}
	\end{proof}
\end{lemma}

\section{Proofs for OSA, MBA and Local-SGD in the finite horizon setting}\label{app:mainproofs}
In this Section and \Cref{sec:proofsonline} we prove convergence results for $ \expe{\mynorm{F''(\ws)(\itePRav-\ws)}} $. The proof technique is the one proposed by Polyak and Judisky in the original article on averaging~\cite{Pol_Jud_1992}. This proof technique has also been used in~\cite{Bac_Mou_2011,God_Saa_2017}. We notice here the following differences, that justify including the proofs:
\begin{enumerate}
	\item Polyak and Judisky were mainly interested in the asymptotic analysis, and the set of assumptions considered was different.
	\item In \cite{Bac_Mou_2011}, the authors prove comparable bounds in the case of bounded gradients. However, their analysis in the smooth and strongly convex setting is not optimal. Precisely, they use a sub-optimal upper bound when controlling the second order moments, that significantly worsens the subsequent proof. This point was underlined in~\cite{Nee_War_Sre_2014,Die_Dur_2018}. The result they provide under our set of assumptions is eventually 1) not optimal, 2) uselessly complex, and 3) only for serial-SGD.
	\item  In \cite{God_Saa_2017}, authors prove a result close to us, using a similar approach for one-shot averaging. Their bounds only apply to decaying step size. Moreover, they rely on the following asymptotic upper bound: $ \expe{\mynorm[2]{\ite_{i,k}^{t}-\ws}} \le C_1 \eta_k^t$: this bound is correct but the constant $ C_1 $ is "asymptotic" (see for e.g., \cite{rakhlin2012making}). On contrary, we use non-asymptotic upper bounds on the second order moment involved. As a consequence, our bounds are both simpler and tighter.
\end{enumerate}
\subsection{Technical Lemmas}
\begin{lemma}[Jensen's Inequality]\label{lem:tech_5}
	For $a_i \in \mathbb{R}^d$, $\norm{\frac{1}{P}\sum_{i=1}^{P}a_i}^2 \leq \frac{1}{P}\sum_{i=1}^{P}\norm{a_i}^2$.
\end{lemma}
\begin{proof}
	The result is an application of Jensen's inequality with the convex function $f(.)=\norm{.}^2$.
\end{proof}
\begin{lemma}[Minkowski's Inequality]\label{lem:tech_6}
	For $a_i \in \mathbb{R}^d$, $\e{\norm{\sum_{i=1}^{P}a_i}^2} \leq \left(\sum_{i=1}^{P}\e{\norm{a_i}^2}^{\frac{1}{2}}\right)^2$
\end{lemma}
\begin{proof}
	The inequality is an application of Minkowski's inequality (or simply triangle's inequality) with the norm $\norm{.}_E = \e{\norm{.}^2}^{\frac{1}{2}}$.
\end{proof}

\subsection{Proof of \Cref{prop:conver_Mini_Batch} (Mini-batch case)}

\Cref{lem:MB_bound} proves the first part of the proposition. We prove the second part of the proposition here following the approach by \cite{Pol_Jud_1992}. Using \Cref{lem:parallel_decomposition}, \Cref{lem:moment_bound} we can obtain an upper bound on $\e{\norm{F''(\ww^\star)(\overline{\overline{\ww}}^C-\ww^\star)}^2}$, which is in-fact a tighter quantity when compared to $\e{\norm{\overline{\overline{\ww}}^C-\ww^\star}^2}$. We prove the following lemma,
\begin{lemma}\label{lem:MB_semifinal}
	Under the Assumptions ~\Cref{hyp:strong_convex},~\Cref{hyp:regularity},~\Cref{ass:def_filt},~\Cref{ass:lip_noisy_gradient_AS},~\Cref{hyp:variancebound} we have,
	\begin{align*}
	\e{\norm{\nabla^2F(\ww^\star)(\ww-\ww^\star)}^2} \leq 4\sum_{i=1}^{5}A_{i,P,C}^2,
	\end{align*}
	where the terms are respectively,
	\begin{align*}
	&A_{1,P,C}^2=\frac{P^2}{T^2\eta^2}\norm{\ww^0-\ww^\star}^2, A_{2,P,C}^2=\frac{P^2}{T^2\eta^2}\left((1-\mu\eta)^C\norm{\ww^0 -\ww^\star}^2 + 2\sigma^2\frac{\eta}{\mu P}\right),\\
	&A_{3,P,C}^2=\frac{P^2M^2}{T^2\mu^2\eta^2}\left(\norm{\ww^0-\ws}^2 +  \frac{C20\eta^2}{P} \sigma^{2}\right)^2, A_{4,P,C}^2= \frac{2\sigma^2}{T},\\ 
	&A_{3,P,C}^2=\frac{2L^2P}{T^2}\left(\frac{1}{\mu\eta}\norm{\ww^0 -\ww^\star}^2 + 2\sigma^2\frac{\left(C\mu\eta-1+(1-\mu\eta)^C\right)}{\mu^2 P}\right).\\
	\end{align*}
	
\end{lemma}

\begin{proof}
	In order to upper bound the expectation we need to separately upper bound all the terms that appear in the result for \Cref{lem:parallel_decomposition}. But before that we can actually simplify the result with constant step size and using $N^t=1\ \forall t\in[C]$ as follows, 
	\begin{align*}
	F''(\ww^\star)(\overline{\overline{\ww}}^C - \ws)  &= \frac{\ww^0-\ws}{C\eta} - \frac{\hat{\ww}^C-\ws}{C\eta} + \frac{1}{T}\sum_{t=1}^{C}\sum_{i=1}^{P}\delta_{i,1}^t + \frac{1}{T}\sum_{t=1}^{C}\sum_{i=1}^{P}\xi_{i,1}^t.
	\end{align*}
	Now we bound each of the terms in the above decomposition one by one. For the first term,
	\begin{align*}
	\e{\norm{\frac{1}{C\eta}\left(\ww^0-\ws\right)}^2} &= \frac{P^2}{T^2\eta^2}\norm{\ww^0-\ww^\star}^2 = A_{1,P,C}^2.
	\end{align*}
	For the second term using \Cref{lem:MB_bound},
	\begin{align*}
	\e{\norm{\frac{1}{C\eta}\left(\hat{\ww}^C-\ws\right)}^2} &= \frac{P^2}{T^2\eta^2}\e{\norm{\ww^C_{MB}-\ww^\star}^2}\\
	&\leq \frac{P^2}{T^2\eta^2}\left(\prod_{k=1}^{C}(1-\mu\eta)\e{\norm{\ww^0 -\ww^\star}^2} + 2\sigma^2\frac{1}{P}\sum_{k=1}^{C}\prod_{l=k+1}^{C}(1-\mu\eta)\eta^2\right)\\
	&\leq \frac{P^2}{T^2\eta^2}\left((1-\mu\eta)^C\norm{\ww^0 -\ww^\star}^2 + 2\sigma^2\frac{1}{P}\left(\frac{1-(1-\mu\eta)^C}{\mu\eta}\right)\eta^2\right)\\
	&\leq \frac{P^2}{T^2\eta^2}\left((1-\mu\eta)^C\norm{\ww^0 -\ww^\star}^2 + 2\sigma^2\frac{\eta}{\mu P}\right) = A_{2,P,C}^2.
	\end{align*}
	For the third term using \Cref{lem:tech_5} and \Cref{lem:tech_6} we get,
	\begin{align*}
	\e{\norm{\frac{1}{T}\sum_{t=1}^{C}\sum_{i=1}^{P}\delta_{i,1}^t}^2}&=\frac{1}{T^2} \e{\norm{\sum_{t=1}^{C}\sum_{i=1}^{P}\left( F'(\ww_{i,0}^t)-F''(\ww^\star)(\ww_{i,0}^t-\ww^\star)\right)}^2}\\
	&\leq \frac{P}{T^2}\sum_{i=1}^{P} \e{\norm{\sum_{t=1}^{C}\left( F'(\hat{\ww}^{t-1})-F''(\ww^\star)(\hat{\ww}^{t-1}-\ww^\star)\right)}^2}\\
	&\leq \frac{P^2}{T^2}\left(\sum_{t=1}^{C}\sqrt{\e{\norm{\left(F'(\hat{\ww}^{t-1})-F''(\ww^\star)(\hat{\ww}^{t-1}-\ww^\star)\right)}^2}}\right)^2.\\
	\end{align*}
	Now using the upper bound from \Cref{hyp:regularity} followed by \Cref{lem:MB_moment_bound} we get,
	\begin{align*}
	\e{\norm{\frac{1}{T}\sum_{t=1}^{C}\sum_{i=1}^{P}\delta_{i,1}^t}^2} &\leq \frac{P^2M^2}{T^2} \left(\sum_{t=1}^{C}\sqrt{\e{\norm{\hat{\ww}^{t-1}-\ws}^4}}\right)^2\\
	&\leq \frac{P^2M^2}{T^2} \left(\sum_{t=1}^{C}\left((1-\eta\mu)^{t-1}\e{(\hat{\ww}^{0}-\ws)^4}^{1/2} +  \frac{20\eta}{P\mu} \sigma^{2}\right)\right)^2\\
	&\leq \frac{P^2M^2}{T^2} \left(\frac{1-(1-\eta\mu)^{C}}{\eta\mu}\e{(\hat{\ww}^{0}-\ws)^4}^{1/2} +  \frac{20C\eta}{P\mu} \sigma^{2}\right)^2\\
	&\leq \frac{P^2M^2}{T^2\mu^2\eta^2}\left(\norm{\ww^0-\ws}^2 +  \frac{20C\eta^2}{P} \sigma^{2}\right)^2 =A_{3,P,C}^2.
	\end{align*}
	
	For the fourth term, note that we are sampling i.i.d observations and thus the stochastic noise across all machines and iterations is independent and equal to zero in expectation (see \Cref{ass:def_filt}). This implies the first equation below while the second inequality is obtained using \Cref{lem:xi_bound},
	\begin{align*}
	\e{\norm{\frac{1}{T}\sum_{t=1}^{C}\sum_{i=1}^{P}\xi_{i,1}^t}^2} &= \frac{1}{T^2}\sum_{t=1}^{C}\sum_{i=1}^{P}\e{\norm{\xi_{i,1}^t}^2} \leq \frac{1}{T^2}\sum_{t=1}^{C}\sum_{i=1}^{P}\left(2L^2\e{\norm{\ww_{i,0}^t-\ww^\star}^2} + 2\sigma^2\right)\\
	&\leq \frac{2\sigma^2}{T} + \frac{2L^2P}{T^2}\sum_{t=1}^{C}\e{\norm{\ww_{1,0}^t-\ww^\star}^2}.
	\end{align*}
	Now using \Cref{lem:MB_bound} we have,
	\begin{align*}
	\e{\norm{\frac{1}{T}\sum_{t=1}^{C}\sum_{i=1}^{P}\xi_{i,1}^t}^2}&\leq \frac{2\sigma^2}{T} + \frac{2L^2P}{T^2}\sum_{t=1}^{C}\e{\norm{\hat{\ww}^{t-1}_{MB}-\ww^\star}^2}\\
	&\leq \frac{2\sigma^2}{T} + \frac{2L^2P}{T^2}\sum_{t=1}^{C}\left((1-\mu\eta)^{t-1}\norm{\ww^0 -\ww^\star}^2 + 2\sigma^2\frac{\eta\left(1-(1-\mu\eta)^C\right)}{\mu P}\right)\\
	&\leq \frac{2\sigma^2}{T} + \frac{2L^2P}{T^2}\left(\frac{1-(1-\mu\eta)^C}{\mu\eta}\norm{\ww^0 -\ww^\star}^2 + 2\sigma^2\frac{\left(C\mu\eta-(1-(1-\mu\eta)^C)\right)}{\mu^2 P}\right)\\
	&\leq \frac{2\sigma^2}{T} + \frac{2L^2P}{T^2}\left(\frac{1}{\mu\eta}\norm{\ww^0 -\ww^\star}^2 + 2\sigma^2\frac{C\eta}{\mu P}\right)\\
	&= A^2_{4,P,C} + A^2_{5,P,C}.
	\end{align*}
	Now using \Cref{lem:tech_5}, we have proved the lemma.
\end{proof}
It can be seen in the above lemma that there are two kinds of terms: one that depend on the history or initialization and second the ones that depend on the variance bound. This implies that it would be possible to restate \Cref{lem:MB_semifinal} as follows,
\begin{lemma}
	Under the assumptions ~\Cref{hyp:strong_convex},~\Cref{hyp:regularity},~\Cref{ass:def_filt},~\Cref{ass:lip_noisy_gradient_AS},~\Cref{hyp:variancebound} we have,
	\begin{align*}
	\e{\norm{\nabla^2F(\ww^\star)(\ww-\ww^\star)}^2} \leq 4(\hat{A}_{1,P,C}^2 + \hat{A}_{2,P,C}^2)
	\end{align*}
	Where the terms are respectively,
	\begin{align*}
	&\hat{A}_{1,P,C}^2=\frac{\norm{\ww^0-\ww^\star}^2}{\eta^2C^2}\left(1 + (1-\mu\eta)^C + \frac{2M^2}{\mu^2}\norm{\ww^0-\ws}^2 +\frac{2L^2\eta}{\mu P}\right),\\
	&\hat{A}_{2,P,C}^2=\frac{2\sigma^2}{T}\left(1+\frac{P}{T\eta\mu} +\frac{400 M^2 C^2 \eta^2 \sigma^2}{T \mu^2} + \frac{2L^2C\eta}{T \mu}\right).
	\end{align*}
\end{lemma}

Ignoring constants the above constants can be upper bounded as follows,
\begin{align*}
\hat{A}_{1,P,C}^2&\leq \frac{\norm{\ww^0-\ww^\star}^2}{\eta^2C^2}\left(1 + 1 + \frac{2M^2}{\mu^2}\norm{\ww^0-\ws}^2 +\frac{2L^2\eta}{\mu P}\right)\\
&\leq 2\frac{\norm{\ww^0-\ww^\star}^2}{\eta^2C^2}\left(1 + \frac{M^2}{\mu^2}\norm{\ww^0-\ws}^2 +\frac{L^2\eta}{\mu P}\right)\\
&\precsim \frac{\norm{\ww^0-\ww^\star}^2}{\eta^2C^2}\left(1 + \frac{M^2}{\mu^2}\norm{\ww^0-\ws}^2 +\frac{L^2\eta}{\mu P}\right),\\
\hat{A}_{2,P,C}^2&\leq 800\frac{\sigma^2}{T}\left(1+\frac{P}{T\eta\mu} +\frac{M^2 C^2 \eta^2 \sigma^2}{T \mu^2} + \frac{L^2C\eta}{T \mu}\right)\\
&\precsim \frac{\sigma^2}{T}\left(1+\frac{P}{T\eta\mu} +\frac{M^2 C^2 \eta^2 \sigma^2}{T \mu^2} + \frac{L^2C\eta}{T \mu}\right).
\end{align*} 
Thus, we recover \Cref{prop:conver_Mini_Batch}.

\subsection{Proof \Cref{prop:conver_One_Shot} (One-shot averaging case)}
To prove the proposition we need to prove a bound on second moment of the inner iterations followed by a bound on the final average outer iteration. For inner iterations we follow the result from \cite{moulines2011non} as the process on a single worker is completely independent of any other worker. We have the following lemma,

\begin{lemma}\label{lem:OS_semifinal}
	Under the Assumptions ~\Cref{hyp:strong_convex},~\Cref{hyp:regularity},~\Cref{ass:def_filt},~\Cref{ass:lip_noisy_gradient_AS},~\Cref{hyp:variancebound} for constant step size for one shot averaging we have,
	\begin{align*}
	\e{\norm{F''(\ww^\star)(\ww_{i,k}^1-\ww^\star)}^2} \leq 4\sum_{i=1}^{5}B_{i,P,N^1}^2
	\end{align*}
	where the terms are respectively,
	\begin{align*}
	&B_{1,P,N^1}^2=\frac{P^2}{T^2\eta^2}\norm{\ww^0-\ww^\star}^2, B_{2,P,N^1}^2=\frac{P^2}{T^2\eta^2}\left((1-\mu\eta)^{N^1}\norm{\ww^0-\ww^\star}^2 + \frac{2\sigma^2\eta}{\mu}\right),\\
	&B_{3,P,N^1}^2=\frac{P^2M^2}{T^2\mu\eta}\left(\norm{\ww^0-\ws}^2 + 20\eta^2 N^1\sigma^2\right)^2, B_{4,P,N^1}^2=\frac{2\sigma^2}{T},\\
	& B_{5,P,N^1}^2=\frac{2L^2P}{T^2}\left( \frac{1}{\mu\eta}\norm{\ww^0-\ww^\star}^2 + \frac{2\sigma^2N^1\eta}{\mu}\right).
	\end{align*}
\end{lemma}

\begin{proof}
	We follow the same line of proof as before. We can use the decomposition from \Cref{lem:parallel_decomposition} with constant step size and $C=1$, which results in the following simpler decomposition,
	\begin{align*}
	F''(\ww^\star)(\overline{\overline{\ww}}^C - \ws)  &= \frac{\ww^0-\ws}{N^\eta} - \frac{\hat{\ww}^1-\ws}{N^1\eta} + \frac{1}{T}\sum_{k=1}^{N^1}\sum_{i=1}^{P}\delta_{i,k}^1 + \frac{1}{T}\sum_{k=1}^{N^1}\sum_{i=1}^{P}\xi_{i,k}^1
	\end{align*} 
	For the first term, 
	\begin{align*}
	\e{\norm{\frac{\ww^0-\ws}{N^1\eta}}^2} &\leq \frac{P^2}{T^2\eta^2}\norm{\ww^0-\ww^\star}^2= B^2_{1,P,N^1}.
	\end{align*}
	For the second term using \Cref{lem:OS_bound} and rearranging we have,
	\begin{align*}
	\e{\norm{\frac{\hat{\ww}^1-\ws}{N^1\eta}}^2} &= \e{\norm{\frac{1}{PN^1\eta}\sum_{i=1}^{P}\ww_{i,N^1}^1-\ws}^2} \leq \frac{P}{T^2\eta^2}\sum_{i=1}^{P}\e{\norm{\ww_{i,N^1}^1-\ww^\star}^2}\\
	&\leq \frac{P^2}{T^2\eta^2}\left(\prod_{l=1}^{N^1}(1-\mu\eta) \norm{\ww^0-\ww^\star}^2 + 2\sigma^2\sum_{l=1}^{N^1}\prod_{m=l+1}^{N^1}(1-\mu\eta)\eta^2\right)\\
	&\leq \frac{P^2}{T^2\eta^2}\left((1-\mu\eta)^{N^1}\norm{\ww^0-\ww^\star}^2 + 2\sigma^2\frac{1-(1-\mu\eta)^{N^1}}{\mu \eta}\eta^2\right)\\
	&\leq \frac{P^2}{T^2\eta^2}\left((1-\mu\eta)^{N^1}\norm{\ww^0-\ww^\star}^2 + \frac{2\sigma^2\eta}{\mu}\right) = B^2_{2,P,N^1}.
	\end{align*}
	For the third term using \Cref{lem:tech_5} and \Cref{lem:tech_6} we obtain,
	\begin{align*}
	\e{\norm{\frac{1}{T}\sum_{i=1}^{P}\sum_{k=1}^{N^1}\delta_{i,k}^1}^2} &= \frac{1}{T^2}\e{\norm{\sum_{i=1}^{P}\sum_{k=1}^{N^1} F'(\ww_{i,k-1}^t)-F''(\ww^\star)(\ww_{i,k-1}^t-\ww^\star)}^2}\\
	&\leq \frac{P}{T^2}\sum_{i=1}^{P}\e{\norm{\sum_{k=1}^{N^1} F'(\ww_{i,k-1}^t)-F''(\ww^\star)(\ww_{i,k-1}^t-\ww^\star)}^2}\\
	&\leq \frac{P}{T^2}\sum_{i=1}^{P}\left(\sum_{k=1}^{N^1}\sqrt{\e{\norm{F'(\ww_{i,k-1}^1)-F''(\ww^\star)(\ww_{i,k-1}^1-\ww^\star)}^2}}\right)^2
	\end{align*}
	Now first using the upper bound of \Cref{hyp:regularity}, followed by \Cref{lem:moment_bound} and some rearranging we can obtain the following,
	\begin{align*}
	\e{\norm{\frac{1}{T}\sum_{i=1}^{P}\sum_{k=1}^{N^1}\delta_{i,k}^1}^2} &\leq \frac{PM^2}{T^2}\sum_{i=1}^{P}\left(\sum_{k=1}^{N^1}\e{\norm{\ww_{i,k-1}^1-\ws}^4}^{1/2}\right)^2\\
	&\leq \frac{PM^2}{T^2}\sum_{i=1}^{P}\left(\sum_{k=1}^{N^1}\left((1-\mu\eta)^{k-1}\e{\norm{\ww_{i,0}^1-\ws}^4}^{1/2} + \frac{20\eta\sigma^2}{\mu}\right)\right)^2\\
	&\leq \frac{P^2M^2}{T^2}\left(\sum_{k=1}^{N^1}\left((1-\mu\eta)^{k-1}\norm{\ww^0-\ws}^2 + \frac{20\eta\sigma^2}{\mu}\right)\right)^2\\
	&\leq \frac{P^2M^2}{T^2}\left(\frac{1-(1-\mu\eta)^{N^1}}{\mu \eta}\norm{\ww^0-\ws}^2 + \frac{20\eta N^1\sigma^2}{\mu}\right)^2\\	
	&\leq \frac{P^2M^2}{T^2\mu^2\eta^2}\left(\norm{\ww^0-\ws}^2 + 20\eta^2 N^1\sigma^2\right)^2 = B_{3,P,N^1}^2.
	\end{align*}
	
	For the fourth term, using the fact that on different machines noise of the gradient is i.i.d. over different iterations and zero in expectation (\Cref{ass:def_filt})  we obtain,
	\begin{align*}
	\e{\norm{\frac{1}{T}\sum_{i=1}^{P}\sum_{k=1}^{N^1}\xi_{i,k}^1}^2} &= \frac{1}{T^2}\sum_{i=1}^{P}\sum_{k=1}^{N^1}\e{\norm{\xi_{i,k}^1}^2}.
	\end{align*}
	Now using \Cref{lem:xi_bound} we have,
	\begin{align*}
	\e{\norm{\frac{1}{T}\sum_{i=1}^{P}\sum_{k=1}^{N^1}\xi_{i,k}^1}^2} &\leq \frac{1}{T^2}\sum_{i=1}^{P}\sum_{k=1}^{N^1}\left(2L^2\e{\norm{\ww_{i,k-1}^1-\ww^\star}^2} + 2\sigma^2\right)\\
	&\leq \frac{2\sigma^2}{T} + \frac{2L^2}{T^2}\sum_{i=1}^{P}\sum_{k=1}^{N^1}\e{\norm{\ww_{i,k-1}^1-\ww^\star}^2}.
	\end{align*}
	Now using \Cref{lem:OS_bound} we have,
	\begin{align*}
	\e{\norm{\frac{1}{T}\sum_{i=1}^{P}\sum_{k=1}^{N^1}\xi_{i,k}^1}^2} &\leq \frac{2\sigma^2}{T} + \frac{2L^2P}{T^2}\sum_{k=1}^{N^1}\left(\prod_{l=1}^{k-1}(1-\mu\eta) \norm{\ww^0-\ww^\star}^2 + 2\sigma^2\sum_{l=1}^{k-1}\prod_{m=l+1}^{k-1}(1-\mu\eta)\eta^2\right)\\
	&\leq \frac{2\sigma^2}{T} + \frac{2L^2P}{T^2}\sum_{k=1}^{N^1}\left((1-\mu\eta)^{k-1} \norm{\ww^0-\ww^\star}^2 + \frac{2\sigma^2\eta}{\mu}\right)\\
	&\leq \frac{2\sigma^2}{T} + \frac{2L^2P}{T^2}\left( \frac{1}{\mu\eta}\norm{\ww^0-\ww^\star}^2 + \frac{N^1 2\sigma^2\eta}{\mu}\right)\\
	&= B_{4,P,N^1}^2 + B_{5,P,N^1}^2.
	\end{align*}
	Finally using \Cref{lem:tech_6}, concludes the proof. 
\end{proof}

Similar to the mini-batch case, there are two kinds of terms one that depend on the history or initialization and second that depend on the variance bound of the functions. This implies that it would be possible to restate \Cref{lem:OS_semifinal} as follows,
\begin{lemma}
	Under the Assumptions ~\Cref{ass:def_filt}, ~\Cref{hyp:regularity}, ~\Cref{hyp:strong_convex}, ~\Cref{ass:lip_noisy_gradient_AS}, ~\Cref{hyp:variancebound} we have,
	\begin{align*}
	\e{\norm{\nabla^2F(\ww^\star)(\ww-\ww^\star)}^2} \leq 4(\hat{B}_{1,P,N^1}^2 + \hat{B}_{2,P,N^1}^2)
	\end{align*}
	Where the terms are respectively,
	\begin{align*}
	&\hat{B}_{1,P,N^1}^2=\frac{\norm{\ww^0-\ww^\star}^2}{(N^1)^2\eta^2}\left(1+(1-\mu\eta)^{N^1} + \frac{2M^2\eta}{\mu}\norm{\ww^0-\ws}^2 +\frac{2L^2\eta}{P\mu}\right),\\
	&\hat{B}_{2,P,N^1}^2=\frac{2\sigma^2}{T}\left(1+\frac{2L^2\eta}{\mu}+\frac{P^2}{T\mu\eta}+\frac{400M^2\sigma^{2}\eta^2 T }{\mu^{2}}\right).
	\end{align*}
\end{lemma}

On upper-bounding the above two terms while ignoring the constants, 
\begin{align*}
\hat{B}_{1,P,N^1}^2&\leq\frac{\norm{\ww^0-\ww^\star}^2}{(N^1)^2\eta^2}\left(1+1+ \frac{2M^2\eta}{\mu}\norm{\ww^0-\ws}^2 +\frac{2L^2\eta}{P\mu}\right)\\
&\leq 2\frac{\norm{\ww^0-\ww^\star}^2}{(N^1)^2\eta^2}\left(1+\frac{M^2\eta}{\mu}\norm{\ww^0-\ws}^2 +\frac{L^2\eta}{P\mu}\right)\\
&\precsim \frac{\norm{\ww^0-\ww^\star}^2}{(N^1)^2\eta^2}\left(1+\frac{M^2\eta}{\mu}\norm{\ww^0-\ws}^2 +\frac{L^2\eta}{P\mu}\right),\\
\hat{B}_{2,P,N^1}^2&\leq 800\frac{\sigma^2}{T}\left(1+\frac{L^2\eta}{\mu}+\frac{P^2}{T\mu\eta}+\frac{M^2\sigma^{2}\eta^2 T }{\mu^{2}}\right)\\
\hat{B}_{2,P,N^1}^2&\precsim \frac{\sigma^2}{T}\left(1+\frac{L^2\eta}{\mu}+\frac{P^2}{T\mu\eta}+\frac{M^2\sigma^{2}\eta^2 T }{\mu^{2}}\right).\\
\end{align*}
Thus we have recovered \Cref{prop:conver_One_Shot}.

\section{Proofs for OSA, MBA and Local-SGD in the online setting} \label{sec:proofsonline}

Recall that the step size at iteration $ (t,k),\in \un{C}\times \un{N^{t}} $ is defined as $\eta_k^t = \frac{c_\eta}{\left (\sum_{t'=1}^{t-1} \Nt+k\right )^\alpha}$ where $\alpha\in(0,1)$. Though our results can be extended for the entire range of learning rates, we prove results only for $\alpha \in (\frac{1}{2},1)$.

\subsection{Technical Lemmas}
We first state a few technical results which are helpful in the following proofs.
\begin{lemma}\label{lem:tech_1}
	For $\tilde{\eta}_m = \frac{c_\eta}{m^\alpha}$, $\alpha \in (0,1)$ we have $\prod_{m=1}^{t} (1-\mu\tilde{\eta}_m) \leq \exp\left(-\frac{\mu c_\eta t^{1-\alpha}}{2(1-\alpha)}\right)$ .
\end{lemma}
\begin{proof}
	The proof simply follows from applying the inequality $1+x\leq \exp\left(x\right)$, followed by an integral bound over the series as $\sum_{m=1}^{t}\frac{1}{m^\alpha} \geq \frac{1}{2}\int_{0}^{t}\frac{1}{m^\alpha} dm = \frac{t^{1-\alpha}}{1-\alpha}$. Note that it is possible to consider $\alpha= 1$ but the integral bound changes. For brevity we don't include it here. 
\end{proof}

\begin{lemma}\label{lem:tech_2}
	For $\tilde{\eta}_m = \frac{c_\eta}{m^\alpha}$, $\alpha \in (0,1)$ we have \[\sum_{m=1}^{t}(\tilde{\eta}_{m})^2\prod_{l=m+1}^{t}(1-\mu\tilde{\eta}_{l}) \leq \exp\left(-\frac{\mu c_\eta t^{1-\alpha}}{2(1-\alpha)}\left(1 - \frac{1}{2^{1-\alpha}}\right)\right)c_\eta^2\left(1+ \frac{t^{1-2\alpha}-1}{1-2\alpha} \right) + \frac{2c_\eta}{t^\alpha\mu}.\] Further if $\alpha \in (\frac{1}{2},1)$, then for large t, $\sum_{m=1}^{t}(\tilde{\eta}_{m})^2\prod_{l=m+1}^{t}(1-\mu\tilde{\eta}_{l}) \leq \exp\left(-\frac{\mu c_\eta t^{1-\alpha}}{2(1-\alpha)}\left(1 - \frac{1}{2^{1-\alpha}}\right)\right)\frac{2\alpha c_\eta^2}{2\alpha-1} + \frac{2c_\eta}{t^\alpha\mu}$. 
\end{lemma}
\begin{proof}
	First we decompose the term, then use $1+x \leq \exp(x)$, followed by a series of integral bounds like \Cref{lem:tech_1},
	\begin{align*}
	\sum_{m=1}^{t}\tilde{\eta}_{m}^2\prod_{l=m+1}^{t}(1-\mu\tilde{\eta}_{l}) &\leq \sum_{m=1}^{\frac{t}{2}}(\tilde{\eta}_{m})^2\prod_{l=m+1}^{t}(1-\mu\tilde{\eta}_{l}) + \sum_{m=\frac{t}{2}}^{t}(\tilde{\eta}_{m})^2\prod_{l=m+1}^{t}(1-\mu\tilde{\eta}_{l})\\
	&\leq \prod_{l=\frac{t}{2}+1}^{t}(1-\mu\tilde{\eta}_{l})\sum_{m=1}^{\frac{t}{2}}(\tilde{\eta}_{m})^2 + \sum_{m=\frac{t}{2}}^{t}\frac{\tilde{\eta}_{m}}{\mu}\left(\prod_{l=m+1}^{t}(1-\mu\tilde{\eta}_{l}) - \prod_{l=m}^{t}(1-\mu\tilde{\eta}_{l})\right)\\
	&\leq \exp\left(-\mu\sum_{l=\frac{t}{2}+1}^{t}\tilde{\eta}_{l}\right)\sum_{m=1}^{t}(\tilde{\eta}_{m})^2 + \frac{\tilde{\eta}_{\frac{t}{2}}}{\mu}\sum_{m=\frac{t}{2}}^{t}\left(\prod_{l=m+1}^{t}(1-\mu\tilde{\eta}_{l}) - \prod_{l=m}^{t}(1-\mu\tilde{\eta}_{l})\right)\\
	&\leq \exp\left(-\mu c_\eta\frac{t^{1-\alpha} -\left(\frac{t}{2}\right)^{1-\alpha}}{2(1-\alpha)}\right)\sum_{m=1}^{t}\frac{c_\eta^2}{m^{2\alpha}} + \frac{\tilde{\eta}_{\frac{t}{2}}}{\mu}\left(1-\prod_{l=\frac{t}{2}+1}^{t}(1-\mu\tilde{\eta}_{l})\right)\\
	&\leq \exp\left(-\frac{\mu c_\eta t^{1-\alpha}}{2(1-\alpha)}\left(1 - \frac{1}{2^{1-\alpha}}\right)\right)c_\eta^2\left(1+ \frac{t^{1-2\alpha}-1}{1-2\alpha} \right) + \frac{2c_\eta}{t^\alpha\mu}.
	\end{align*}
	The additional condition on $\alpha$ is obtained by simply taking the limiting case for $t\rightarrow \infty$. 
	Also note that this upper bound is tight up to constants (for both terms), especially one could easily show  $  \sum_{m=1}^{t}(\tilde{\eta}_{m})^2\prod_{l=m+1}^{t}(1-\mu\tilde{\eta}_{l}) \geq \frac{c_\eta}{2 t^\alpha\mu}.$
\end{proof}

\begin{lemma}\label{lem:tech_3}
	For the gamma function $\Gamma(s) = \int_{0}^{\infty}y^{s-1}\exp(-y) dy$ we have, $\sum_{t=1}^{C} \exp\left(-at^b\right)\leq \frac{1}{ba^{1/b}} \Gamma(\frac{1}{b})$. 
\end{lemma}
\begin{proof}
	First we use an integral bound as $\sum_{t=1}^{C} \exp\left(-at^b\right)\leq \int_{0}^{\infty} \exp\left(-az^b\right) dz $, followed by the integral substitution $u = az^b$ after which the proof follows from the definition of the gamma function.
\end{proof}

\begin{lemma}\label{lem:tech_8}
	For the gamma function $\Gamma(s) = \int_{0}^{\infty}y^{s-1}\exp(-y) dy$ we have, $\sum_{t=1}^{C} \frac{\exp\left(-at^b\right)}{t^c}\leq \frac{1}{ba^{(1-c)/b}} \Gamma(\frac{1-c}{b})$. 
\end{lemma}
\begin{proof}
	First we use an integral bound as $\sum_{t=1}^{C} \frac{\exp\left(-at^b\right)}{t^c}\leq \int_{0}^{\infty} \frac{\exp\left(-az^b\right)}{z^c} dz $, followed by the integral substitution $u = az^b$ after which the proof follows from the definition of the gamma function.
\end{proof}

\begin{lemma}\label{lem:tech_4}
	For $a \in (0,1)$,  $\sum_{t=1}^{C}\frac{1}{t^{1-a}} \leq \frac{C^{a}}{a}$. 
\end{lemma}
\begin{proof}
	It is a simple application of the integral bound on a decreasing function, $\sum_{t=1}^{C}\frac{1}{t^{1-a}} \leq \int_{0}^{C} x^{a-1} dx = \frac{C^a}{a}$.
\end{proof}

\begin{lemma}[Weighted Minkowski]\label{lem:tech_7}
	For $b_i \in \mathbb{R}$ and $a_i\in \mathbb{R}^d$, we have $\e{\norm{\sum_{i=1}^{P}a_ib_i}^2}\leq \left(\sum_{i=1}^{P}b_i\sqrt{\e{\norm{a_i}^2}}\right)^2$. 
\end{lemma}
\begin{proof}
	We consider again the norm $\norm{.}_E = \e{\norm{.}^2}^{\frac{1}{2}}$. Now the above result follows by first applying triangle inequality as $\norm{\sum_{i=1}^{P}a_ib_i}_E \leq \sum_{i=1}^{P}\norm{a_ib_i}_E$, followed by Holder's inequality to give $\sum_{i=1}^{P}b_i\norm{a_i}_E$.
\end{proof}
\subsection{Proof of \Cref{prop:conver_Mini_Batch_DSS} (Mini-batch Averaging Case)}
We have the following lemma for mini-batch averaging for the decreasing step-size case, 
\begin{lemma}\label{lem:MB_semifinal_DSS}
	Under the Assumptions ~\Cref{hyp:strong_convex}, ~\Cref{hyp:regularity}, ~\Cref{ass:def_filt}, ~\Cref{ass:lip_noisy_gradient_AS}, ~\Cref{hyp:variancebound} we have for mini-batch averaging,
	\begin{align*}
	\e{\norm{\nabla^2F(\ww^\star)(\ww-\ww^\star)}^2} \leq 5\sum_{i=1}^{6}C_{i,P,C}^2.
	\end{align*}
	Where the terms are,
	\begin{align*}
	C_{1,P,C}^2=&\frac{1}{C^2c_\eta^{2}}\norm{\ww^0-\ww^\star}^2,\\ 
	C_{2,P,C}^2=&\frac{4}{C^{2-2\alpha}c_\eta^{2}}\Bigg(\exp\left(-\frac{\mu c_\eta C^{1-\alpha}}{2(1-\alpha)}\right)\norm{\ww^0 -\ww^\star}^2\\&+ \frac{2\sigma^2}{P}\left(\exp\left(-\frac{\mu C^{1-\alpha}}{2(1-\alpha)}\left(1 - \frac{1}{2^{1-\alpha}}\right)\right)\frac{2\alpha c_\eta^2}{2\alpha-1} + \frac{2 c_\eta}{C^\alpha \mu}\right)\Bigg),\\
	C_{3,P,C}^2=&\frac{P^2\alpha^2}{T^2c_\eta^2}\Bigg(\beta_1\norm{\ww^0 -\ww^\star}^2 + \beta_2\frac{\sigma^2}{P} +\beta_3\frac{\sigma^2 C^{\alpha}}{P}\Bigg),\\
	C_{4,P,C}^2=&\frac{P^2M^2}{T^2}\left(2\beta_1^2\norm{\ww^0-\ws}^{4} +2\frac{400\sigma^{4}}{P^2}\left(\beta_2^2 + \beta_3^2C^{2-2\alpha}\right) \right),\\
	C_{5,P,C}^2=&\frac{2\sigma^2}{T} + \frac{2L^2P}{T^2}\left(\beta_1\norm{\ww^0 -\ww^\star}^2 + \beta_2 \frac{\sigma^2}{P} + \beta_3 \frac{\sigma^2 C^{1-\alpha}}{P}\right).
	\end{align*}
	And the constants are,
	\begin{align*}
	&\beta_1 = \frac{2^{\frac{1+3\alpha}{1-\alpha}}(1-\alpha)^{\frac{4\alpha-2}{1-\alpha}}}{(\mu c_\eta)^{\frac{2\alpha}{1-\alpha}}} \Gamma(\frac{\alpha}{1-\alpha})^2,\beta_2 = \frac{4^{\frac{1+2\alpha -\alpha^2}{(1-\alpha)}} (1-\alpha)^{\frac{2\alpha -1}{(1-\alpha)}}c_\eta^2}{(2\alpha-1)\left(\mu c_\eta (2^{1-\alpha}-1)\right)^{\frac{2\alpha}{(1-\alpha)}}} \Gamma\left(\frac{\alpha}{1-\alpha}\right)^2,\beta_3 = \frac{32c_\eta}{\alpha^2\mu},\\
	&\beta_4 = \frac{2^{\frac{1}{1-\alpha}} (1-\alpha)^{\frac{\alpha}{1-\alpha}} }{(\mu c_\eta)^{\frac{1}{1-\alpha}}}\Gamma\left(\frac{1}{1-\alpha}\right),\beta_5 = \frac{2^{\frac{3-2\alpha}{1-\alpha}} (1-\alpha)^{\frac{\alpha}{1-\alpha}}\alpha c_\eta^2}{(2\alpha-1)\left(\mu c_\eta (2^{1-\alpha}-1)\right)^{\frac{1}{1-\alpha}}}\Gamma\left(\frac{1}{1-\alpha}\right), \beta_6 = \frac{2c_\eta}{(1-\alpha)\mu}.
	\end{align*}
	
\end{lemma}

\begin{proof}
	Using again the decomposition in \Cref{lem:parallel_decomposition}, we can obtain the following simpler version for mini-batch averaging,
	\begin{align*}
	F''(\ww^\star)(\overline{\overline{\ww}}^C - \ws)  &= \frac{\ww^0-\ws}{C\eta_1^1} - \frac{\hat{\ww}^C-\ws}{C\eta_{2}^C} -\frac{1}{T}\sum_{t=1}^{C}\sum_{i=1}^{P}\left(\ww_{i,1}^t -\ws \right)\left(\frac{1}{\eta_1^t} - \frac{1}{\eta_{2}^t} \right)\\
	&+ \frac{1}{T}\sum_{t=1}^{C}\sum_{i=1}^{P}\delta_{i,1}^t + \frac{1}{T}\sum_{t=1}^{C}\sum_{i=1}^{P}\xi_{i,1}^t.
	\end{align*}
	Note again that we assume $\alpha \in (\frac{1}{2},1)$, just for the sake of brevity. For the first term,
	\begin{align*}
	\e{\norm{\frac{\ww^0-\ws}{C\eta_1^1}}^2} &= \frac{1}{C^2c_\eta^{2}}\norm{\ww^0-\ww^\star}^2 = C_{1,P,C}^2.
	\end{align*}
	
	For the second term using \Cref{lem:MB_bound_DSS}, followed by \Cref{lem:tech_1} and \Cref{lem:tech_2} we obtain,
	\begin{align*}
	\e{\norm{\frac{\hat{\ww}^C-\ws}{C\eta_{2}^C}}^2} &= \frac{(C+1)^{2\alpha}}{C^2c_\eta^{2}}\e{\norm{\ww^C_{MB}-\ww^\star}^2}\\
	&\leq \frac{2^{2\alpha}}{C^{2-2\alpha}c_\eta^{2}}\left(\prod_{m=1}^{C}(1-\mu\tilde{\eta}_{m})\e{\norm{\ww^0 -\ww^\star}^2} + 2\sigma^2\frac{1}{P}\sum_{m=1}^{C}(\tilde{\eta}_{m})^2\prod_{l=m+1}^{C}(1-\mu\tilde{\eta}_{l})\right)\\
	&\leq \frac{4}{C^{2-2\alpha}c_\eta^{2}}\Bigg(\exp\left(-\frac{\mu c_\eta C^{1-\alpha}}{2(1-\alpha)}\right)\norm{\ww^0 -\ww^\star}^2\\
	&+ \frac{2\sigma^2}{P}\left(\exp\left(-\frac{\mu C^{1-\alpha}}{2(1-\alpha)}\left(1 - \frac{1}{2^{1-\alpha}}\right)\right)\frac{2\alpha c_\eta^2}{2\alpha-1} + \frac{2c_\eta}{C^\alpha\mu}\right)\Bigg) = C_{2,P,C}^2\\
	\end{align*}
	
	For the third term using \Cref{lem:tech_7} and $(t+1)^\alpha-t^\alpha \leq \alpha t^{\alpha-1}$,
	\begin{align*}
	&\e{\norm{\frac{1}{T}\sum_{t=1}^{C}\sum_{i=1}^{P}\left(\ww_{i,1}^t -\ws \right)\left(\frac{1}{\eta_1^t} - \frac{1}{\eta_{2}^t} \right)}^2}\\
	&\leq \frac{1}{T^2c_\eta^2}\e{\norm{\sum_{t=1}^{C}\sum_{i=1}^{P}\left(\ww_{i,1}^t -\ws \right)\left( (t+1)^\alpha - t^\alpha \right)}^2}\\
	&\leq \frac{P^2\alpha^2}{T^2c_\eta^2}\left(\sum_{t=1}^{C}\left((t+1)^\alpha-t^\alpha\right)\sqrt{\e{\norm{\sum_{i=1}^{P}\left(\ww_{i,1}^t -\ws \right)}^2}} \right)^2\\
	&\leq \frac{P^2\alpha^2}{T^2c_\eta^2}\left(\sum_{t=1}^{C}t^{\alpha-1}\sqrt{\e{\norm{\ww_{MB}^t -\ws}^2}} \right)^2.
	\end{align*}
	Now using \Cref{lem:MB_bound_DSS}, \Cref{lem:tech_1}, \Cref{lem:tech_2} and $\sqrt{a+b} \leq \sqrt{a} + \sqrt{b}$ we get,
	\begin{small}
		\begin{align*}
		&\e{\norm{\frac{1}{T}\sum_{t=1}^{C}\sum_{i=1}^{P}\left(\ww_{i,1}^t -\ws \right)\left(\frac{1}{\eta_1^t} - \frac{1}{\eta_{2}^t} \right)}^2}\\
		&\leq \frac{P^2\alpha^2}{T^2c_\eta^2}\left(\sum_{t=1}^{C}t^{\alpha-1}\sqrt{\prod_{m=1}^{t}(1-\mu\tilde{\eta}_{m})\norm{\ww^0 -\ww^\star}^2 + 2\sigma^2\frac{1}{P}\sum_{m=1}^{t}(\tilde{\eta}_{m})^2\prod_{l=m+1}^{t}(1-\mu\tilde{\eta}_{l})} \right)^2\\
		&\leq \frac{P^2\alpha^2}{T^2c_\eta^2}\left(\sum_{t=1}^{C}t^{\alpha-1}\sqrt{\exp\left(-\frac{\mu c_\eta t^{1-\alpha}}{2(1-\alpha)}\right)\norm{\ww^0 -\ww^\star}^2 + \frac{2\sigma^2}{P}\left(\exp\left(-\frac{\mu c_\eta t^{1-\alpha}}{2(1-\alpha)}\left(1 - \frac{1}{2^{1-\alpha}}\right)\right)\frac{2\alpha c_\eta^2}{2\alpha-1} + \frac{2c_\eta}{t^\alpha\mu}\right)} \right)^2\\
		&\leq \frac{P^2\alpha^2}{T^2c_\eta^2}\Bigg(\sum_{t=1}^{C}t^{\alpha-1}\Bigg(\exp\left(-\frac{\mu c_\eta t^{1-\alpha}}{4(1-\alpha)}\right)\norm{\ww^0 -\ww^\star} + \sqrt{\frac{2\sigma^2}{P}\exp\left(-\frac{\mu c_\eta t^{1-\alpha}}{2(1-\alpha)}\left(1 - \frac{1}{2^{1-\alpha}}\right)\right)\frac{2\alpha c_\eta^2}{2\alpha-1}}\\ &+\sqrt{\frac{4c_\eta\sigma^2}{Pt^\alpha\mu}} \Bigg)\Bigg)^2\\
		&\leq \frac{P^2\alpha^2}{T^2c_\eta^2}\Bigg(\sum_{t=1}^{C}t^{\alpha-1}\exp\left(-\frac{\mu c_\eta t^{1-\alpha}}{4(1-\alpha)}\right)\norm{\ww^0 -\ww^\star} + \sum_{t=1}^{C}t^{\alpha-1}\sqrt{\frac{2\sigma^2 c_\eta^2 }{P(2\alpha-1)}\exp\left(-\frac{\mu c_\eta t^{1-\alpha}}{2(1-\alpha)}\left(1 - \frac{1}{2^{1-\alpha}}\right)\right)}\\
		&+\sum_{t=1}^{C}t^{\frac{\alpha}{2}-1}\sqrt{\frac{4c_\eta\sigma^2}{P\mu}}\Bigg)^2\\
		&\leq \frac{P^2\alpha^2}{T^2c_\eta^2}\Bigg(\sum_{t=1}^{C}t^{\alpha-1}\exp\left(-\frac{\mu c_\eta t^{1-\alpha}}{4(1-\alpha)}\right)\norm{\ww^0 -\ww^\star} + \sqrt{\frac{2\sigma^2 c_\eta^2 }{P(2\alpha-1)}}\sum_{t=1}^{C}t^{\alpha-1}\exp\left(-\frac{\mu c_\eta t^{1-\alpha}}{4(1-\alpha)}\left(1 - \frac{1}{2^{1-\alpha}}\right)\right)\\
		&+\sqrt{\frac{4c_\eta\sigma^2}{P\mu}}\sum_{t=1}^{C}\frac{1}{t^{1-\frac{\alpha}{2}}}\Bigg)^2.
		\end{align*}
	\end{small}	
	Now using \Cref{lem:tech_8} (with $b=1-\alpha$, $c=1-\alpha$ and $a=\frac{\mu c_\eta}{4(1-\alpha)}$), followed by using \Cref{lem:tech_8} again (with $a=\frac{\mu c_\eta}{4(1-\alpha)}\left(1-\frac{1}{2^{1-\alpha}}\right)$, $b=1-\alpha$ and $c=1-\alpha$) and \Cref{lem:tech_4} (with $a=\frac{\alpha}{2}$) we get, 
	\begin{align*}
	&\e{\norm{\frac{1}{T}\sum_{t=1}^{C}\sum_{i=1}^{P}\left(\ww_{i,1}^t -\ws \right)\left(\frac{1}{\eta_1^t} - \frac{1}{\eta_{2}^t} \right)}^2}\\
	&\leq \frac{P^2\alpha^2}{T^2c_\eta^2}\Bigg(\frac{4^{\frac{\alpha}{1-\alpha}}(1-\alpha)^{\frac{2\alpha-1}{1-\alpha}}}{(\mu c_\eta)^{\frac{\alpha}{1-\alpha}}} \Gamma(\frac{\alpha}{1-\alpha})\norm{\ww^0 -\ww^\star} + \sqrt{\frac{2\sigma^2 c_\eta^2 }{P(2\alpha-1)}}\frac{2^{\frac{\alpha(3-\alpha)}{1-\alpha}}(1-\alpha)^{\frac{2\alpha-1}{1-\alpha}}}{\left(\mu c_\eta (2^{1-\alpha}-1)\right)^{\frac{\alpha}{1-\alpha}}} \Gamma(\frac{\alpha}{1-\alpha})\\
	&+\sqrt{\frac{4c_\eta\sigma^2}{P\mu}}\frac{2C^{\frac{\alpha}{2}}}{\alpha}\Bigg)^2.
	\end{align*}
	Finally using \Cref{lem:tech_5} and re-organizing with constants defined as above,
	\begin{align*}
	&\e{\norm{\frac{1}{T}\sum_{t=1}^{C}\sum_{i=1}^{P}\left(\ww_{i,1}^t -\ws \right)\left(\frac{1}{\eta_1^t} - \frac{1}{\eta_{2}^t} \right)}^2}\\
	&\leq \frac{P^2\alpha^2}{T^2c_\eta^2}\Bigg(2\frac{4^{\frac{2\alpha}{1-\alpha}}(1-\alpha)^{\frac{4\alpha-2}{1-\alpha}}}{(\mu c_\eta)^{\frac{2\alpha}{1-\alpha}}} \Gamma(\frac{\alpha}{1-\alpha})^2\norm{\ww^0 -\ww^\star}^2 + 2\frac{2\sigma^2 c_\eta^2 }{P(2\alpha-1)}\frac{4^{\frac{\alpha(3-\alpha)}{(1-\alpha)}} (1-\alpha)^{\frac{2\alpha -1}{(1-\alpha)}}}{\left(\mu c_\eta (2^{1-\alpha}-1)\right)^{\frac{2\alpha}{(1-\alpha)}}} \Gamma\left(\frac{\alpha}{1-\alpha}\right)^2\\
	&+2\frac{4c_\eta\sigma^2}{P\mu}\frac{4C^{\alpha}}{\alpha^2}\Bigg)\\
	&\leq \frac{P^2\alpha^2}{T^2c_\eta^2}\Bigg(\beta_1\norm{\ww^0 -\ww^\star}^2 + \beta_2\frac{\sigma^2}{P} +\beta_3\frac{\sigma^2 C^{\alpha}}{P}\Bigg) = C^2_{3,P,C}.
	\end{align*}
	For the fourth term first proceeding as in \Cref{lem:MB_semifinal} with \Cref{lem:tech_5} and \Cref{lem:tech_6} we can obtain, 
	\begin{align*}
	\e{\norm{\frac{1}{T}\sum_{t=1}^{C}\sum_{i=1}^{P}\delta_{i,1}^t}^2}&=\frac{1}{T^2} \e{\norm{\sum_{t=1}^{C}\sum_{i=1}^{P}\left( F'(\ww_{i,0}^t)-F''(\ww^\star)(\ww_{i,0}^t-\ww^\star)\right)}^2}\\
	&\leq \frac{P}{T^2}\sum_{i=1}^{P} \e{\norm{\sum_{t=1}^{C}\left( F'(\hat{\ww}^{t-1})-F''(\ww^\star)(\hat{\ww}^{t-1}-\ww^\star)\right)}^2}\\
	&\leq \frac{P}{T^2}\sum_{i=1}^{P} \left(\sum_{t=1}^{C}\sqrt{\e{\norm{\left(F'(\hat{\ww}^{t-1})-F''(\ww^\star)(\hat{\ww}^{t-1}-\ww^\star)\right)}^2}}\right)^2\\
	&\leq \frac{PM^2}{T^2}\sum_{i=1}^{P} \left(\sum_{t=1}^{C}\sqrt{\e{(\hat{\ww}^{t-1} - \ws)^4}}\right)^2\\
	&\leq \frac{P^2M^2}{T^2}\left(\sum_{t=1}^{C}\sqrt{\e{(\ww_{MB}^{t-1} - \ws)^4}}\right)^2.
	\end{align*}
	Now using \Cref{lem:moment_bound_DSS}, followed by \Cref{lem:tech_1} and \Cref{lem:tech_2} we get\footnote{Note that we ignore t=1 in second inequality for second term as we have already incorporated it in the first term},
	\begin{align*}
	\e{\norm{\frac{1}{T}\sum_{t=1}^{C}\sum_{i=1}^{P}\delta_{i,1}^t}^2}&\leq \frac{P^2M^2}{T^2}\left(\sum_{t=1}^{C}\left(\prod_{j=1}^{t-1}\left(1-\tilde{\eta}_j\mu\right)\norm{\ww^0-\ws}^{2} +  \frac{20\sigma^{2}}{P}\sum_{j=1}^{t-1}(\tilde{\eta}_j)^2\prod_{l=j+1}^{t-1}(1-\mu\tilde{\eta}_l)\right)\right)^2\\
	&\leq \frac{P^2M^2}{T^2}\Bigg(\sum_{t=1}^{C}\exp\left(-\frac{\mu c_\eta (t-1)^{1-\alpha}}{2(1-\alpha)}\right)\norm{\ww^0-\ws}^{2}\\
	&+ \sum_{t=2}^{C}\frac{20\sigma^{2}}{P}\left(\exp\left(-\frac{\mu c_\eta (t-1)^{1-\alpha}}{2(1-\alpha)}\left(1 - \frac{1}{2^{1-\alpha}}\right)\right)\frac{2 \alpha c_\eta^2}{2\alpha-1} + \frac{2c_\eta}{(t-1)^\alpha\mu}\right) \Bigg)^2\\
	&\leq \frac{P^2M^2}{T^2}\Bigg(\sum_{t=1}^{C}\exp\left(-\frac{\mu c_\eta (t-1)^{1-\alpha}}{2(1-\alpha)}\right)\norm{\ww^0-\ws}^{2}\\
	&+ \sum_{t=1}^{C}\frac{20\sigma^{2}}{P}\left(\exp\left(-\frac{\mu c_\eta t^{1-\alpha}}{2(1-\alpha)}\left(1 - \frac{1}{2^{1-\alpha}}\right)\right)\frac{2\alpha c_\eta^2 }{2\alpha-1} + \sum_{t=1}^{C}\frac{2c_\eta}{t^\alpha\mu}\right) \Bigg)^2.
	\end{align*}
	Now using \Cref{lem:tech_3} (with $b=1-\alpha$ and $a=\frac{\mu c_\eta}{2(1-\alpha)}$), followed by \Cref{lem:tech_3} again (with $a=\frac{\mu c_\eta}{2(1-\alpha)}\left(1 - \frac{1}{2^{1-\alpha}}\right)$ and $b=1-\alpha$), followed by \Cref{lem:tech_4} (with $a=1-\alpha$) and \Cref{lem:tech_5} we get,
	\begin{align*}
	&\e{\norm{\frac{1}{T}\sum_{t=1}^{C}\sum_{i=1}^{P}\delta_{i,1}^t}^2}\\
	&\leq \frac{P^2M^2}{T^2}\Bigg(\frac{2^{\frac{1}{1-\alpha}} (1-\alpha)^{\frac{\alpha}{1-\alpha}} }{(\mu c_\eta)^{\frac{1}{1-\alpha}}}\Gamma\left(\frac{1}{1-\alpha}\right)\norm{\ww^0-\ws}^{2}\\
	&+\frac{20\sigma^{2}}{P}\left(\frac{2^{\frac{2-\alpha}{1-\alpha}} (1-\alpha)^{\frac{\alpha}{1-\alpha}} }{\left(\mu c_\eta (2^{1-\alpha}-1)\right)^{\frac{1}{1-\alpha}}}\Gamma\left(\frac{1}{1-\alpha}\right)\frac{2\alpha c_\eta^2 }{2\alpha-1} + \frac{2c_\eta C^{1-\alpha}}{(1-\alpha)\mu}\right) \Bigg)^2\\
	&\leq \frac{P^2M^2}{T^2}\Bigg(2\frac{2^{\frac{2}{1-\alpha}} (1-\alpha)^{\frac{2\alpha}{1-\alpha}} }{(\mu c_\eta)^{\frac{2}{1-\alpha}}}\Gamma\left(\frac{1}{1-\alpha}\right)^2\norm{\ww^0-\ws}^{4}\\
	&+2\frac{400\sigma^{4}}{P^2}\left(\frac{2^{\frac{4-2\alpha}{1-\alpha}} (1-\alpha)^{\frac{2\alpha}{1-\alpha}} }{\left(\mu c_\eta (2^{1-\alpha}-1)\right)^{\frac{2}{1-\alpha}}}\Gamma\left(\frac{1}{1-\alpha}\right)^2\frac{4\alpha^2 c_\eta^4 }{(2\alpha-1)^2} + \frac{4c_\eta^2 C^{2-2\alpha}}{(1-\alpha)^2\mu^2}\right) \Bigg)
	\end{align*}
	Bounding again with the constants defined above,
	\begin{align*}
	\e{\norm{\frac{1}{T}\sum_{t=1}^{C}\sum_{i=1}^{P}\delta_{i,1}^t}^2} \leq \frac{P^2M^2}{T^2}\left(2\beta_4^2\norm{\ww^0-\ws}^{4} +2\frac{400\sigma^{4}}{P^2}\left(\beta_5^2 + \beta_6^2C^{2-2\alpha}\right) \right) = C_{4,P,C}^2.
	\end{align*}
	For the fifth term, proceeding as in \Cref{lem:MB_semifinal},
	\begin{align*}
	\e{\norm{\frac{1}{T}\sum_{t=1}^{C}\sum_{i=1}^{P}\xi_{i,1}^t}^2} &= \frac{1}{T^2}\sum_{t=1}^{C}\sum_{i=1}^{P}\left(2L^2\e{\norm{\ww_{i,0}^t-\ww^\star}^2} + 2\sigma^2\right)\\
	&\leq \frac{2\sigma^2}{T} + \frac{2L^2P}{T^2}\sum_{t=1}^{C}\e{\norm{\ww_{1,0}^t-\ww^\star}^2}\\
	&\leq \frac{2\sigma^2}{T} + \frac{2L^2P}{T^2}\sum_{t=1}^{C}\e{\norm{\hat{\ww}^{t-1}_{MB}-\ww^\star}^2}.
	\end{align*}
	Now using \Cref{lem:MB_bound_DSS}, \Cref{lem:tech_1} and \Cref{lem:tech_2} like before,
	\begin{align*}	
	\e{\norm{\frac{1}{T}\sum_{t=1}^{C}\sum_{i=1}^{P}\xi_{i,1}^t}^2}&\leq \frac{2\sigma^2}{T} + \frac{2L^2P}{T^2}\sum_{t=1}^{C}\Bigg(\exp\left(-\frac{\mu c_\eta}{2(1-\alpha)}t^{1-\alpha}\right)\norm{\ww^0 -\ww^\star}^2\\
	&+ \frac{2\sigma^2}{P}\exp\left(-\frac{\mu c_\eta t^{1-\alpha}}{2(1-\alpha)}\left(1 - \frac{1}{2^{1-\alpha}}\right)\right)\frac{2\alpha c_\eta^2 }{2\alpha-1} + \frac{4\sigma^2c_\eta}{Pt^\alpha\mu}\Bigg).
	\end{align*}
	Further using \Cref{lem:tech_3} (with $b=1-\alpha$ and $a=\frac{\mu c_\eta}{2(1-\alpha)}$), followed by \Cref{lem:tech_3} again (with $a=\frac{\mu c_\eta}{2(1-\alpha)}\left(1 - \frac{1}{2^{1-\alpha}}\right)$ and $b=1-\alpha$), followed by \Cref{lem:tech_4} (with $a=1-\alpha$) and the constants as used above we get,
	\begin{align*}
	\e{\norm{\frac{1}{T}\sum_{t=1}^{C}\sum_{i=1}^{P}\xi_{i,1}^t}^2}&\leq \frac{2\sigma^2}{T} + \frac{2L^2P}{T^2}\Bigg(\frac{2^{\frac{1}{1-\alpha}} (1-\alpha)^{\frac{\alpha}{1-\alpha}} }{(\mu c_\eta)^{\frac{1}{1-\alpha}}}\Gamma\left(\frac{1}{1-\alpha}\right)\norm{\ww^0 -\ww^\star}^2\\
	&+ \frac{2^{\frac{2-\alpha}{1-\alpha}} (1-\alpha)^{\frac{\alpha}{1-\alpha}} }{\left(\mu c_\eta (2^{1-\alpha}-1)\right)^{\frac{1}{1-\alpha}}}\Gamma\left(\frac{1}{1-\alpha}\right)\frac{2\alpha c_\eta^2 }{2\alpha-1} + \frac{2c_\eta C^{1-\alpha}}{(1-\alpha)\mu}\Bigg)\\
	&\leq \frac{2\sigma^2}{T} + \frac{2L^2P}{T^2}\left(\beta_4\norm{\ww^0 -\ww^\star}^2 + \beta_5 \frac{\sigma^2}{P} + \beta_6 \frac{\sigma^2 C^{1-\alpha}}{P}\right) = C_{5,P,C}^2.
	\end{align*}
	Finally using \Cref{lem:tech_5} we have proved the lemma.
\end{proof}

The following lemma separates the terms above into bias and variance terms, following which we can easily prove \Cref{prop:conver_Mini_Batch_DSS},

\begin{lemma}\label{lem:MBAonline}
	Under the Assumptions ~\Cref{hyp:strong_convex}, ~\Cref{hyp:regularity}, ~\Cref{ass:def_filt}, ~\Cref{ass:lip_noisy_gradient_AS}, ~\Cref{hyp:variancebound} we have for mini-batch averaging,
	\begin{align*}
	\e{\norm{\nabla^2F(\ww^\star)(\ww-\ww^\star)}^2} \leq 5\left(\hat{C}_{1,P,C}^2 + \hat{C}_{2,P,C}^2\right) 
	\end{align*}
	Where for constants defined as above the terms are,
	\begin{align*}
	\hat{C}_{1,P,C}^2&=\frac{\norm{\ww^0-\ww^\star}^2}{C^2c_\eta^{2}}\left(1+ 4C^{2\alpha}\exp\left(-\frac{\mu c_\eta C^{1-\alpha}}{2(1-\alpha)}\right) + \alpha^2\beta_1 + 2M^2c_\eta^2\beta_1^2\norm{\ww^0-\ws}^2 + \frac{2L^2\beta_1c_\eta^2}{P}\right),\\
	\hat{C}_{2,P,C}^2&=\frac{2\sigma^2}{T}\Bigg(1+\frac{8\alpha C^{2\alpha-1}}{2\alpha-1}\exp\left(-\frac{\mu C^{1-\alpha}}{2(1-\alpha)}\left(1 - \frac{1}{2^{1-\alpha}}\right)\right) + \frac{8}{C^{1-\alpha}c_\eta \mu} + \frac{\alpha^2 \beta_2}{2 C c_\eta^2} + \frac{\alpha^2 \beta_3}{2 C^{1-\alpha} c_\eta^2}\\
	&+\frac{400 M^2 \sigma^2}{T}\left(\beta_2^2 + \beta_3^2 C^{2-2\alpha}\right) + \frac{L^2}{T}\left(\beta_2 + \beta_3 C^{1-\alpha}\right)\Bigg).
	\end{align*}
\end{lemma}

To get \Cref{prop:conver_Mini_Batch_DSS}, we upper bound every term up to constants depending only on $ \alpha $. Specifically, we use $ \beta_1\precsim  {(\mu c_\eta)^{-\frac{1}{1-\alpha}}}$, $ \beta_2\precsim  {(\mu c_\eta)^{-\frac{\alpha}{1-\alpha}}}$,  and $ \beta_3\precsim  \frac{ c_\eta}{\mu}$.

\subsection{Proof of \Cref{prop:conver_Mini_Batch_DSS} (One-shot Averaging case)}	
The analysis for the one-shot case is very similar to the mini-batch case, just like the constant step-size case. In fact at many place the communications $C$ of MBA get replaced by $N^1$ and the form of the bound remains the same. This intuitive conversion strengthens our analysis, which smoothly extends to both the extreme cases.

\begin{lemma}\label{lem:OS_semifinal_DSS}
	Under the Assumptions ~\Cref{hyp:strong_convex},~\Cref{hyp:regularity},~\Cref{ass:def_filt}, ~\Cref{ass:lip_noisy_gradient_AS},~\Cref{hyp:variancebound} for decreasing step size, for one shot averaging we have,
	\begin{align*}
	\e{\norm{\nabla^2F(\ww^\star)(\ww_{i,k}^1-\ww^\star)}^2} \leq 5\sum_{i=1}^{6}D_{i,P,C}^2
	\end{align*}
	where the terms are,
	\begin{align*}
	&D_{1,P,N^1}^2=\frac{P^2}{T^2c_\eta^{2}}\norm{\ww^0-\ww^\star}^2,D_{2,P,N^1}^2=\frac{4}{(N^1)^{2-2\alpha} c_\eta^2}\left(\exp\left(-\frac{\mu c_\eta (N^1)^{1-\alpha}}{1-\alpha}\right)\norm{\ww^0-\ww^\star}^2 + \frac{2\sigma^2 c_\eta}{\mu}\right),\\
	&D_{3,P,N^1}^2=\frac{P^2\alpha^2}{T^2c_\eta^2}\left(4\beta^2\norm{\ww^0 -\ww^\star}^2 + \frac{2\sigma^2 (N^1)^{2\alpha} c_\eta}{\mu \alpha^2} \right), D_{4,P,N^1}^2=\frac{P^2M^2}{T^2}\left(\beta\norm{\ww^0-\ws}^2 + \frac{20 \sigma^2 N^1 c_\eta}{\mu}\right)^2,\\
	&D_{5,P,N^1}^2=\frac{2\sigma^2}{T}, D_{6,P,N^1}^2 = \frac{2L^2P}{T^2}\left(\beta\norm{\ww^0-\ww^\star}^2 + \frac{2\sigma^2 N^1 c_\eta}{\mu}\right). 
	\end{align*}
	And the constants are $\beta_1 = 1 + \left(\frac{(1-\alpha)^\alpha}{\mu c_\eta}\right)^{\frac{1}{1-\alpha}}\Gamma\left(\frac{1}{1-\alpha}\right)$ and $\beta_2 = \left(2^\alpha\frac{(1-\alpha)^{2\alpha -1}}{(\mu c_\eta)^\alpha}\right)^{\frac{1}{1-\alpha}}\Gamma\left(\frac{\alpha}{1-\alpha}\right)$.
	
\end{lemma}

\begin{proof}
	We follow an analysis similar to \cite{God_Saa_2017}. We can simplify the decomposition from \Cref{lem:parallel_decomposition} for one outer phase as follows,
	\begin{align*}
	F''(\ww^\star)(\overline{\overline{\ww}}^C - \ws)  &= \frac{\ww^0-\ws}{N^1\eta_1^1} - \frac{\hat{\ww}^1-\ws}{N^1\eta_{N^1+1}^1} -\frac{1}{T}\sum_{i=1}^{P}\sum_{k=1}^{N^1}\big(\ww_{i,k}^1 -\ws \big)\Big(\frac{1}{\eta_k^1} - \frac{1}{\eta_{k+1}^1} \Big)\\
	&+ \frac{1}{T}\sum_{k=1}^{N^1}\sum_{i=1}^{P}\delta_{i,k}^1 + \frac{1}{T}\sum_{k=1}^{N^1}\sum_{i=1}^{P}\xi_{i,k}^1.
	\end{align*}
	For the first term, 
	\begin{align*}
	\e{\norm{\frac{\ww^0-\ws}{N^1\eta_1^1}}^2} &\leq \frac{P^2}{T^2c_\eta^{2}}\norm{\ww^0-\ww^\star}^2 = D^2_{1,P,N^1}.
	\end{align*}
	For the second term note that the inner iterate bound is independent for different machines using \Cref{lem:OS_bound_DSS} for say machine $1$, followed by \Cref{lem:tech_1} and \Cref{lem:tech_2} we get,
	\begin{align*}
	\e{\norm{\frac{\hat{\ww}^1-\ws}{N^1\eta_{N^1+1}^1}}^2} &\leq \frac{(N^1+1)^{2\alpha}}{{(N^1)}^{2} c_\eta^2}\e{\norm{\frac{1}{P}\sum_{i=1}^{P}\left(\ww_{i,N^1}^1-\ws\right)}^2}\\
	&\leq \frac{2^{2\alpha}}{(N^1)^{2-2\alpha} c_\eta^2}\e{\norm{\ww_{1,N^1}^1-\ws}^2}\\
	&\leq \frac{4}{(N^1)^{2-2\alpha} c_\eta^2}\left(\prod_{m=1}^{N^1}(1-\mu\eta_m^1) \norm{\ww^0-\ww^\star}^2 + 2\sigma^2\sum_{m=1}^{N^1}(\eta_m^1)^2\prod_{l=m+1}^{N^1}(1-\mu\eta_l^1)\right)\\
	&\leq \frac{4}{(N^1)^{2-2\alpha} c_\eta^2}\left(\exp\left(-\frac{\mu c_\eta {(N^1)}^{1-\alpha}}{1-\alpha}\right)\norm{\ww^0-\ww^\star}^2 + \frac{2\sigma^2 c_\eta}{\mu}\right) = D^2_{2,P,N^1}.\\
	\end{align*}
	
	For the third term using $(k+1)^\alpha - k^\alpha \leq \alpha k^{\alpha-1}$, \Cref{lem:tech_7}, and noting that the individual bounds on inner iterates for different machines are the same, thus using machine $1$ for brevity we can obtain,
	\begin{align*}
	\e{\norm{\frac{1}{T}\sum_{i=1}^{P}\sum_{k=1}^{N^1}\left(\ww_{i,k}^1 -\ws \right)\left(\frac{1}{\eta_k^1} - \frac{1}{\eta_{k+1}^1} \right)}^2} &\leq \frac{P^2\alpha^2}{T^2 c_\eta^2}\e{\norm{\sum_{k=1}^{N^1}k^{\alpha - 1}\left(\ww_{1,k}^1 -\ws \right)}^2}\\
	&\leq \frac{P^2\alpha^2}{T^2 c_\eta^2}\left(\sum_{k=1}^{N^1}k^{\alpha - 1}\sqrt{\e{\norm{\ww_{1,k}^1 -\ws}^2 }}\right)^2.\\
	\end{align*}
	Now using \Cref{lem:OS_bound_DSS}, \Cref{lem:tech_1}, \Cref{lem:tech_2} and $\sqrt{a+b} \leq \sqrt{a} + \sqrt{b}$ we get,
	\begin{align*}
	&\e{\norm{\frac{1}{T}\sum_{k=1}^{N^1}\sum_{i=1}^{P}\left(\ww_{i,k}^1 -\ws \right)\left(\frac{1}{\eta_k^1} - \frac{1}{\eta_{k+1}^1} \right)}^2}\\
	&\leq \frac{P^2\alpha^2}{T^2c_\eta^2}\left(\sum_{k=1}^{N^1}k^{\alpha-1}\sqrt{\e{\prod_{m=1}^{k}(1-\mu\tilde{\eta}_{m})\norm{\ww^0 -\ww^\star}^2 + 2\sigma^2\sum_{m=1}^{k}(\tilde{\eta}_{m})^2\prod_{l=m+1}^{k}(1-\mu\tilde{\eta}_{l})}} \right)^2\\
	&\leq \frac{P^2\alpha^2}{T^2c_\eta^2}\left(\sum_{k=1}^{N^1}k^{\alpha-1}\sqrt{\exp\left(-\frac{\mu c_\eta k^{1-\alpha}}{1-\alpha}\right)\norm{\ww^0 -\ww^\star}^2 + \frac{2\sigma^2 c_\eta}{\mu}} \right)^2\\
	&\leq \frac{P^2\alpha^2}{T^2c_\eta^2}\left(\sum_{k=1}^{N^1}k^{\alpha-1}\left(\exp\left(-\frac{\mu c_\eta k^{1-\alpha}}{2(1-\alpha)}\right)\norm{\ww^0 -\ww^\star} + \sqrt{\frac{2\sigma^2 c_\eta}{ \mu}}\right) \right)^2.
	\end{align*}
	Now using \Cref{lem:tech_8} again with $b=1-\alpha$ and $a=\frac{\mu c_\eta}{2(1-\alpha)}$ with $\beta_2$ defined as above and \Cref{lem:tech_4} we get, 
	\begin{align*}
	&\e{\norm{\frac{1}{T}\sum_{k=1}^{N^1}\sum_{i=1}^{P}\left(\ww_{i,k}^1 -\ws \right)\left(\frac{1}{\eta_k^1} - \frac{1}{\eta_{k+1}^1} \right)}^2}\\
	&\leq \frac{P^2\alpha^2}{T^2c_\eta^2}\left(\left(2^\alpha\frac{(1-\alpha)^{2\alpha -1}}{(\mu c_\eta)^\alpha}\right)^{\frac{1}{1-\alpha}}\Gamma\left(\frac{\alpha}{1-\alpha}\right)\norm{\ww^0 -\ww^\star} + \sqrt{\frac{2\sigma^2 (N^1)^{2\alpha} c_\eta}{\mu \alpha^2}} \right)^2\\
	&\leq \frac{P^2\alpha^2}{T^2c_\eta^2}\left(\beta_2\norm{\ww^0 -\ww^\star} + \sqrt{\frac{2\sigma^2 (N^1)^{2\alpha} c_\eta}{P\mu \alpha^2}} \right)^2\\
	&\leq \frac{P^2\alpha^2}{T^2c_\eta^2}\left(2\beta_2^2\norm{\ww^0 -\ww^\star}^2 + \frac{4\sigma^2 (N^1)^{2\alpha} c_\eta}{\mu \alpha^2} \right) = D_{3,P,N^1}^2.
	\end{align*}
	
	Now for the fourth term proceeding as in \Cref{lem:OS_semifinal} with \Cref{lem:tech_5} and \Cref{lem:tech_6} we can obtain ,
	\begin{align*}
	\e{\norm{\frac{1}{T}\sum_{i=1}^{P}\sum_{k=1}^{N^1}\delta_{i,k}^1}^2} &= \frac{1}{T^2}\e{\norm{\sum_{i=1}^{P}\sum_{k=1}^{N^1} F'(\ww_{i,k-1}^t)-F''(\ww^\star)(\ww_{i,k-1}^t-\ww^\star)}^2}\\
	&\leq \frac{P}{T^2}\sum_{i=1}^{P}\e{\norm{\sum_{k=1}^{N^1} F'(\ww_{i,k-1}^t)-F''(\ww^\star)(\ww_{i,k-1}^t-\ww^\star)}^2}\\
	&\leq \frac{P}{T^2}\sum_{i=1}^{P}\left(\sum_{k=1}^{N^1}\sqrt{\e{\norm{F'(\ww_{i,k-1}^1)-F''(\ww^\star)(\ww_{i,k-1}^1-\ww^\star)}^2}}\right)^2
	\end{align*}
	Now first using the upper bound of \Cref{hyp:regularity}, followed by \Cref{lem:moment_bound_DSS}, \Cref{lem:tech_1}, \Cref{lem:tech_2} and \Cref{lem:tech_3} we can obtain the following,
	\begin{align*}
	\e{\norm{\frac{1}{T}\sum_{i=1}^{P}\sum_{k=1}^{N^1}\delta_{i,k}^1}^2} &\leq \frac{PM^2}{T^2}\sum_{i=1}^{P}\left(\sum_{k=1}^{N^1}\e{\norm{\ww_{i,k-1}^1-\ws}^4}^{1/2}\right)^2\\
	&\leq \frac{P^2M^2}{T^2}\left(\sum_{k=1}^{N^1}\left(\prod_{j=1}^{k-1}(1-\eta_j^1\mu)\norm{\ww^0-\ws}^2 + 20 \sigma^{2}\sum_{j=1}^{k-1}\prod_{l=j+1}^{k-1}(1-\mu\eta_{l}^1)(\eta_j^1)^2\right)\right)^2\\
	&\leq \frac{P^2M^2}{T^2}\left(\sum_{k=1}^{N^1}\left(\exp\left(-\frac{\mu c_\eta (k-1)^{1-\alpha}}{1-\alpha}\right)\norm{\ww^0-\ws}^2 + \frac{20 \sigma^2 c_\eta}{\mu}\right)\right)^2\\
	&\leq \frac{P^2M^2}{T^2}\left(\left(1+ \left(\frac{(1-\alpha)^\alpha}{\mu c_\eta}\right)^{\frac{1}{1-\alpha}}\Gamma\left(\frac{1}{1-\alpha}\right)\right)\norm{\ww^0-\ws}^2 + \frac{20 \sigma^2 N^1 c_\eta}{\mu}\right)^2\\
	&\leq \frac{P^2M^2}{T^2}\left(\beta_1\norm{\ww^0-\ws}^2 + \frac{20 \sigma^2 N^1 c_\eta}{\mu}\right)^2 = D^2_{4,P,N^1}.
	\end{align*}
	
	For the fifth term, using the fact that for different machines noise is independent, zero in expectation (\Cref{ass:def_filt})  we obtain,
	\begin{align*}
	\e{\norm{\frac{1}{T}\sum_{i=1}^{P}\sum_{k=1}^{N^1}\xi_{i,k}^1}^2} &= \frac{1}{T^2}\sum_{i=1}^{P}\sum_{k=1}^{N^1}\e{\norm{\xi_{i,k}^1}^2}.
	\end{align*}
	Now using \Cref{lem:xi_bound} we have,
	\begin{align*}
	\e{\norm{\frac{1}{T}\sum_{i=1}^{P}\sum_{k=1}^{N^1}\xi_{i,k}^1}^2} &\leq \frac{1}{T^2}\sum_{i=1}^{P}\sum_{k=1}^{N^1}\left(2L^2\e{\norm{\ww_{i,k-1}^1-\ww^\star}^2} + 2\sigma^2\right)\\
	&\leq \frac{2\sigma^2}{T} + \frac{2L^2}{T^2}\sum_{i=1}^{P}\sum_{k=1}^{N^1}\e{\norm{\ww_{i,k-1}^1-\ww^\star}^2}.\\
	\end{align*}
	Now using \Cref{lem:OS_bound_DSS}, followed by \Cref{lem:tech_1}, \Cref{lem:tech_2} and \Cref{lem:tech_3} with definition of $\beta$ as before, and we have,
	\begin{align*}
	\e{\norm{\frac{1}{T}\sum_{i=1}^{P}\sum_{k=1}^{N^1}\xi_{i,k}^1}^2}
	&\leq \frac{2\sigma^2}{T} + \frac{2L^2P}{T^2}\sum_{k=1}^{N^1}\left(\prod_{m=1}^{k-1}(1-\mu\eta_m^1) \norm{\ww^0-\ww^\star}^2 + 2\sigma^2\sum_{m=1}^{k-1}(\eta_m^1)^2\prod_{l=m+1}^{k-1}(1-\mu\eta_l^1)\right)\\
	&\leq \frac{2\sigma^2}{T} + \frac{2L^2P}{T^2}\sum_{k=1}^{N^1}\left(\exp\left(-\frac{\mu c_\eta (k-1)^{1-\alpha}}{1-\alpha}\right)\norm{\ww^0-\ww^\star}^2 + \frac{2\sigma^2 c_\eta}{\mu}\right)\\
	&\leq \frac{2\sigma^2}{T} + \frac{2L^2P}{T^2}\left(\left(1+ \left(\frac{(1-\alpha)^\alpha}{\mu c_\eta}\right)^{\frac{1}{1-\alpha}}\Gamma\left(\frac{1}{1-\alpha}\right)\right)\norm{\ww^0-\ww^\star}^2 + \frac{2\sigma^2 N^1 c_\eta}{\mu}\right)\\
	&\leq \frac{2\sigma^2}{T} + \frac{2L^2P}{T^2}\left(\left(1+ \left(\frac{(1-\alpha)^\alpha}{\mu c_\eta}\right)^{\frac{1}{1-\alpha}}\Gamma\left(\frac{1}{1-\alpha}\right)\right)\norm{\ww^0-\ww^\star}^2 + \frac{2\sigma^2 N^1 c_\eta}{\mu}\right)\\
	&\leq \frac{2\sigma^2}{T} + \frac{2L^2P}{T^2}\left(\beta\norm{\ww^0-\ww^\star}^2 + \frac{2\sigma^2 N c_\eta}{\mu}\right) = D^2_{5,P,N^1} + D^2_{6,P,N^1}.\\
	\end{align*}
	Thus using \Cref{lem:tech_5} we have proved the lemma. 
\end{proof}

We can get the following lemma combining the bias and variance terms separately,

\begin{lemma}
	Under the Assumptions ~\Cref{hyp:strong_convex},~\Cref{hyp:regularity},~\Cref{ass:def_filt},~\Cref{ass:lip_noisy_gradient_AS},~\Cref{hyp:variancebound} for decreasing step size, for one shot averaging we have,
	\begin{align*}
	\e{\norm{\nabla^2F(\ww^\star)(\ww-\ww^\star)}^2} \leq 5\left(\hat{D}_{1,P,N^1}^2 + \hat{D}_{2,P,N^1}^2\right) 
	\end{align*}
	Where for constants defined as above the terms are,
	\begin{small}
		\begin{align*}
		\hat{D}_{1,P,N^1}^2&=\frac{\norm{\ww^0-\ww^\star}^2}{(N^1)^2c_\eta^{2}}\left(1+ 4(N^1)^{2\alpha}\exp\left(-\frac{\mu c_\eta (N^1)^{1-\alpha}}{2(1-\alpha)}\right) + \alpha^2\beta_1 + 2M^2c_\eta^2\beta_1^2\norm{\ww^0-\ws}^2 + \frac{2L^2\beta_1c_\eta^2}{P}\right),\\
		\hat{D}_{2,P,N^1}^2&=\frac{2\sigma^2}{T}\Bigg(1+\frac{8\alpha P (N^1)^{2\alpha-1}}{2\alpha-1}\exp\left(-\frac{\mu (N^1)^{1-\alpha}}{2(1-\alpha)}\left(1 - \frac{1}{2^{1-\alpha}}\right)\right) + \frac{8P}{(N^1)^{1-\alpha}c_\eta \mu} + \frac{\alpha^2 P \beta_2}{2 N^1 c_\eta^2} + \frac{\alpha^2 P \beta_3}{2 (N^1)^{1-\alpha} c_\eta^2}\\
		&+\frac{400 M^2 P \sigma^2}{N^1}\left(\beta_2^2 + \beta_3^2 (N^1)^{2-2\alpha}\right) + \frac{L^2}{N^1}\left(\beta_2 + \beta_3 (N^1)^{1-\alpha}\right)\Bigg).
		\end{align*}
	\end{small}	
\end{lemma}

\section{Brief overview of distributed optimization}\label{app:relatedwork}
The above three schemes (OSA, MBA, Local-SGD) are the most studied synchronous parallel schemes. However, communication latencies often make it difficult to use these algorithms for large-scale problems. Thus many alternative parallelization schemes which minimize communication or perform better have been studied. The major problem with some of these variants is that they are often difficult to tune, are not as stable and don't scale well to non-convex optimization problems. Result-wise, most of the machine learning packages use centralized mini-batch synchronous SGD.   

\textbf{Asynchronous SGD:} These techniques are characterized by avoiding a centralized synchronization, using delayed updates, maintaining parameter server estimates and being fault tolerant. Some of the notable references in a chronological order are \cite{langford2009slow,feng2011hogwild,agarwal2011delayed,paine2013GPU,Li2014communication,zhang2014deep,keuper2015parallel,de2015efficient, feyzmahdavian2015mini,lian2015asynchronous,mania2015perturbed,zhao2015fast,duchi2015asynchronous,chen2016revisiting,lian2017decentralized,pedregosa2017Nonsmooth,lian2017asynchronous,leblond2018improve, alistarh2018convergence}.\\     

\textbf{Federated optimization:} This setting is characterized by a huge number of mobile user devices, which run their local model in a decentralized manner with  often unbalanced data, but aim to train jointly. Many research questions still remain open but the direction is very relevant for distributed AI. Some references are \cite{jakub2015federated,jakub2016federated,mcmahan2016federated}.\\   

\textbf{Compressed Communication:} A common strategy to combat the communication overhead is to introduce lossless or lossy compression of exchanged information, often the gradients. Some of the work in this direction can be found in \cite{zhang2017zipml,wen2017terngrad,wangni2017gradient,sa2015taming,Na2017OnchipTO,gupta2015deep,alistarh2016randomized,khirirat2018distributed}.\\ 

\textbf{Non-SGD methods:} Many other optimization algorithms (coordinate descent, quasi newton, etc.) have also been studied in the parallel setting, owing to their better distributivity or convergence for some applications compared to the SGD algorithm. Some of them are \cite{2011boydADMM} (ADMM), \cite{shamir2014communication} (DANE), \cite{zhang2015communication} (DiSCO), \cite{reddi2016aide} (AIDE), \cite{ma2017distributed,smith2016cocoa,ma2015adding} (COCOA) and some of the references therein. Recently \cite{scaman2017optimal} gave provably optimal algorithms for the strongly convex and smooth functions for both synchronous and asynchronous cases. More broadly speaking, variance reduction methods are often the methods of choice in better understood, convex optimization problems [add reference]. Yet, their usage in the deep learning community has been relatively scarce, and often they are more difficult to parallelize [add reference]. Some of the works for instance are \cite{reddi2015variance,zhao2016fast,de2015efficient,lee2015distributed}. Among second order methods, quasi newton methods like distributed L-BFGS \cite{najafabadi2017large, simsekli2018MCMC} are also widely popular among the machine learning community.\\

\textbf{Communication Lower Bounds:} On a broader level our work is related to communication lower bounds which arise from information and learning-theoretic considerations. Unfortunately, these bounds are difficult to match for convex optimization as they are provided in \cite{arjevani2015communication}. Similar bounds have also been provided for the generally easier statistical estimation setting in \cite{duchi2014optimality,braverman2015communication,zhang2013information}.\\      

\textbf{Feature Distribution:} As clearly evident training data is not the only element of our optimization scheme which can be parallelized. Often in many problems in natural language processing and linear estimation, the features number in hundreds of thousands, and it might be of some interest to distribute the features alongside or beside training data. Some relevant references are \cite{lee2014model,ma2015partitioning,smith2016cocoa,chen2016communication,fang2018large}.\\

There has also been work in parallelizing stochastic optimization algorithms for specific problems (like PCA) in the past, for e.g., \cite{mcdonald2009efficient,mcdonald2010distributed,meng2012distributed, zhuang2013fast,li2014fast,Chin2015fast,Oh2015fast}.

We also provide a brief overview of some other techniques in distributed optimization in \Cref{app:relatedwork}.
\begin{table}
\resizebox{\linewidth}{!}{%
	\begin{tabular}{|c|c|p{8cm}|}
		\hline
		Reference  &  Setting  & Limitations\\ \hline 
		Zhang et. al. \cite{zhang2012communication}  &  OSA & Small learning rates $\frac{c}{\mu t}$; $\mu$ often unknown; Non-asymptotic bound on single worker convergence rate is used (\cite{rakhlin2012making}); \\
		Jain et. al. \cite{Jai_Kak_Kid_2016} & OSA, MBA & Results for least square regression (LSR) in finite horizon setting only; \\
		Godichon et. al. \cite{godichon2017rates} &  OSA & Uses uniform gradient bound \Cref{hyp:addit_noise} and thus not usable for LSR; Non-asymptotic result (\cite{rakhlin2012making}) is used;\\
		Stich \cite{stich2018local} & Local SGD & Small learning rates $\frac{c}{\mu t}$; $\mu$ often unknown; Uses uniform gradient bound \Cref{hyp:addit_noise} and thus not usable for LSR; Doesn't capture the need for an adaptive communication frequency \cite{Zha_DeS_Re_2016}; Doesn't extend to one-shot averaging, implying it is not tight enough; \\
		\hline
	\end{tabular}%
}
	\vspace{0.5em}
	\caption{Limitations of the previously existing results.}\label{tab:limit}
\end{table}

\end{document}